\documentclass[pdflatex,sn-mathphys-num]{sn-jnl}

\usepackage{amsmath,amsfonts}
\usepackage{array}
\usepackage{subfig}
\usepackage{textcomp}
\usepackage{stfloats}
\usepackage{url}
\usepackage{verbatim}
\usepackage{graphicx}
\usepackage{booktabs,makecell}
\usepackage{multirow}

\usepackage{siunitx}
\usepackage{amssymb}
\allowdisplaybreaks
\usepackage{hyperref}

\usepackage{wrapfig}
\usepackage{rotating}
\usepackage[skip=0.333\baselineskip]{caption}
\usepackage{relsize}
\usepackage{xfrac}
\usepackage{tabularx}
\usepackage[thinlines]{easytable}
\usepackage{mathtools}

\usepackage{longtable}
\usepackage{colortbl}
\usepackage{breqn}
\usepackage{soul}
\usepackage{tikz}
\usepackage{dcolumn}
\usepackage{threeparttable}
\usepackage{float}
\usepackage{enumitem}
\usepackage{afterpage}
\usepackage{pdflscape}
\usepackage{lipsum}
\usepackage[linesnumbered,ruled,vlined]{algorithm2e}
\usepackage[title]{appendix}%

\newcolumntype{C}[1]{>{\centering\arraybackslash}p{#1}}

\theoremstyle{thmstyleone}%
\theoremstyle{thmstyletwo}%

\theoremstyle{thmstylethree}%

\begin{document}

\title{Multi-Objective Optimization-Based Anonymization of Structured Data for Machine Learning Application}

\author*[1]{\fnm{Yusi} \sur{Wei}}\email{yw825@drexel.edu}

\author[1]{\fnm{Hande} \sur{Benson}}

\author[2]{\fnm{Joseph K} \sur{Agor}}

\author[3]{\fnm{Muge} \sur{Capan}}

\affil*[1]{\orgdiv{College of Business}, \orgname{Drexel University}, \orgaddress{\city{Philadelphia}, \state{PA}, \country{United States}}}

\affil[2]{\orgname{Johns Hopkins University Applied Physics Laboratory}, \orgaddress{\city{Laurel}, \state{MD}, \country{United States}}}

\affil[3]{\orgdiv{College of Engineering}, \orgname{University of Massachusetts Amherst}, \orgaddress{\city{Amherst}, \state{MA}, \country{United States}}}

\abstract
 {Organizations are collecting vast amounts of data, but they often lack the capabilities needed to fully extract insights. As a result, they increasingly share data with external experts, such as analysts or researchers, to gain value from it. However, this practice introduces significant privacy risks. Various techniques have been proposed to address privacy concerns in data sharing. However, these methods often degrade data utility, impacting the performance of machine learning (ML) models. Our research identifies key limitations in existing optimization models for privacy preservation, particularly in handling categorical variables, and evaluating effectiveness across diverse datasets. We propose a novel multi-objective optimization model that simultaneously minimizes information loss and maximizes protection against attacks. This model is empirically validated using diverse datasets and compared with two existing algorithms. We assess information loss, the number of individuals subject to linkage or homogeneity attacks, and ML performance after anonymization. The results indicate that our model achieves lower information loss and more effectively mitigates the risk of attacks, reducing the number of individuals susceptible to these attacks compared to alternative algorithms in some cases. Additionally, our model maintains comparable ML performance relative to the original data or data anonymized by other methods. Our findings highlight significant improvements in privacy protection and ML model performance, offering a comprehensive and extensible framework for balancing privacy and utility in data sharing.
}


\maketitle

\section{INTRODUCTION}
Data has become a critical asset for generating insights and supporting informed decision-making across diverse domains. In the entertainment industry, for example, Netflix leverages big data to power its recommendation algorithms, resulting in estimated annual savings of \$1 billion \cite{insideainews2018netflix}. In manufacturing, 72\% of executives report relying on data to boost productivity and efficiency \cite{bcg2020share}. In healthcare, clinical and genetic data are essential for advancing disease research and developing personalized treatments \cite{stark2025call}. These examples underscore the central role data plays in driving innovation, improving outcomes, and enabling evidence-based strategies. As a result, organizations are collecting vast amounts of data, but the ability to extract actionable insights often lies beyond the reach of those collecting it. These data owners may lack the technical expertise, analytical tools, or dedicated personnel to fully leverage the data they hold. Therefore, they often share this data with external analysts, such as researchers, consultants, or third-party experts, who have the skills to unlock the insights. 

This kind of collaboration is essential for innovation and impact. However, the increasing sharing of data raises serious concerns about privacy. As data sharing becomes more common, so do the risks associated with unauthorized access, re-identification, and data breaches. In 2024, privacy breaches affected 211\% more individuals compared to the previous year \cite{idtheft2024}. The financial impact is also escalating: global spending on data privacy protection is projected to reach approximately \$1.67 billion in 2024, representing a 24.6\% increase from 2023 \cite{dataprotection2024}. Industries such as financial services and healthcare account for 40\% of all reported breaches, with healthcare data breaches posing especially severe consequences. Many of these attacks have disrupted critical care delivery, in some cases leading to life-threatening outcomes for patients \cite{aha2024}. These trends underscore the urgent need to protect individual privacy.

Individuals have a fundamental right to control their personal information, and data owners—ranging from corporations to government agencies—have both legal and ethical responsibilities to uphold this right \cite{arbuckle2020building}. Regulatory frameworks such as the General Data Protection Regulation (GDPR) \cite{gdpr} in Europe, the California Consumer Privacy Act (CCPA) \cite{ccpa}, and the Health Insurance Portability and Accountability Act (HIPAA) \cite{hipaa} in the United States establish stringent requirements for data protection and impose significant penalties for non-compliance. In addition to regulatory mandates, maintaining user trust is essential for organizational reputation and operational success \cite{brough2022bulletproof}. Nonetheless, a growing body of research has shown that datasets de-identified under existing regulatory frameworks may still be vulnerable to re-identification and other privacy attacks \cite{el2011systematic, benitez2010evaluating, kwok2011harder, janmey2018re}. This highlights the limitations of current protections and reinforces the need for more effective privacy-preserving data sharing techniques.

A variety of privacy-preserving techniques have been developed to address the privacy risks associated with data publishing and sharing. Among these, anonymization is one of the most widely used approaches, replacing specific values with more generalized representations, such as age ranges or geographic regions, to reduce identifiability. A foundational model in this category is $k$-anonymity, which ensures that each individual’s record is indistinguishable from at least $k-1$ others in the dataset, thereby mitigating the risk of direct identification \cite{sweeney2002k}. However, $k$-anonymity does not explicitly protect sensitive information, such as disease history, financial records, which are not only highly confidential but also particularly vulnerable to re-identification. The leakage of such information can have serious consequences, including financial harm, discrimination, and in extreme cases, threats to patient safety. For instance, one report found that 2\% of hospital data breaches compromised sensitive medical information, affecting the health privacy of approximately 2.4 million patients \cite{healthcaredive2019breaches}. The limitations of $k$-anonymity in protecting sensitive attributes have led to the development of more robust privacy models, including $l$-diversity \cite{machanavajjhala2007diversity}, which ensures a minimum level of diversity in sensitive attribute values within each group, and $t$-closeness \cite{li2006t}, which limits the distributional distance of sensitive attributes between each group and the overall dataset. These models reflect the growing recognition that protecting sensitive information requires more than just preventing identity disclosure—it also demands mechanisms that reduce the risk of privacy attacks targeting sensitive content.

While these methods are effective in enhancing privacy, they inevitably alter the original data, often leading to significant information loss. This degradation in data quality can adversely affect the performance of machine learning (ML) models, resulting in reduced accuracy and reliability. Given the increasing reliance on ML in data-driven decision-making, preserving data utility while protecting privacy is essential. Consequently, a significant body of research focuses on developing algorithms and frameworks that achieve a balance between privacy protection and data utility \cite{zaki2024securing, turgay2023perturbation}. Optimization models can be effective in addressing the trade-off between privacy and utility. However, we have identified the following limitations in the application of optimization models for privacy preservation:
\begin{itemize}
    \item \textbf{Handling of Categorical Variables}: Information loss quantifies the deviation between the original and the privacy-preserved data. However, many existing studies struggle to effectively measure information loss for categorical variables within optimization models \cite{liang2020optimization}.
    \item \textbf{Evaluation with Diverse Datasets}: Data from a wide range of sectors, including financial, healthcare, retail, and education, are increasingly at risk of privacy attacks \cite{ranjan2025behavioural,al2024big,MARTIN2020449,reidenberg2018achieving}. Therefore, evaluating the effectiveness of privacy preservation models should involve diverse datasets. However, many models are tested on only a single dataset, limiting the understanding of their robustness and generalizability \cite{slijepvcevic2021k}.
\end{itemize}
Additionally, most existing frameworks adopt a structure where information loss is minimized subject to predefined privacy constraints. Given the critical importance of protecting sensitive information for individuals, we argue that a multi-objective framework is a more natural and flexible alternative. It enables the simultaneous optimization of both data utility and privacy, allowing solutions to be tailored to the varying needs and preferences of data owners or users. Therefore, the objective of our research is to address these identified limitations and advance the field of privacy-preserving data sharing through the development of a refined, multi-objective optimization model. The main contributions of this study are as follows:
\begin{enumerate}
    \item \textbf{Development of a multi-objective optimization model}: We propose a novel optimization model that simultaneously minimizes information loss and enhances the protection of sensitive information. To accurately capture information loss, our formulation distinguishes between numerical and categorical variables when measuring information loss. For the protection of sensitive attributes, we incorporate entropy as an objective function, leveraging its ability to reflect the diversity of sensitive values within each group. In parallel, the model enforces $k$-anonymity through constraints to mitigate the risk of re-identification. By combining these elements, our approach offers a comprehensive framework that balances privacy preservation with data utility, supporting reliable use in machine learning applications.
    \item \textbf{Evaluation of model effectiveness}: We empirically validate the effectiveness of our model using diverse datasets from finance or healthcare to ensure broad applicability and robustness. We execute our proposed model and the state-of-the-art from literature, and we quantify the level of information loss, the effectiveness against attacks, and the performance of ML models as measured by the F1 score for each approach.
\end{enumerate}

The remainder of this paper is structured as follows. In Section~\ref{sec:related work}, we present an overview of the related work on algorithms and models for privacy preservation. In Section~\ref{sec:prereq work}, we discuss the techniques used in our proposed optimization model. In Section~\ref{sec:proposed work}, we introduce our proposed multi-objective optimization model for privacy preservation. In Section~\ref{sec:experiment}, we explain the datasets used in this study and the experimental design. In Section~\ref{sec:results}, we report the experimental results and present the performance of ML algorithms with the proposed and alternative models. In Section~\ref{sec:conclusion}, we summarize our findings and provide directions for future research.

\section{LITERATURE REVIEW}\label{sec:related work}

In this section, we will introduce the concepts and definitions pertinent to this research topic and review the related works. This literature review will provide the necessary background and context for understanding the state of the art and the contributions of our research.

\subsection{Privacy Risks in Data Sharing}
\begin{figure}
    \centering
    \includegraphics[width=\columnwidth]{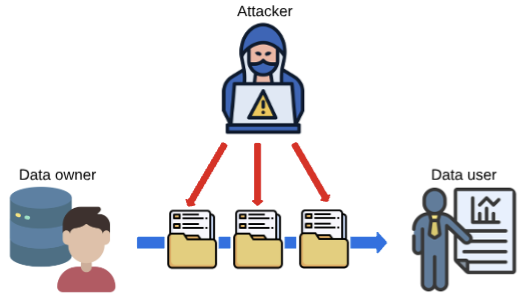}
    \caption{Overview of privacy risks in data sharing}
    \label{fig:ppds}
\end{figure}

Figure~\ref{fig:ppds} illustrates a typical scenario of privacy risks in the context of data sharing. In this setting, a data owner intends to share data with an external data user, such as a researcher or an organization, for statistical analysis or machine learning model development. However, adversaries may attempt to re-identify individuals in the dataset or infer sensitive information, posing significant threats to personal privacy.

Table~\ref{tab: original data1}(a) presents an example of an original dataset. In this dataset, some attributes—such as age, ZIP code, and income—are not unique identifiers by themselves but can uniquely identify individuals when combined. These attributes are referred to as quasi-identifiers (QIs). An adversary may exploit QIs in conjunction with publicly available data or external information to re-identify individuals, known as a {\em linkage attack}. For example, if an adversary knows that a 28-year-old woman living in ZIP code 19103 earns \$51,348, they could match this information to a unique entry in the dataset and consequently learn her medical condition. The medical condition represents a sensitive attribute (SA) variable containing private or confidential information about an individual. While SAs are not used for re-identification directly, they are often the target of inference in many privacy attacks. Therefore, protecting sensitive attributes is a critical aspect of privacy-preserving data publishing.

To mitigate the risk of linkage attacks, {\em $k$-anonymity} has been proposed. This technique ensures that each individual in the released dataset is indistinguishable from at least $k-1$ others based on QIs. Table~\ref{tab: original data1}(b) illustrates a $3$-anonymized version of the dataset, where individuals are grouped into equivalence classes—clusters of entries that share the same values for QIs. 

While $k$-anonymity effectively reduces the risk of re-identification through QIs, it does not guarantee protection of sensitive attributes. For example, in the first equivalence class in Table~\ref{tab: original data1}(b), all individuals have the same diagnosis—diabetes. If an adversary can associate someone with this group using QIs, they can still infer the person’s sensitive information. This type of privacy breach is referred to as a {\em homogeneity attack}, which arises when all values of a sensitive attribute within an equivalence class are identical. This vulnerability can persist even after applying $k$-anonymity.

To address this issue, {\em $l$-diversity} was introduced. This method ensures that each equivalence class contains at least $l\geq2$ distinct sensitive values, thereby reducing the risk of inference. As shown in Table~\ref{tab: original data1}(c), the modified dataset satisfies both $k$-anonymity and $l$-diversity, providing a more robust defense against both linkage and homogeneity attacks.

Despite these techniques, a central challenge remains: preserving individual privacy while maintaining sufficient data utility for downstream applications such as data mining and predictive modeling. The goal of privacy-preserving data publishing, therefore, is to strike a balance between minimizing privacy risks and retaining useful information in the released data \cite{fung2010privacy}. This trade-off highlights the need for advanced models and frameworks capable of simultaneously addressing privacy threats and utility requirements.

\begin{table}
    \centering
    \caption{Explanation of anonymization using example of original and anonymized data}
    \label{tab: original data1}
    \begin{tabular}{c}
         (a) Original Data \\
         \begin{tabular}{llll}
                \toprule
                Age & ZIP Code & Income & Disease  \\ \hline
                28  & 19103    & 51,348 & Diabetes \\ \hline
                29  & 19104    & 54,981 & Diabetes \\ \hline
                26  & 19104    & 52,003 & Diabetes \\ \hline
                30  & 19104    & 60,010 & Asthma \\ \hline
                31  & 19104    & 60,614 & Cancer   \\ \hline
                35  & 19103    & 64,715 & Asthma   \\ \hline
                40  & 19102    & 71,222 & Cancer   \\ \hline
                42  & 19102    & 74,820 & Flu      \\ \hline
                44  & 19102    & 74,173 & Obesity      \\ \bottomrule
            \end{tabular} \\ \\
        (b) 3-Anonymized Data Vulnerable to Homogeneity Attack \\
         \begin{tabular}{p{2cm}llll}
                \toprule
                \textbf{Equivalence Class} & \multicolumn{3}{c}{\textbf{QIs}} & \textbf{SA}                      \\ \hline
                                           & Age                               & ZIP Code    & Income  & Disease  \\ \hline
                \multirow{3}{*}{1}         & 25-30                             & 1910*       & 50K–55K & Diabetes \\ \cmidrule{2-5}
                                           & 25-30                             & 1910*       & 50K–55K & Diabetes \\ \cmidrule{2-5}
                                           & 25-30                             & 1910*       & 50K–55K & Diabetes \\ \hline
                \multirow{3}{*}{2}         & 30-35                             & 1910*       & 60K–65K & Asthma   \\ \cmidrule{2-5}
                                           & 30-35                             & 1910*       & 60K–65K & Cancer   \\ \cmidrule{2-5}
                                           & 30-35                             & 1910*       & 60K–65K & Asthma   \\ \hline
                \multirow{3}{*}{3}         & 40-45                             & 1910*       & 70K–75K & Cancer   \\ \cmidrule{2-5}
                                           & 40-45                             & 1910*       & 70K–75K & Flu      \\ \cmidrule{2-5}
                                           & 40-45                             & 1910*       & 70K–75K & Obesity      \\ \bottomrule
            \end{tabular} \\ \\
        (c) 3-Anonymized Data with 3-Diversity \\
         \begin{tabular}{p{2cm}llll}
                \toprule
                \textbf{Equivalence Class} & \multicolumn{3}{c}{\textbf{QIs}} & \textbf{SA}                      \\ \hline
                                           & Age                               & ZIP Code    & Income  & Disease  \\ \hline
                \multirow{5}{*}{1}         & 25-35                             & 1910*       & 50K–65K & Diabetes \\ \cmidrule{2-5}
                                           & 25-35                             & 1910*       & 50K–65K & Diabetes \\ \cmidrule{2-5}
                                           & 25-35                             & 1910*       & 50K–65K & Diabetes \\ \cmidrule{2-5}
                                           & 25-35                             & 1910*       & 50K–65K & Asthma \\ \cmidrule{2-5}
                                           & 25-35                             & 1910*       & 50K–65K & Cancer   \\ \hline
                \multirow{4}{*}{2}         & 35-45                             & 1910*       & 65K–75K & Asthma   \\ \cmidrule{2-5}
                                           & 35-45                             & 1910*       & 65K–75K & Cancer   \\ \cmidrule{2-5}
                                           & 35-45                             & 1910*       & 65K–75K & Flu 
                                           \\ \cmidrule{2-5}
                                           & 35-45                             & 1910*       & 65K–75K & Obesity 
                                           \\ \bottomrule
            \end{tabular}
    \end{tabular}
\end{table}


\subsection{Identification of Privacy Attack Risk}\label{sec: attack identification}
To identify individuals at risk of a linkage attack, we assume that adversaries can leverage publicly available sources to gain auxiliary information about the released dataset. Under this assumption, the probability of correct re-identification for an individual is given by $1/|E_g|$, where $|E_g|$ denotes the size of the equivalence class $g$ to which the individual belongs \cite{ElEmamKhaled2013MtPo}. An individual is considered at risk if this probability exceeds a predefined threshold $\tau$, that is, if $1/|E_g| \geq \tau$. 

In Table~\ref{tab: original data1}(a), each row represents a distinct equivalence class, resulting in a re-identification probability of 1 for every individual. If we set $\tau = 0.5$, then all individuals (a total of 9) are considered at risk of a linkage attack. However, after applying $k$-anonymity with $k = 3$, as shown in Table~\ref{tab: original data1}(b), the size of each equivalence class becomes 3, yielding a re-identification probability of $1/3 < 0.5$. In this case, no individual is at risk.

Although the choice of $\tau$ may vary depending on the acceptable level of risk, a fundamental strategy to mitigate linkage attacks remains the same: increasing the size of equivalence classes. As the class size grows, the likelihood of correct re-identification decreases, effectively reducing the number of individuals deemed at risk. Therefore, controlling the number and size of equivalence classes is critical for enhancing privacy protection against linkage attacks.

Machanavajjhala et al. \cite{machanavajjhala2007diversity} demonstrated that individuals in equivalence classes where all sensitive attribute values are identical are susceptible to homogeneity attacks. In Table~\ref{tab: original data1}(b), all individuals in equivalence class $E_1$ share the same sensitive attribute value, ``Diabetes''. As a result, the number of individuals at risk of a homogeneity attack in this table is $|E_1| = 3$. To mitigate this risk, the $l$-diversity principle can be applied. After enforcing $l$-diversity with $l = 3$, as shown in Table~\ref{tab: original data1}(c), each equivalence class contains at least three distinct sensitive attribute values. Consequently, no individual is considered at risk of a homogeneity attack.

\subsection{Privacy Preservation Algorithms and Models}

\begin{table}
\centering
\caption{Summary of Methods in Privacy Preservation}
\label{tab:privacy_preservation_summary}
\renewcommand{\arraystretch}{1.2}
\begin{tabular}{lp{3cm}p{3cm}}
\toprule
\multicolumn{3}{c}{Privacy Preservation Algorithms and Models} \\
\hline
\multirow{2}{2cm}{\raggedright Perturbative Methods: The original values in the dataset are modified} & \raggedright Semantic Methods: Adding noise & $\epsilon$-differential privacy, $(\epsilon, \sigma)$-differential privacy\\
\cmidrule{2-3}
 & \raggedright Syntactic Methods: Employing a clustering framework & $k$-anonymity, $l$-diversity, $t$-closeness, $\beta$-likeness, $\theta$-sensitive $k$-anonymity\\
\hline
\multirow{2}{1.5cm}{\raggedright Non-perturbative Methods: The original values in the dataset are not modified} & \raggedright Encryption: Ensuring that only authorized users can decrypt and access it & Advanced Encryption Standard (AES), RSA Algorithm, Homomorphic Encryption\\
\cmidrule{2-3}
 & \raggedright Federated Learning: Training ML models across multiple decentralized devices or servers while keeping the data localized & Federated Averaging (FedAvg), Secure Aggregation Protocols\\
\bottomrule
\end{tabular}
\end{table}

Privacy preservation methods can be classified into two main categories based on whether the original values in the dataset are modified: perturbative and non-perturbative methods \cite{willenborg-2001,domingo2001disclosure}, as outlined in Table~\ref{tab:privacy_preservation_summary}. Perturbative methods involve distorting the original data before its publication to protect individual privacy. These methods can be further divided into semantic and syntactic models \cite{khan2022tau}. Differential privacy is one of the semantic privacy models. 

On the other hand, syntactic privacy models employ a clustering framework to form equivalence classes, ensuring that data within each class is indistinguishable from one another \cite{khan2022tau}. This process is also called data anonymization. Syntactic approaches include $k$-anonymity \cite{sweeney2002k, sweeney2002achieving}, $l$-diversity \cite{machanavajjhala2007diversity}, $t$-closeness \cite{li2006t}, $\beta$-likeness \cite{cao2012publishing} and $\theta$-sensitive $k$-anonymity \cite{khan2020theta}. Syntactic privacy models can be further categorized into microaggregation \cite{domingo2005ordinal} and generalization \cite{fung2005top,wang2004bottom}. Microaggregation replaces the QI values in an equivalence class with the centroid of the equivalence class. In contrast, generalization replaces QI values with broader, less specific values, such as intervals. 

Numerous generalization-based algorithms have been developed to achieve data anonymization, including the works of \cite{khan2022tau, chen2023sensitivity, wang2020enhanced, amiri2023enhancing}, etc. Additionally, Doka et al.\cite{doka2015k} and Liang and Samavi\cite{liang2020optimization} introduced an optimization model grounded in generalization principles. Liang and Samavi\cite{liang2020optimization} employed the Loss Metric, an information loss evaluation metric proposed by \cite{iyengar2002transforming}, as its objective function and incorporates constraints to achieve $k$-anonymity. However, this optimization model encounters challenges when dealing with categorical variables \cite{liang2020optimization}. The nature of the objective function, which calculates the difference between the maximum and minimum values, is not inherently meaningful for categorical data. For instance, calculating the subtraction between the 7th category and the 1st category does not provide a useful measure of information loss. In addition, \cite{doka2015k} and \cite{liang2020optimization} merely consider $k$-anonymity and do not prevent homogeneity attacks, which is the issue addressed by $l$-diversity.

Microaggregation-based algorithms have been also extensively explored, including  \cite{batista2022privacy, singh2022social, aleroud2024privacy, abidi2018hybrid, wu2019micro}. In addition, Aminifar et al.\cite{aminifar2021diversity} proposed an optimization model based on microaggregation. Their model's objective function is geared towards minimizing the sum of within-group distances. To achieve this, they established a QI space, where individual records were represented as points within this space. These points were subsequently grouped into QI clusters. Within each cluster, the QI values of the records were replaced with the centroid values of the respective QI group. This clustering process was optimized under specified constraints to produce a database that satisfied the criteria for $k$-anonymity, $l$-diversity, and $t$-closeness. However, a notable limitation in their approach lies in the use of the Manhattan distance metric as the objective function. This choice poses challenges when dealing with categorical variables, as Manhattan distance is not inherently suitable for measuring dissimilarity between categorical attributes.

In addition, several studies \cite{dewri2011exploring, lin2024exploring, halawi2023multi, sugitha2024multi, ahamad2022multi, jahan2025analysis, sadeghi2025optimizing, jahan2024dynamic} have identified two primary objectives in privacy preservation: maximizing privacy and maximizing utility. These works highlight the adaptability of multi-objective optimization models in addressing a wide range of privacy concerns across various data types. The multi-objective framework has proven to be highly versatile, enabling researchers to balance competing goals and accommodate diverse data structures. However, many of these studies \cite{jahan2025analysis, sadeghi2025optimizing, jahan2024dynamic} primarily focus on enhancing the performance of heuristic algorithms—such as genetic algorithms and particle swarm optimization—to more efficiently explore the solution space and incur less computational cost. Less attention is given to tailoring the optimization model itself to specific domain needs or real-world constraints. In contrast, this study emphasizes the development of a novel, domain-specific optimization model that explicitly captures the privacy-utility trade-off in structured data. While a heuristic algorithm is employed to solve the model, it functions purely as a computational tool; the core contribution lies in the formulation of the model rather than the improvement of the solution method.

\subsection{Effects of Privacy Preservation Models on the Performance of ML Models}

ML algorithms are widely applied to data analysis, making it imperative to examine the effect of anonymization on their performance. Evaluating the impact of data anonymization is crucial for assessing the trade-off between privacy preservation and the accuracy of ML models. 

 Oprescu et al.\cite{oprescu2022energy} assessed the impact of $k$-anonymity on ML models, including Logistic Regression, $k$-Nearest Neighbor, and Gradient Boosting algorithms. They implemented $k$-anonymity through two approaches: generalization and suppression, as well as microaggregation. Their findings indicated that, particularly for larger and more complex datasets, the decline in model accuracy was minimal. Additionally, they observed that the effect of $k$-anonymity significantly depends on the specific dataset and the anonymization technique used.

Senavirathne and Torra\cite{senavirathne2020role} investigated the effects of various anonymization techniques, including generalization, microaggregation, and differential privacy, on deep neural networks using three different datasets. Their findings revealed that current data anonymization methods fail to achieve an optimal trade-off between privacy and utility, highlighting the need for new methods to overcome these challenges. Their study also indicates that when the level of anonymization is low, the accuracy of ML models remains comparable to that of the original accuracy, and there is a substantial decline in data utility for multi-class classification problems.

Pitoglou et al.\cite{pitoglou2022exploring} evaluated the impact of data anonymization on the performance of various ML models, including Logistic Regression, Decision Trees, $k$-Nearest Neighbors, Support Vector Machines, and Gaussian Naive Bayes. They employed the Mondrian algorithm, a greedy anonymization technique that ensures $k$-anonymity through generalization, and tested its performance under different combinations of QIs and values of $k$ on real-world healthcare data. Their findings indicate that the degree of accuracy loss in ML models varies based on the choice of QIs and the level of anonymity (i.e., the value of $k$) used during anonymization. Moreover, the selection of QIs significantly influences the performance of ML models on anonymized datasets. They emphasize the importance of tuning hyperparameters in ML models when assessing the impact of anonymization, suggesting that appropriate anonymization techniques and carefully chosen hyperparameters can mitigate the negative effects of anonymization.

Based on this literature review, we conclude the following:
\begin{itemize}
\item The impact of \textbf{different anonymization algorithms} on ML varies.
\item The impact of anonymization on ML differs with \textbf{different datasets (including different sizes)}.
\item The impact of \textbf{different levels of anonymization} exhibits various outcomes in ML.
\item For \textbf{different ML models}, the impact of anonymization varies.
\end{itemize}

After reviewing 1106 papers published between 2005 and 2025 across Google Scholar, ACM Digital Library, IEEE Xplore, Wiley Online Library, Web of Science, and ABI/INFORM, we have identified 16 papers that are highly relevant to our study. These papers reveal a significant gap in the application of optimization models for privacy preservation. Thus, our objective is to address these limitations and contribute to the advancement of privacy-preserving techniques through the refinement and innovation of multi-objective optimization models. In addition, we aim to systematically evaluate the impact of different hyperparameter settings on the performance of ML models. We intend to provide guidance on which hyperparameter settings are most suitable for specific datasets and ML models. Our research aims to offer actionable guidance for researchers preparing data and using data to make decisions or gain insights.

\section{PRELIMINARIES}\label{sec:prereq work}

Table~\ref{tab: original data1}(a) is an example of original data with 9 records. Let $x_{ij}$ represent the value of $j$th QI for record $i$, and let its anonymized value be denoted as $x'_{ij}$. After anonymization, assume that we have $n_E$ equivalence classes, which are sets $E_g$, $g = 1,\ldots,n_E$ wherein all records have the same anonymized values for the QIs.

\subsection{Entropy $l$-diversity}

Machanavajjhala et al.\cite{machanavajjhala2007diversity} proposed the $l$-diversity principle to mitigate homogeneity attacks by enforcing that each equivalence class contains at least $l \geq 2$ distinct values for a given sensitive attribute. To quantify the degree of diversity of sensitive attributes within each equivalence class, they leveraged the information-theoretic concept of entropy. The entropy of an equivalence class $E_g$ for a given sensitive attribute whose possible values are represented as the set $SA$ is computed as follows:

\begin{equation}\label{eq1:entropy l-diverse}
    \text{entropy}(E_g, SA) = \sum\limits_{s \in SA}-\frac{n(E_g,s)}{|E_g|}\log_2\left(\frac{n(E_g,s)}{|E_g|}\right)
\end{equation}
where $n(E_g,s)$ refers to the number of individuals whose sensitive attribute value is $s$ in equivalence class $E_g$, and $|E_g|$ refers to the size of equivalence class $g$. For instance, we compute the entropy values for the sensitive attribute `Disease' within three equivalence classes using Table~\ref{tab: original data1}(b): $\text{entropy}(E_1, \text{`Disease'})=0$, $\text{entropy}(E_2, \text{`Disease'})=0.91$, and $\text{entropy}(E_3, \text{`Disease'})=1.58$. As previously highlighted, records within $E_1$ are vulnerable to a homogeneity attack. The calculated $\text{entropy}(E_1, \text{`Disease'})$ represents the minimum entropy value across the three equivalence classes. This observation underscores that a lower entropy represents a higher degree of identical sensitive attribute values, implying a greater risk of homogeneity attacks. Conversely, a higher entropy, such as $\text{entropy}(E_3, \text{`Disease'})=1.58$, suggests a lower risk of homogeneity attacks, as it indicates a more diverse group of sensitive attribute values in the dataset. 

The entropy of a sensitive attribute $SA$ is calculated as the minimum entropy across all equivalence classes for $SA$:
    \begin{align}\label{eq2:entropy l-diverse}
        \text{entropy}(SA) = & \min_{g \in 1..n_E} \Bigl\{\text{entropy}(Eg, SA)\Bigr\} 
    \end{align}
Maximizing the total entropy of all sensitive attributes can enhance protection against homogeneity attacks.

\subsection{Information Loss}
Information loss (IL) refers to the deviation of the original data from the anonymized data\cite{mauger2020multi}. It can also serve as a surrogate measure of data utility, with higher information loss typically indicating lower utility for downstream tasks. For numeric data \cite{domingo2001disclosure, domingo2002practical}, the deviation between $x_{ij}$ and $x'_{ij}$ is measured as
\begin{equation} \label{eq:SSE(NQI)}
    (x_{ij}-x'_{ij})^2,
\end{equation}
whereas for categorical data, it is computed as follows:
\begin{equation} \label{eq:SSE(CQI)2}
    \delta(x_{ij},x'_{ij}) = \begin{cases}
    1, \quad x_{ij} \neq x'_{ij} \\
    0, \quad x_{ij} = x'_{ij} .
    \end{cases} 
\end{equation}
There is already deviation present in the dataset, as well.  For numeric data, we measure this deviation as 
\begin{equation} \label{eq:SST(NQI)}
    (x_{ij}-\bar{x}_{j})^2,
\end{equation}
where $\bar{x}_{j}$ is the mean value for QI $j$ across all records.  For categorical data, we measure it as 
\begin{equation} \label{eq:SST(CQI)}
    \delta(x_{ij},\widehat{x}_{j}),
\end{equation}
where $\widehat{x}_{j}$ is the mode for QI $j$ across all records.
To calculate IL, the total deviation across all QIs and records is scaled by the total deviation present in the data itself.


\section{PROPOSED APPROACH}\label{sec:proposed work}
In this section, we describe our proposed multi-objective anonymization model (MO-OBAM) that aims to simultaneously minimize information loss, maximize the protection for sensitive attributes, and maintain $k$-anonymity. 

\subsection{Problem Statement}
In privacy-preserving data sharing, the trade-off between privacy and utility can be addressed in two primary ways: through constraint-based formulations or multi-objective optimization models. In constraint-based approaches, one objective—typically minimizing information loss—is optimized, while privacy is enforced through constraints such as $k$-anonymity or $l$-diversity. This structure is illustrated in Equations~\eqref{eq: k constraint} and~\eqref{eq: l constraint}, where Equation~\eqref{eq: k constraint} ensures that each equivalence class contains at least $k$ individuals, and Equation~\eqref{eq: l constraint} guarantees that each class includes at least $l$ distinct sensitive attribute values. This method offers interpretability and is widely used in the literature (e.g., \cite{liang2020optimization}, \cite{zheng2019effective}, \cite{10.1145/2714576.2714590}, \cite{10.1007/978-3-030-66172-4_5}). However, it can be restrictive and may not fully capture the nuanced trade-offs between privacy and utility.
\begin{align}
    \text{minimize} &\quad IL \nonumber \\
    \text{subject to } &\quad |E_g| \geq k, \forall E_g \label{eq: k constraint}\\
    & \quad |\{s \in SA : n(E_g, s) > 0\}| \geq l \quad \forall E_g \label{eq: l constraint}
\end{align}

To provide greater flexibility, we adopt a multi-objective optimization framework in which both privacy and utility are treated as competing objectives: we minimize information loss while maximizing privacy. This allows us to generate a set of trade-off solutions, giving data owners the ability to explore different anonymization strategies depending on their priorities. While multi-objective models often rely on heuristic or metaheuristic methods to navigate complex solution spaces, in our study, the heuristic method is used solely as a computational tool to solve the model efficiently, without being the focus of the methodological contribution.


\subsection{Proposed Model: MO-OBAM}
Suppose we are provided a dataset that has $n$ records and $n_{QI}$ QIs. Among the QIs, $n_{NQI}$ of them are numerical while $n_{CQI}$ are categorical ($n_{QI}=n_{NQI}+n_{CQI}$). WLOG, we assume that the QIs are ordered such that the indices $j=1,\ldots, n_{NQI}$ correspond to numerical data and $j=n_{NQI}+1,\ldots, n_{NQI}+n_{CQI}$ correspond to categorical data. There are also $n_{SA}$ sensitive attributes. The values of sensitive attribute $j$ for record $i$ are denoted as $SA_{ij}$.

In our model, anonymization of the original data results in clusters with respect to QIs of records where each record can only belong to one cluster. The centroids of the clusters are used to replace the original values of QIs in the data to achieve anonymization so that each cluster is an equivalence class. Denoting the number of clusters in anonymized data as $n_C$, we have two sets of decision variables: the set of centroids for clusters, $q_c=\{q_{c1},\ldots,q_{cn_{QI}}\}$, $c=1,\ldots n_C$, and binary variables $w_{ic}$ representing the membership of record $i$ in cluster $c$, $i=1,\ldots n$ and $c=1,\ldots n_C$. Each record is assigned to only one cluster; therefore, we have an assignment constraint as follows:
\begin{equation}\label{eq:assignment}
    \sum_{c=1}^{n_C} w_{ic} = 1, \quad i=1,\ldots,n.
\end{equation}

To formulate information loss, we first apply Equation~\eqref{eq:SSE(NQI)} to calculate the deviation after anonymizing numerical data:
\begin{equation}\label{eq:SSE(NQI)app}
    \sum_{i=1}^{n} w_{ic}\sum_{j=1}^{n_{NQI}}(x_{ij}-q_{cj})^2, \quad c=1,\ldots,n_C. 
\end{equation}
Then we apply Equation~\eqref{eq:SSE(CQI)2} to calculate the deviation after anonymizing categorical data:
\begin{equation}
    \sum_{i=1}^{n} w_{ic} \sum_{j=n_{NQI}+1}^{n_{NQI}+n_{CQI}} \delta( \,x_{ij}, q_{cj}), \quad c=1,\ldots,n_C \label{eq:SSE(CQI)app}.
\end{equation}
To clarify the computation, we define the distance between each record $i$ and centroids of cluster $c$ as follows:
\begin{align}\label{eq:distance}
\sum_{j=1}^{n_{NQI}} (x_{ij} - q_{cj})^2 + \sum_{j=n_{NQI}+1}^{n_{NQI}+n_{CQI}} \delta(x_{ij}, q_{cj}).
\end{align}
Therefore, the total deviation due to anonymization is calculated as follows:
\begin{align}\label{eq:SSE app}
 \sum_{c=1}^{n_C} \sum_{i=1}^{n} w_{ic} \Bigg( \sum_{j=1}^{n_{NQI}} (x_{ij} - q_{cj})^2 
 + \sum_{j=n_{NQI}+1}^{n_{NQI}+n_{CQI}} \delta(x_{ij}, q_{cj}) \Bigg)
\end{align}

And then we can apply Equation~\eqref{eq:SST(NQI)} and~\eqref{eq:SST(CQI)} to calculate total deviation present in the data itself.
\begin{equation}\label{eq:SST app}
    \sum_{j=1}^{n_{NQI}}\sum_{i=1}^{n}(x_{ij}-\overline{X}_j)^2+\sum_{j=n_{NQI}+1}^{n_{NQI}+n_{CQI}}\sum_{i=1}^{n}\delta(x_{ij}, \widehat{X}_j)
\end{equation}
where $\overline{X}_j$ is the mean of the $j$th NQI  over the entire dataset, and $\widehat{X}_j$ is the mode of the $j$th CQI over the entire dataset. One of our objectives is to minimize information loss. The information loss $IL$ resulting from the anonymization process is formulated as $IL=\frac{\eqref{eq:SSE app}}{\eqref{eq:SST app}}$. 


Another objective of our model is to maximize the protection of sensitive information. To quantify this protection, we incorporate entropy as a measure in the objective function. The intuition is that low entropy within a cluster indicates that the majority of individuals in that cluster share the same sensitive attribute value, making it easier for an adversary to infer sensitive information, hence representing lower privacy protection.

To assess the privacy level, we calculate the entropy of each cluster with respect to a given sensitive attribute. Specifically, for the $j$th sensitive attribute, let $SA_j$ denote the set of possible values. The entropy of each equivalence class is computed using Equation~\eqref{eq1:entropy l-diverse}. Let $w_c$, $c = 1, \ldots, n_C$, represent the equivalence classes induced by the clustering assignment $w$. Then the entropy for each $w_c$ with respect to $SA_j$ is given by:
\begin{align}
    \text{entropy}(w_c,SA_{j}) = & \sum\limits_{s \in SA_{j}}-p(w_c,s)\log_2\left(p(w_c,s)\right), \label{eq:entropy1} \nonumber \\
    j= n_{NQI}+n_{CQI}&+1,\ldots,n_{NQI}+n_{CQI}+n_{SA} \\
   p(w_c,s) = & \frac{\sum\limits_{i=1}^{n}w_{ic}\mathbb{I}(SA_{ij}=s)}{\sum\limits_{i=1}^{n}w_{ic}} \label{eq:entropy fraction}  \\
    \mathbb{I}(SA_{ij}=s)= & \begin{cases}
        1 & SA_{ij}=s \\
        0 & \text{otherwise}
    \end{cases} \label{eq:indicator}
\end{align}
where $\sum\limits_{i=1}^{n}w_{ic}\mathbb{I}(SA_{ij}=s)$ is to the number of records with sensitive value $s$ in cluster $c$, $\sum\limits_{i=1}^{n}w_{ic}$ is the number of records in cluster $c$, and $p(w_c,s)$ is the fraction of records in cluster $c$ with sensitive value equal to $s$. Then, Equation~\eqref{eq2:entropy l-diverse} can be specified as follows:
\begin{align}
    \text{entropy}(SA_{j}) = & \min_{c \in 1 \ldots n_C} \Bigl\{\text{entropy}(w_c, SA_{j})\Bigr\} \label{eq:entropy2}
\end{align}

To enhance the protection of sensitive information, we aim to maximize the minimum entropy across all clusters. Equation~\eqref{obj2:sum of entropy} is the second objective function in our model.
\begin{alignat}{1}\label{obj2:sum of entropy}
\sum\limits_{j=n_{NQI}+n_{CQI}+1}^{n_{NQI}+n_{CQI}+n_{SA}}\text{entropy}(SA_{j}) 
\end{alignat}
This approach ensures that even the least diverse cluster maintains a high level of uncertainty regarding sensitive attribute values, thereby reducing the risk of homogeneity attacks.

To ensure $k$-anonymity, we add a constraint to the model that each cluster must contain at least $k$ records:
\begin{equation}\label{eq:k_anonymity}
   \sum_{i=1}^{n}w_{ic} \geq k, \quad c = 1,\ldots,n_C.
\end{equation}
This constraint ensures that each cluster has enough records to protect against linkage attacks.

Putting all of the equations together gives the following multi-objective model:
\begin{align*}
    & \text{minimize}_{q,w} \quad IL - \lambda \cdot \eqref{obj2:sum of entropy} \\
    & \text{subject to}  \quad ~\eqref{eq:assignment},~\eqref{eq:entropy1},~\eqref{eq:entropy fraction},~\eqref{eq:indicator},~\eqref{eq:entropy2},~\eqref{eq:k_anonymity}
\end{align*}
where $\lambda$ is a hyperparameter to balance the two objective functions. Table~\ref{tab:parameters table} summarizes all notations in the model. 

\begin{table}
    \centering
    \caption{Summary of notations in our model}
    \label{tab:parameters table}    
    \begin{tabular}{m{1.5cm}m{5cm}} \toprule
    \textbf{Notation} & \textbf{Definition}  \\\hline
    $i$ & Record index \\\hline
    $j$ & Variable index \\\hline
    $c$ & Cluster index \\\hline
    $n$ & Total number of records in the entire data \\\hline
    $n_{NQI}$ & Number of numeric quasi-identifiers \\\hline
    $n_{CQI}$ & Number of categorical quasi-identifiers \\\hline
    $n_{SA}$ & Number of sensitive attributes \\\hline
    $n_C$ & Number of clusters \\\hline
    $\overline{X_j}$ & Mean of the $j$th numeric quasi-identifier \\\hline
    $\widehat{X_j}$ & Mode of the $j$th categorical quasi-identifier \\\hline
    $x_{ij}$ & Value of the $i$th record for the $j$th variable in QIs \\\hline
    $s$ & Sensitive value \\\hline
    $SA_{ij}$ & Value of the $i$th record for the $j$th sensitive attribute  \\\hline
    $k$ & Number of records in a cluster \\\hline
    $\lambda$ & Controls the trade-off between objective functions \\\hline
    $q_{cj}$ & Centroid of the $c$th cluster for the $j$th variable \\\hline
    $w_{ic}$ & Membership of the $i$th record in the $c$th cluster \\\bottomrule
    \end{tabular}
    
    \end{table}

In summary, the proposed multi-objective optimization model has two sets of decision variables: $q_{cj}$, representing the cluster centroids used to replace the original values in the data, and $w_{ic}$, representing the cluster membership for each record. In addition, the objective function of the optimization model serves two purposes. Firstly, it aims to minimize information loss during the anonymization process, ensuring that the anonymized data retains as much useful information as possible. Secondly, it seeks to maximize the sum of entropy to enhance protection for sensitive information, mitigating homogeneity attacks. The constraints ensure that each record is assigned to only one cluster and each cluster has at least $k$ records to protect against linkage attacks.


\subsection{Optimization Approach}
\begin{algorithm}[ht]
\caption{Optimization of MO-OBAM using PSO}\label{alg:pso_mo_obam}
\DontPrintSemicolon
\KwData{$n_{\text{particles}}$, $n_{\text{iterations}}$, original dataset, $n_C$, $\lambda$, $k$, $l_{multi}$}
\KwResult{Optimal anonymization with minimized IL, maximized entropy, and satisfied $k$-anonymity}

Initialize particles by randomly selecting values from QIs\;
Initialize personal best and global best solutions\;

\For{$i = 1$ \KwTo $n_{\text{iterations}}$}{
    \ForEach{particle}{
        \ForEach{record in the dataset}{
            Compute Equation~\eqref{eq:distance} for each cluster\;
            Assign the record to the cluster with the minimum value\;
        }
        Compute fitness: $\text{fit} = IL - \lambda * \eqref{obj2:sum of entropy} - l_{multi} * \sum (k - \text{cluster size})$\;
        \If{fitness better than personal best}{
            Update personal best\;
        }
    }
    Update global best based on best particle\;
    Update positions using PSO rules\;
}
\Return best-found anonymization\;
\end{algorithm}

Since the two objective functions operate on different scales, normalization may be required to ensure that the sensitivity of $\lambda$ is appropriately maintained. To solve this optimization problem, we employ the Particle Swarm Optimization (PSO) algorithm \cite{kennedy1995particle}, a population-based metaheuristic inspired by the social behavior of birds flocking or fish schooling. PSO is well-suited for complex, nonlinear optimization problems and efficiently explores the solution space by iteratively updating the position and velocity of each particle based on both individual experience and the collective experience of the swarm.

Algorithm~\ref{alg:pso_mo_obam} gives a detailed description of the PSO for solving the MO-OBAM model. We first define the number of particles ($n_{\text{particles}}$), where each particle represents a candidate solution for the model. Specifically, each solution is represented by a matrix of decision variables $q_{cj}$. We also specify the number of iterations ($n_{\text{iterations}}$), which determines the duration of the optimization process. Each particle is initialized by randomly selecting values from the original quasi-identifiers to construct its $q_{cj}$ matrix. During each iteration, an additional set of decision variables $w_{ic}$ is determined based on the centroids $q_{cj}$ defined by the particle. Using these assignments, we compute the objective value and evaluate any constraint violations. Each particle retains its best-performing solution over time, referred to as its \textit{personal best}, while the best solution across the entire swarm is tracked as the \textit{global best}. After each iteration, particles update their $q_{cj}$ values by taking into account their current $q_{cj}$, personal best, and the global best solution. This process continues until the predefined number of iterations is reached. The final global best solution corresponds to the optimized anonymized dataset. For a detailed analysis of the convergence properties of PSO, see Xu and Yu \cite{XU201865}.

Additionally, to evaluate the effectiveness of our proposed model in terms of information loss, protection against attacks, and the impact on ML model performance, we conduct a comparative analysis using anonymized data from our model and two existing algorithms alongside the original datasets, serving as the baseline. Therefore, we assess several hypotheses regarding our proposed model. The hypotheses are as follows:
\begin{itemize}
    \item[H1:] Information loss resulting from our model is lower than the two alternative algorithms.
    \item[H2:] Our model provides further protection against both linkage and homogeneity attacks by reducing the number of records that are at risk of such attacks 
    \item[H3:] Our model will not negatively impact the performance of ML models, as measured by the F1 score.
\end{itemize}


\section{EXPERIMENTIAL SETUP}\label{sec:experiment}
\begin{figure*}[!t]
    \centering
    \includegraphics[width=\columnwidth]{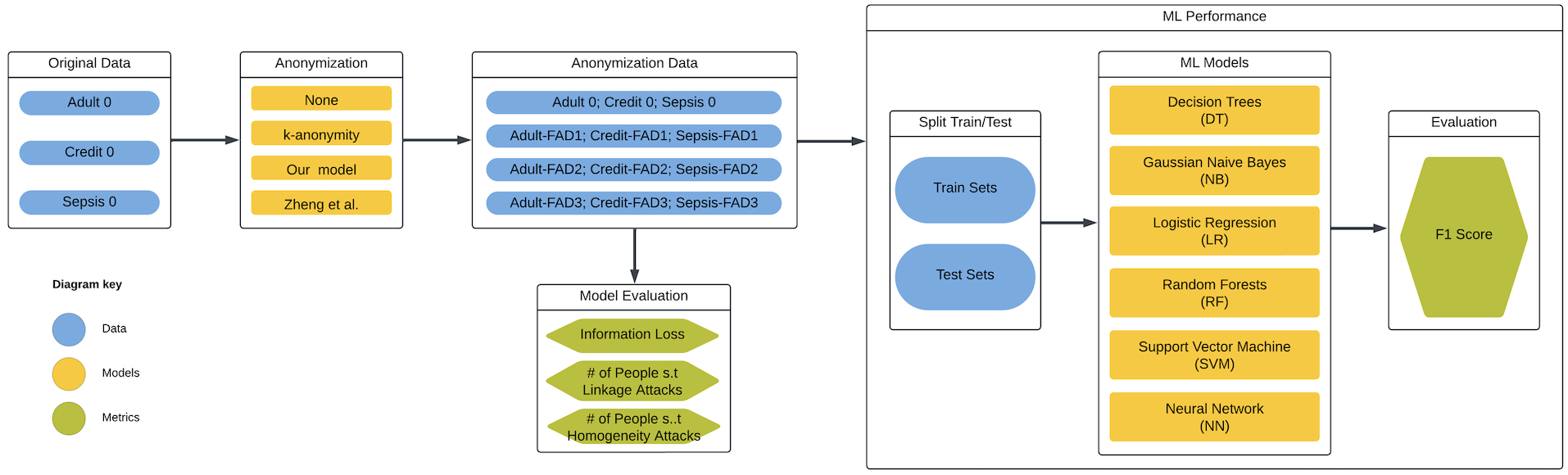}
    \caption{Process of the experiment}
    \label{fig:flow chart}
\end{figure*}

In this section, we delve into the datasets used in our study, outline our strategies for tuning hyperparameters, and describe our experimental design. Figure~\ref{fig:flow chart} delineates the experimental process, illustrating the steps undertaken in our investigation.
\subsection{Data}
In this study, we utilized three distinct datasets: the Adult dataset, the German Credit dataset, and the Sepsis Patient dataset, as summarized in Table~\ref{tab: datasets}.

The German Credit dataset~\cite{statlog_(german_credit_data)_144} is used to classify individuals into categories of good or bad credit risks based on a set of attributes. The QIs encompass age, personal status (including marital status and sex), and job type, while the sensitive attributes include checking account status and saving account status.

The Adult dataset~\cite{adult_2}, also referred to as the Census Income dataset, aims to predict whether an individual's income exceeds \$50,000 per year. It features QIs such as age, race, sex, and marital status, with occupation as the sensitive attribute.

The sepsis patient dataset is composed of retrospectively collected EHR data from two hospitals of a single tertiary-care healthcare system in the United States (in total, 1100 in-hospital beds). The data collection was performed from patients admitted to these hospitals between July 2013 and December 2015. The inclusion criteria consisted of patient age $\geq$ 18 at arrival and visit types of in-patient, Emergency Department only, or observational visits. The QIs in the dataset include age, the number of visits to the hospital, the number of days spent in the hospital, gender, race, and ethnicity. The dataset includes 30 sensitive attributes, which indicate whether a patient has been previously diagnosed with specific diseases such as tumors, hypertension, or blood loss during a prior visit before the current visit. The sepsis flag is the target variable for ML models. The sepsis flag indicates that the patient was discharged with a sepsis-related International Classification of Diseases (ICD) code in their chart based on meeting clinical sepsis criteria during their hospitalization.

\begin{table}
\centering
    \caption{Information about datasets used in this study} 
    \begin{tabular}{m{3em}>{\centering\arraybackslash}m{5em}>{\centering\arraybackslash}m{5em}>{\centering\arraybackslash}m{5em}>{\centering\arraybackslash}m{5em}>{\centering\arraybackslash}m{5em}>{\centering\arraybackslash}m{5em}}\toprule
    Dataset       & Number of Records & Number of Attributes & Number of NQIs & Number of CQIs & Number of SAs & Number of Classes \\\hline
    German credit & 1000              & 21                   & 1              & 2              & 2             & 2   \\\hline
    Adult         & 45222             & 15                   & 1              & 3              & 1             & 2      \\\hline
    Sepsis patient       & 119871            & 106                  & 3              & 3              & 30            & 2     \\\bottomrule
    \end{tabular}   
\label{tab: datasets}
\end{table}

\subsection{Tuning Hyperparameters}
In our model, three key hyperparameters—$n_C$, $\lambda$, and $k$—play crucial roles. The optimal values for these hyperparameters hinge upon several factors including the dataset's size, diversity, sensitivity, and its intended use. We aim to provide guidance on tuning these hyperparameters for effective application of our model.

\begin{enumerate}
    \item $n_C$: This hyperparameter is paramount in our model as it directly influences the diversity of QIs in the anonymized data. It dictates the degree of information loss, resistance against attacks, performance of ML models, and computational efficiency. The lower bound of $n_C$ is 1, so, all data points are aggregated into a single cluster, implying that they share identical QIs combination. Conversely, the upper bound of $n_C$ corresponds to the total number of unique combinations of QIs obtained by concatenating their values from the original dataset. Practically, we may constrain the range of $n_C$ to a subset of this full range, such as starting with 4 clusters or 10 clusters and increasing it by 10 and extending up to 20\% of the upper bound. A smaller $n_C$ leads to more data points being assigned to the same clusters, enhancing robustness against attacks and reducing computational overhead. However, this comes at the cost of increased information loss and potentially diminished ML model performance. 
    
    \item $\lambda$: It controls the trade-off between two competing objective functions. On the one hand, the model aims to minimize IL during the process of anonymization. On the other hand, the model also aims to maximize the protection of sensitive attributes. It ranges from 0 to 1. Our approach initializes $\lambda$ at 0.0001 and iteratively increases it by a factor of 10 until reaching 1. However, for binary sensitive attributes, a higher $\lambda$ is preferable as binary sensitive attributes are more vulnerable to homogeneity attacks. Thus, we initially prioritize defense against such attacks and subsequently adjust $n_C$ based on IL considerations. As $\lambda$ approaches 1, the model prioritizes defense against homogeneity attacks, whereas a value closer to 0 prioritizes information loss minimization. This nuanced adjustment ensures a tailored approach to balancing privacy preservation and utility in the anonymization process.
    \item $k$: This hyperparameter enforces the $k$-anonymity requirement, ensuring that each record is indistinguishable from at least $k-1$ other records. The selection of $k$ should consider factors such as the size of the dataset and the acceptable level of re-identification risk. The minimum value for $k$ is 2, and it must also satisfy the constraint $k \leq \frac{n}{n_C}$. In this study, we adopt $k$ values of 5, 10, 15, and 20 based on El Emam's study \cite{ElEmamKhaled2013CMT}. However, it is essential to acknowledge that in our model, achieving $k$-anonymity becomes more straightforward when the selected value of $k$ is substantially lower than the $\frac{n}{n_C}$.
\end{enumerate}
In this study, we investigate the parameter space of three crucial hyperparameters by establishing intervals for each and assessing the model at different points within these intervals. This approach resembles a grid search, allowing us to assess model performance across a range of hyperparameter values. However, to automate the hyperparameter tuning process, various packages in R or Python, such as rBayesianOptimization \cite{yan2016rbayesianoptimization}, can be utilized. These tools enable efficient exploration of the hyperparameter space, aiding in the selection of optimal values for enhanced model performance.

\subsection{Experimental Design}
\subsubsection{Baseline Analysis}
\begin{itemize}
    \item \textbf{Initial risk level: }To evaluate the baseline privacy risk in the three datasets, we quantify the number of individuals vulnerable to linkage and homogeneity attacks, following the methodology described in Section~\ref{sec: attack identification}. Specifically, we apply risk thresholds of $\tau = 0.05$, $0.075$, and $0.1$, as recommended in El Emam’s study \cite{ElEmamKhaled2013CMT}.
    \item \textbf{Initial ML performance: }We leverage three datasets to train and evaluate ML models, specifically Decision Trees (DT), Gaussian Naive Bayes (NB), Logistic Regression (LR), Random Forests (RF), Support Vector Machine (SVM), and Neural Network (NN). We employ a training and test set division for 100 iterations and document ML performance measured by the F1 score for each iteration.
\end{itemize}

\subsubsection{Anonymization Process}
In the subsequent phase of the experiment, we applied three anonymization algorithms, namely the $k$-anonymity algorithm proposed by Domingo-Ferrer and Torra\cite{domingo2005ordinal}, the algorithm introduced by Zheng et al.\cite{zheng2019effective}, and our model. The $k$-anonymity algorithm proposed by \cite{domingo2005ordinal} exclusively addresses linkage attacks and serves as the baseline algorithm for the anonymization process. Conversely, the algorithm proposed by \cite{zheng2019effective} shares similar objectives as our model, which provide protection against both linkage and homogeneity attacks. Since this paper focuses on establishing a foundational framework that targets these two fundamental privacy attacks, we selected these two existing algorithms for comparison due to their close alignment with the scope of our study.

\subsubsection{Model Evaluations}
 We compare the information loss of the datasets to evaluate the effectiveness of the algorithms in maintaining information. We also calculate the number of individuals subject to linkage and homogeneity attacks in the anonymized datasets to examine the effectiveness of protection against attacks. We will compare our model with the $k$-anonymity algorithm to determine if our model offers superior protection against linkage attacks, if $k$-anonymity alone is insufficient in addressing homogeneity attacks, and if our model can provide advanced protection against them. We will also compare our model with the algorithm introduced by \cite{zheng2019effective}. This comparative analysis will provide insights into the effectiveness of our model relative to state-of-the-art anonymization techniques.

\subsubsection{Machine Learning Performance}
We leverage the anonymized datasets to train and evaluate ML models. The objective is to comprehensively evaluate the influence of our model on the performance of ML models. Therefore, we undertake two comparisons. Firstly, we compare the ML performance of our model with initial ML performance. Secondly, we compare our model's ML performance with two alternative algorithms. These comparisons enable us to thoroughly evaluate the effectiveness of our model in enhancing ML outcomes. To achieve comparisons, we employ the statistical test on the F1 scores gathered from 100 iterations. 

\section{EXPERIMENTAL RESULTS}\label{sec:results}

In this section, we present the model evaluation and ML performance across different datasets and scenarios. Section~\ref{sec:model eval} details the model evaluation results, where Section~\ref{sec:model eval-MOOBAM} discusses the evaluation of our proposed model, MO-OBAM and Section~\ref{sec:model eval-comparison}, we compare model evaluation results of MO-OBAM with two alternative algorithms. Section~\ref{sec:ml perfor} focuses on the ML performance results. Specifically, Section~\ref{sec:ml perfor-MOOBAM} outlines the ML performance of MO-OBAM, while Section~\ref{sec:ml perfor-comparison} provides a comparative analysis of ML performance between MO-OBAM and the alternative algorithms.

The models under consideration vary in the number of hyperparameters they incorporate. Specifically, the algorithm proposed by \cite{domingo2005ordinal} introduces a single hyperparameter, $k$, which is essential for maintaining the $k$-anonymity requirement. In contrast, the algorithm proposed by \cite{zheng2019effective} introduces an additional hyperparameter, $l$, while our model includes two more hyperparameters: $n_C$ and $\lambda$. Despite these differences, all three models share the parameter $k$. Therefore, we focus on presenting results corresponding to different values of $k$ for $k$-anonymity, specifically $k = 5, 10, 15, 20$.

\subsection{Model Evaluations}\label{sec:model eval}

\subsubsection{MO-OBAM}\label{sec:model eval-MOOBAM}

\begin{figure}[]
\centering
\subfloat[Information Loss]{
\includegraphics[scale=0.1]{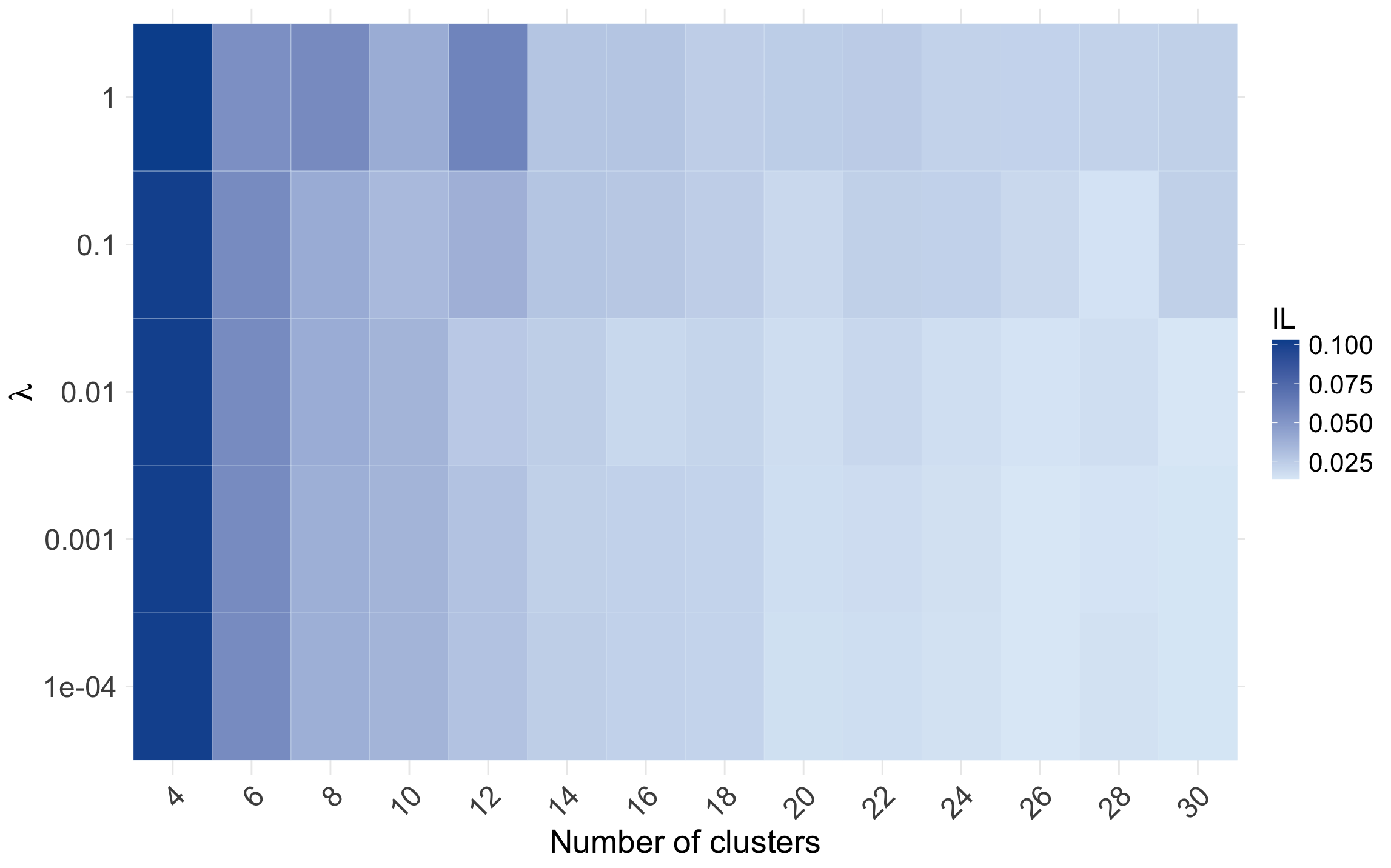}
\label{fig: German credit-infoloss-5}}\hfill
\subfloat[Number of People s.t Linkage Attacks]{
\includegraphics[scale=0.1]{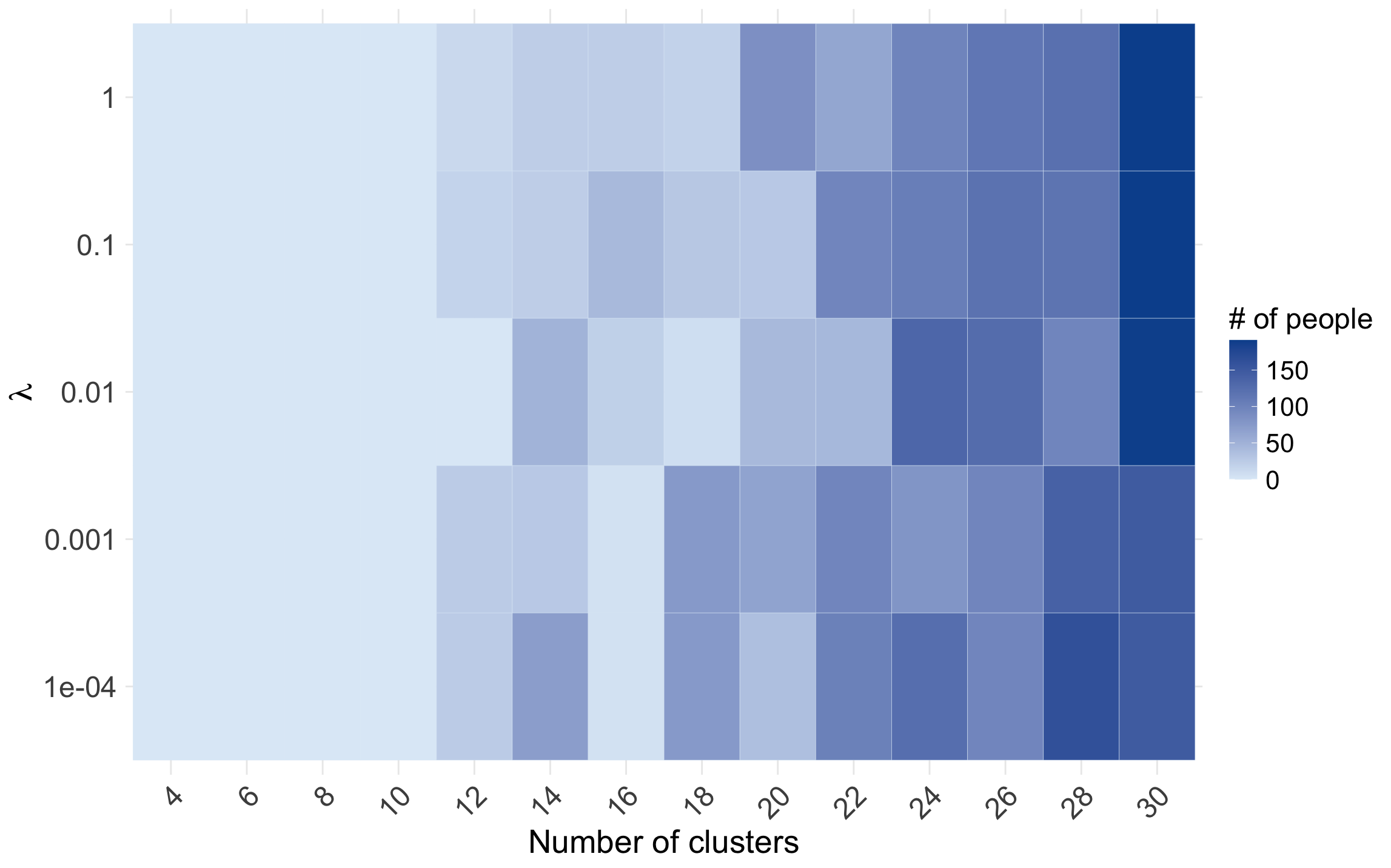}
\label{fig: German credit-LA-5}}\hfill
\subfloat[Number of People s.t Homogeneity Attacks]{
\includegraphics[scale=0.1]{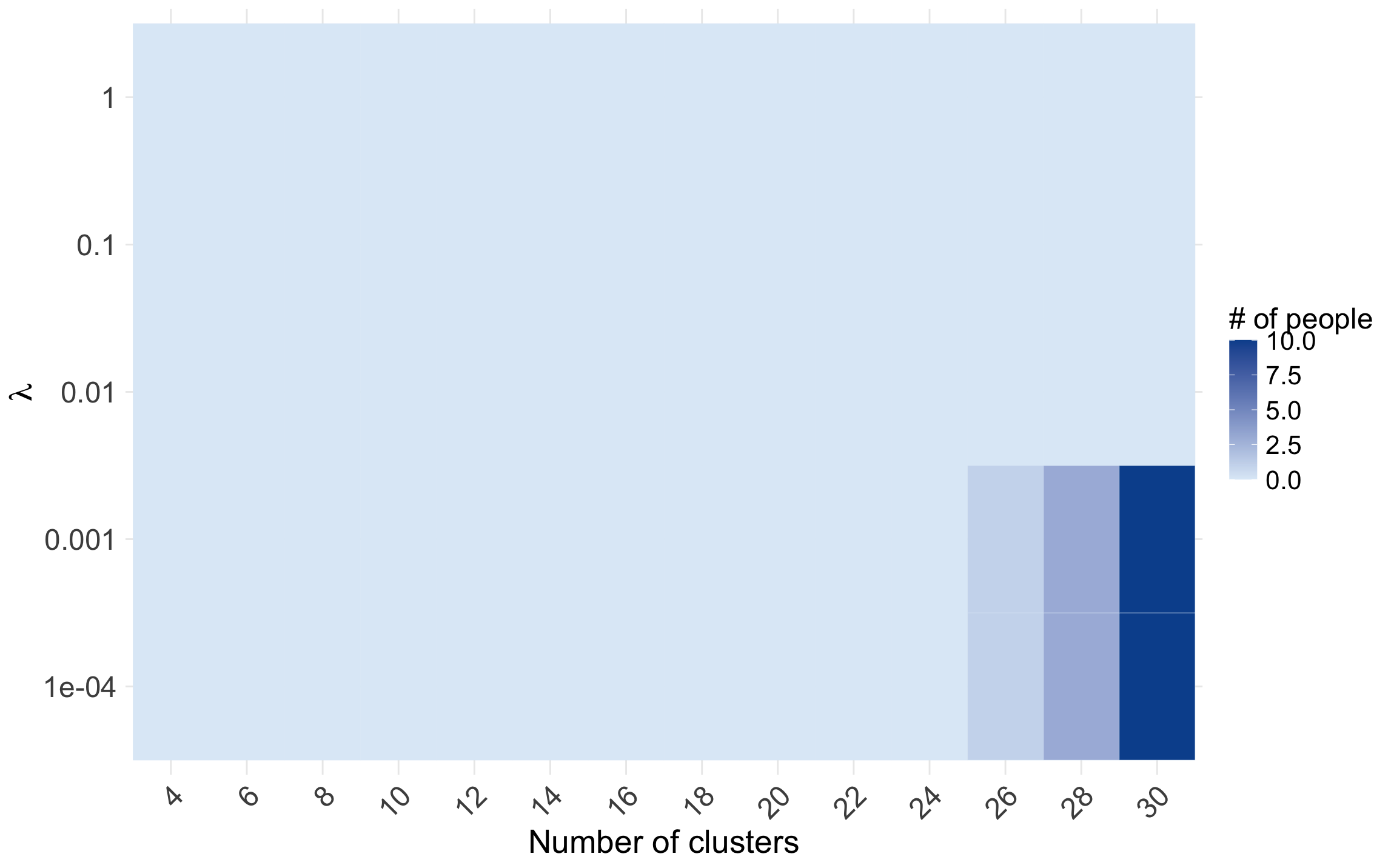}
\label{fig: German credit-HA-5}}
\caption{Impact of $n_C$ and $\lambda$ on privacy preservation using the German credit German credit dataset $(k=5)$}
\label{fig: German credit-5}
\end{figure}

In this section, we use the German credit dataset as an example to demonstrate the impact of hyperparameter changes in our model on information loss, and the number of individuals susceptible to linkage and homogeneity attacks. Figure~\ref{fig: German credit-5} shows how each hyperparameter change affects these metrics when $k=5$. The x-axis represents $n_C$ (number of clusters), ranging from 4 to 30, while the y-axis represents $\lambda$, which varies exponentially from 1 to 0.0001. The color gradient indicates the level of information loss, the number of individuals at risk of linkage or homogeneity attacks, with darker blue areas representing higher values and lighter blue areas representing lower values.

Figure~\ref{fig: German credit-infoloss-5} illustrates how information loss varies with different combinations of $n_C$ and $\lambda$ values for $k=5$. It is evident that as $n_C$ increases while holding $\lambda$ constant, there is a consistent decrease in information loss. Conversely, when $n_C$ is fixed, increasing $\lambda$ results in higher information loss. This trend holds consistent across different values of $k$.

Figure~\ref{fig: German credit-LA-5} shows how the number of individuals susceptible to linkage attacks varies with different combinations of $n_C$ and $\lambda$ values when $k=5$ and $\tau=0.05$. This figure indicates that as $n_C$ increases while holding $\lambda$ constant, there is a consistent increase in the number of individuals at risk of linkage attacks. Similarly, increasing $\lambda$ while holding $n_C$ constant also increases the number of individuals at risk. Interestingly, when $n_C$ ranges from 4 to 10, no individuals are at risk of linkage attacks, demonstrating that our model provides sufficient protection against such attacks with a smaller number of clusters. This occurs because fewer clusters lead to more individuals per cluster, mitigating the risk of linkage attacks.

Figure~\ref{fig: German credit-HA-5} depicts how the number of individuals susceptible to homogeneity attacks varies with different combinations of $n_C$ and $\lambda$ values when $k=5$. The figure reveals that when $\lambda$ is small, prioritizing the minimization of the objective function over information loss, certain individuals remain vulnerable to homogeneity attacks, especially with larger $n_C$ values. In addition, as $n_C$ increases while holding $\lambda$ constant, the number of individuals at risk of homogeneity attacks also increases. However, the large white area in the figure indicates that no individuals are at risk of homogeneity attacks in most combinations of $n_C$ and $\lambda$ values, underscoring the effectiveness of our model in mitigating such attacks.

These results demonstrate that information loss, and the number of individuals susceptible to linkage and homogeneity attacks are influenced by the number of clusters. Observing the three plots vertically, the darker blue areas in Figure~\ref{fig: German credit-5} are inversely related. An increase in the number of clusters typically reduces information loss but increases the risk of linkage and homogeneity attacks. This is because a greater number of clusters leads to a wider diversity of QI values, thereby reducing information loss. However, as the number of clusters grows, fewer individuals are allocated to each cluster, increasing the risk of attacks. Therefore, to achieve robust protection against attacks while maintaining data utility, an optimal range for $n_C$ is generally in the middle area.

\subsubsection{Comparative Analysis of Model Evaluation}\label{sec:model eval-comparison}

To comprehensively evaluate the models, we systematically explore various values for each hyperparameter. In Appendix A, we provide a detailed overview of the selected hyperparameter values for each model. As each combination of hyperparameters results in specific levels of information loss and varying susceptibility to linkage and homogeneity attacks, due to the space constraints, presenting all possible values is impractical. Consequently, we present the results primarily based on varying values of $k$ ($k=5,10,15,20$) and values for $l$, $n_C$, and $\lambda$ that promote lower and higher protection against homogeneity attacks for \cite{zheng2019effective} and our model. Table~\ref{tab: German credit model evaluations} present model evaluation results using German credit dataset, and Table~\ref{tab: Adult model evaluations} display model evaluation results using Adult dataset. For the sepsis patient data, all sensitive attributes are binary. In \cite{zheng2019effective}, they have identified the value of 2 as optimal for maximizing protection against homogeneity attacks. Following this principle, we only compare our results when we promote higher protection against homogeneity attacks with theirs. Hence, Table~\ref{tab: sepsis model evaluations2} present evaluations with higher promotion of protection against homogeneity attacks. 

\subsubsection*{Compare to $k$-anonymity} 
In the algorithm proposed by \cite{domingo2005ordinal}, the hyperparameter $k$ dictates the number of clusters, with higher values of $k$ resulting in a decreased number of clusters. The results of $k$-anonymity presented in Table~\ref{tab: German credit model evaluations} through~\ref{tab: sepsis model evaluations2} illustrate that as $k$ increases from 5 to 20, the number of clusters decreases while information loss increases. Concurrently, the number of individuals vulnerable to attacks decreases. However, even when $k=5$, there remain individuals susceptible to homogeneity attacks across all three datasets. Notably, in the sepsis patient dataset, which comprises the most significant number of sensitive attributes, individuals are still at risk of homogeneity attacks even when $k=20$.

Upon comparing our proposed model with the $k$-anonymity algorithm, several critical insights emerge for each dataset. 
\begin{itemize}
    \item \textbf{German credit:} When a lower promotion of protection against homogeneity attacks is proposed, our model consistently demonstrates lower information loss across all values of $k$ despite having fewer clusters. This indicates that our approach preserves data utility more effectively. In terms of the number of individuals susceptible to linkage attacks, our model shows significant improvements. For $\tau=0.05$, our model results in substantially fewer individuals at risk of linkage attacks in 3 out of 4 different $k$ values compared to $k$-anonymity. For $\tau=0.075$, our model continues to demonstrate fewer individuals at risk in 2 out of 4 different $k$ values. However, for $\tau=0.1$, our model achieves fewer individuals at risk in only 1 out of 4 $k$ values. Regarding homogeneity attacks, our model performs better at $k=5$, showing fewer individuals at risk compared to $k$-anonymity. However, for $k$ values of 10, 15, and 20, our model shows an increase in the number of individuals susceptible to homogeneity attacks compared to $k$-anonymity. When emphasizing higher protection against homogeneity attacks, our model exhibits higher information loss across all values of $k$ due to the significantly fewer clusters compared to $k$-anonymity. Because of fewer clusters, our model provides more robust protection against both linkage and homogeneity attacks. The significantly fewer clusters in our model lead to a scenario where no individuals are susceptible to linkage or homogeneity attacks, highlighting the effectiveness of our approach in safeguarding sensitive data.
    \item \textbf{Adult:} When a lower promotion of protection against homogeneity attacks is required, our model displays higher information loss across all values of $k$ due to the fewer clusters compared to $k$-anonymity. For protection against linkage attacks, our model consistently shows fewer individuals at risk. For $\tau=0.05$, our model results in significantly fewer individuals susceptible to linkage attacks in 3 out of 4 different $k$ values compared to $k$-anonymity. For $\tau=0.075$, our model continues to demonstrate fewer individuals at risk in 2 out of 4 different $k$ values. For $\tau=0.1$, our model achieves fewer individuals at risk in only 1 out of 4 $k$ values. Regarding the protection against homogeneity attacks, our model performs better at $k=5$, showing fewer individuals at risk compared to $k$-anonymity. However, for $k$ values of 10, 15, and 20, our model shows an increase in the number of individuals susceptible to homogeneity attacks compared to $k$-anonymity. When emphasizing higher protection against homogeneity attacks, our model exhibits higher information loss across all values of $k$ due to the significantly fewer clusters compared to $k$-anonymity. With fewer clusters, our model provides more robust protection against both types of attacks.
    \item \textbf{Sepsis patient:} In the Sepsis Patient dataset, our model consistently demonstrates lower information loss across all values of $k$ despite having fewer clusters. Regarding linkage attacks, our model performs equally or better, showing fewer individuals at risk except for $k=15$ when $\tau=0.05$ and $\tau=0.1$. For homogeneity attacks, our model shows no individuals at risk across all $k$ values, unlike $k$-anonymity, which consistently leaves some individuals vulnerable.
\end{itemize}

Overall, when $k$-anonymity shows an 8\% to 35\% decrease in the number of individuals at risk of linkage attacks compared to the baseline, our model achieves approximately a 96\% to 98\% decrease, which indicates our model's superior ability to protect against linkage attacks. Moreover, our model provides advanced protection against homogeneity attacks, significantly reducing the number of individuals at risk.

\subsubsection*{Compare to algorithm proposed by\cite{zheng2019effective}}
In the algorithm proposed by \cite{zheng2019effective}, both $k$ and $l$ play a role in determining the number of clusters. Specifically, for a fixed value of $l$, increasing $k$ results in fewer clusters. The results of \cite{zheng2019effective} in each table from~\ref{tab: German credit model evaluations} to~\ref{tab: sepsis model evaluations2} can illustrate this trend, showing that as $k$ increases, information loss also increases, while the number of individuals susceptible to attacks, particularly linkage attacks, decreases. Conversely, for a given value of $k$, an increase in $l$ results in a decrease in the number of clusters, so the information loss is increased, but the number of individuals vulnerable to linkage attacks decreases. 

When comparing our model to the algorithm proposed by \cite{zheng2019effective}, several key observations emerge.
\begin{itemize}
    \item \textbf{German credit:} In scenarios emphasizing lower promotion of protection against homogeneity attacks, our model exhibits higher information loss for $k=5,10$, and $15$ due to fewer clusters compared to the algorithm proposed by \cite{zheng2019effective}. However, for $k=20$, our model achieves lower information loss despite having fewer clusters. Regarding protection against linkage attacks, our model significantly outperforms the algorithm proposed by Zheng et al. in several instances. For $\tau=0.05$, our model results in significantly fewer individuals susceptible to linkage attacks in 3 out of 4 different $k$ values. For $\tau=0.075$, this superior performance is observed in 2 out of 4 $k$ values, and for $\tau=0.1$, it is seen in 1 out of 4 $k$ values. However, when evaluating the number of individuals susceptible to homogeneity attacks, the algorithm by \cite{zheng2019effective} demonstrates superior performance. In situations where higher promotion of protection against homogeneity attacks is prioritized, our model incurs higher information loss across all $k$ values. This is due to the significantly fewer clusters used in our approach. Despite this increased information loss, our model exhibits superior performance in terms of protection against both linkage and homogeneity attacks compared to the algorithm proposed by \cite{zheng2019effective}
    \item \textbf{Adult:} In scenarios emphasizing lower promotion of protection against homogeneity attacks, the algorithm proposed by \cite{zheng2019effective} achieves significantly fewer clusters, resulting in markedly lower information loss compared to our model. Despite this, our model offers more robust protection against linkage attacks in various scenarios. Specifically, our model demonstrates superior performance for $\tau=0.05$ with $k=5, 10, 15$, for $\tau=0.075$ with $k=5, 10$, and for $\tau=0.1$ with $k=5$. On the number of individuals susceptible to homogeneity attacks, \cite{zheng2019effective} outperforms our model. This trend is consistent with observations from the German credit dataset. In scenarios emphasizing higher promotion of protection against homogeneity attacks, our model achieves comparable performance in terms of protection against both linkage and homogeneity attacks when compared to the algorithm proposed by \cite{zheng2019effective}. However, our model manages to achieve lower information loss.
    \item \textbf{Sepsis patient:} In this dataset, our model achieves comparable performance in terms of protection against both linkage and homogeneity attacks when compared to the algorithm proposed by \cite{zheng2019effective}. However, our model manages to achieve significantly lower information loss.
\end{itemize}

In conclusion, our model consistently offers superior protection against linkage attacks and comparable protection against homogeneity attacks compared to the algorithm proposed by \cite{zheng2019effective}. While the algorithm by \cite{zheng2019effective} achieves lower information loss in scenarios with lower promotion of homogeneity attack protection, our model provides a more balanced approach, excelling in privacy protection and maintaining lower information loss in scenarios emphasizing higher promotion of homogeneity attack protection.

\begin{table}
    \centering
    \caption{Comparison of model results for the German Credit dataset. The columns labeled $\tau=0.05$, $\tau=0.075$, and $\tau=0.1$ indicate the number of individuals at risk of linkage attacks. The column labeled HA represents the number of individuals at risk of homogeneity attacks. The ``Baseline'' row refers to the original German Credit dataset.}
    \begin{tabular}{c}
     (a) Hyperparameter values that promote lower protection against homogeneity attacks \\    
    \resizebox{\textwidth}{!}{\begin{tabular}{lclccccc}\toprule
Model       & \# of clusters  & Hyperparameter Values        & IL     & $\tau$=0.05 & $\tau$=0.075 & $\tau$=0.1 & HA  \\ \hline
Baseline    & 310 &                              &        & 959    & 828     & 698   & 275 \\ \hline
$k$-anonymity & 149 & ($k$=5)                        & 0.0166 & 940    & 790     & 570   & 50  \\
Zheng et al & 191 & ($k$=5, $l$=2)                   & 0.0061 & 899    & 778     & 532   & 0   \\
MO-OBAM   & 30  & ($k$=5, $\lambda$=0.0001, $n_C$=30)  & 0.0147 & 148    & 63      & 29    & 10  \\ \hline
$k$-anonymity & 90  & ($k$=10)                       & 0.0255 & 810    & 810     & 0     & 0   \\
Zheng et al & 101 & ($k$=10, $l$=2)                  & 0.0106 & 860    & 860     & 0     & 0   \\
MO-OBAM   & 30  & ($k$=10, $\lambda$=0.0001, $n_C$=30) & 0.0153 & 118    & 71      & 22    & 5   \\ \hline
$k$-anonymity & 63  & ($k$=15)                       & 0.0408 & 885    & 0       & 0     & 0   \\
Zheng et al & 67  & ($k$=15, $l$=2)                  & 0.0141 & 820    & 10      & 0     & 0   \\
MO-OBAM   & 30  & ($k$=15, $\lambda$=0.0001, $n_C$=30) & 0.0153 & 106    & 40      & 40    & 3   \\ \hline
$k$-anonymity & 49  & ($k$=20)                       & 0.0545 & 0      & 0       & 0     & 0   \\
Zheng et al & 51  & ($k$=20, $l$=2)                  & 0.0164 & 0      & 0       & 0     & 0   \\
MO-OBAM   & 30  & ($k$=20, $\lambda$=0.0001, $n_C$=30) & 0.0148 & 125    & 57      & 31    & 2   \\ \bottomrule
\end{tabular}}  \\ \\
    (b) Hyperparameter values that promote higher protection against homogeneity attacks \\
    \resizebox{\textwidth}{!}{\begin{tabular}{lclccccc}\toprule
Model         & \# of clusters & Hyperparameter Values          & IL     & $\tau$=0.05 & $\tau$=0.075 & $\tau$=0.1 & HA  \\ \hline
Baseline      & 310   &                                &        & 959         & 828          & 698        & 275 \\ \hline
$k$-anonymity & 149   & ($k$=5)                        & 0.0166 & 940         & 790          & 570        & 50  \\
Zheng et al   & 46    & ($k$=5, $l$=4)                 & 0.0209 & 268         & 109          & 16         & 0   \\
MO-OBAM     & 4     & ($k$=5, $\lambda$=1, $n_C$=4)  & 0.1027 & 0           & 0            & 0          & 0   \\ \hline
$k$-anonymity & 90    & ($k$=10)                       & 0.0255 & 810         & 810          & 0          & 0   \\
Zheng et al   & 47    & ($k$=10, $l$=4)                & 0.0190 & 288         & 143          & 0          & 0   \\
MO-OBAM     & 4     & ($k$=10, $\lambda$=1, $n_C$=4) & 0.1027 & 0           & 0            & 0          & 0   \\ \hline
$k$-anonymity & 63    & ($k$=15)                       & 0.0408 & 885         & 0            & 0          & 0   \\
Zheng et al   & 41    & ($k$=15, $l$=4)                & 0.0190 & 339         & 12           & 0          & 0   \\
MO-OBAM     & 4     & ($k$=15, $\lambda$=1, $n_C$=4) & 0.1027 & 0           & 0            & 0          & 0   \\ \hline
$k$-anonymity & 49    & ($k$=20)                       & 0.0545 & 0           & 0            & 0          & 0   \\
Zheng et al   & 40    & ($k$=20, $l$=4)                & 0.0195 & 10          & 10           & 0          & 0   \\
MO-OBAM     & 4     & ($k$=20, $\lambda$=1, $n_C$=4) & 0.1027 & 0           & 0            & 0          & 0   \\ \bottomrule
\end{tabular}} 
    \end{tabular}
    \label{tab: German credit model evaluations} 
\end{table}

\begin{table}
\caption{Comparison of model results for the Adult dataset. The columns labeled $\tau=0.05$, $\tau=0.075$, and $\tau=0.1$ indicate the number of individuals at risk of linkage attacks. The column labeled HA represents the number of individuals at risk of homogeneity attacks. The ``Baseline'' row refers to the original Adult dataset.}
\begin{tabular}{c}
(a) Hyperparameter values that promote lower protection against homogeneity attacks \\
\resizebox{\textwidth}{!}{\begin{tabular}{lclccccc}\toprule
Model         & \# of clusters & Hyperparameter Values                 & IL     & $\tau$=0.05 & $\tau$=0.075 & $\tau$=0.1 & HA  \\ \hline
Baseline      & 1900  &                                       &        & 6506        & 4906         & 3910       & 634 \\ \hline
$k$-anonymity & 1232  & ($k$=5)                               & 0.0010 & 5925        & 4590         & 3050       & 5   \\
Zheng et al   & 1066  & ($k$=5, $l$=2)                        & 0.0006 & 5077        & 3680         & 2280       & 0   \\
MO-OBAM     & 100   & ($k$=5, $\lambda$=0.0001, $n_C$=100)  & 0.0116 & 103         & 69           & 47         & 3   \\ \hline
$k$-anonymity & 930   & ($k$=10)                              & 0.0022 & 5080        & 5080         & 0          & 0   \\
Zheng et al   & 895   & ($k$=10, $l$=2)                       & 0.0013 & 4770        & 4770         & 0          & 0   \\
MO-OBAM     & 100   & ($k$=10, $\lambda$=0.0001, $n_C$=100) & 0.0116 & 103         & 69           & 47         & 3   \\ \hline
$k$-anonymity & 774   & ($k$=15)                              & 0.0034 & 4167        & 0            & 0          & 0   \\
Zheng et al   & 779   & ($k$=15, $l$=2)                       & 0.0019 & 4084        & 0            & 0          & 0   \\
MO-OBAM     & 100   & ($k$=15, $\lambda$=0.0001, $n_C$=100) & 0.0116 & 103         & 69           & 47         & 3   \\ \hline
$k$-anonymity & 673   & ($k$=20)                              & 0.0045 & 0           & 0            & 0          & 0   \\
Zheng et al   & 733   & ($k$=20, $l$=2)                       & 0.0025 & 0           & 0            & 0          & 0   \\
MO-OBAM     & 100   & ($k$=20, $\lambda$=0.0001, $n_C$=100) & 0.0116 & 103         & 69           & 47         & 3   \\ \bottomrule
\end{tabular}} \\ \\ 
(b) Hyperparameter values that promote higher protection against homogeneity attacks \\
\resizebox{\textwidth}{!}{\begin{tabular}{lclccccc}\toprule
Model         & \# of clusters & Hyperparameter Values          & IL     & $\tau$=0.05 & $\tau$=0.075 & $\tau$=0.1 & HA  \\ \hline
Baseline      & 1900  &                                &        & 6506        & 4906         & 3910       & 634 \\ \hline
$k$-anonymity & 1232  & ($k$=5)                        & 0.0010 & 5925        & 4590         & 3050       & 5   \\
Zheng et al   & 12    & ($k$=5, $l$=14)                & 0.1189 & 0           & 0            & 0          & 0   \\
MO-OBAM     & 4     & ($k$=5, $\lambda$=1, $n_C$=4)  & 0.1074 & 0           & 0            & 0          & 0   \\ \hline
$k$-anonymity & 930   & ($k$=10)                       & 0.0022 & 5080        & 5080         & 0          & 0   \\
Zheng et al   & 12    & ($k$=10, $l$=14)               & 0.1251 & 0           & 0            & 0          & 0   \\
MO-OBAM     & 4     & ($k$=10, $\lambda$=1, $n_C$=4) & 0.1074 & 0           & 0            & 0          & 0   \\ \hline
$k$-anonymity & 774   & ($k$=15)                       & 0.0034 & 4167        & 0            & 0          & 0   \\
Zheng et al   & 14    & ($k$=15, $l$=14)               & 0.1201 & 0           & 0            & 0          & 0   \\
MO-OBAM     & 4     & ($k$=15, $\lambda$=1, $n_C$=4) & 0.1074 & 0           & 0            & 0          & 0   \\ \hline
$k$-anonymity & 673   & ($k$=20)                       & 0.0045 & 0           & 0            & 0          & 0   \\
Zheng et al   & 12    & ($k$=20, $l$=14)               & 0.1251 & 0           & 0            & 0          & 0   \\
MO-OBAM     & 4     & ($k$=20, $\lambda$=1, $n_C$=4) & 0.1074 & 0           & 0            & 0          & 0   \\ \bottomrule
\end{tabular}}
\end{tabular}
\label{tab: Adult model evaluations}    
\end{table}

\begin{table}
\caption{Comparison model results with promoting higher protection against homogeneity attacks for the original Sepsis Patient dataset. The columns labeled $\tau=0.05$, $\tau=0.075$, and $\tau=0.1$ indicate the number of individuals at risk of linkage attacks. The column labeled HA represents the number of individuals at risk of homogeneity attacks. The ``Baseline'' row refers to the original Sepsis Patient dataset.}
\begin{tabular}{c}
     \resizebox{\textwidth}{!}{\begin{tabular}{lclccccc}\toprule
Model         & \# of clusters & Hyperparameter Values                 & IL     & $\tau=0.05$ & $\tau=0.075$ & $\tau=0.1$ & HA     \\ \hline
Baseline      & 23553 &                                       &        & 56113       & 48396        & 41445      & 14937 \\ \hline
$k$-anonymity & 9999  & ($k$=5)                               & 0.0050 & 53225       & 46250        & 35480      & 6905   \\
Zheng et al   & 4471  & ($k$=5, $l$=2)                        & 0.0114 & 18174       & 11656        & 7486       & 0      \\
MO-OBAM     & 3240  & ($k$=5, $\lambda$=1, $n_C$=3240)      & 0.0032 & 15680       & 7375         & 2954       & 0      \\ \hline
$k$-anonymity & 6420  & ($k$=10)                              & 0.0089 & 47990       & 47990        & 0          & 2260   \\
Zheng et al   & 3206  & ($k$=10, $l$=2)                       & 0.0122 & 17068       & 12929        & 0          & 0      \\
MO-OBAM     & 2310  & ($k$=10, $\lambda$=1, $n_C$=2310)     & 0.0037 & 6939        & 1960         & 0          & 0      \\ \hline
$k$-anonymity & 4883  & ($k$=15)                              & 0.0122 & 46355       & 0            & 0          & 780    \\
Zheng et al   & 3085  & ($k$=15, $l$=2)                       & 0.0125 & 21607       & 0           & 0          & 0      \\
MO-OBAM     & 1740  & ($k$=15, $\lambda$=1, $n_C$=1740)     & 0.0046 & 2020        & 0           & 0         & 0      \\ \hline
$k$-anonymity & 4003  & ($k$=20)                              & 0.0146 & 0           & 0            & 0          & 320    \\
Zheng et al   & 2437  & ($k$=20, $l$=2)                       & 0.0137 & 0           & 0            & 0          & 0      \\
MO-OBAM     & 820   & ($k$=20, $\lambda$=1, $n_C$=820) & 0.0066 & 0           & 0            & 0          & 0      \\ \bottomrule
\end{tabular}}
\end{tabular}
\label{tab: sepsis model evaluations2}
\end{table}

\subsection{Machine Learning Performance}\label{sec:ml perfor}
In this section, we explore the impact of anonymization models on the performance of ML models. We employ six distinct ML algorithms, namely Decision Trees (DT), Logistic Regression (LR), Gaussian Naive Bayes (NB), Random Forests (RF), Neural Networks (NN), and Support Vector Machines (SVM), and evaluate their performance using the F1 score. To evaluate differences in ML performance between utilizing original datasets (referred to as "Baseline" henceforth) and anonymized datasets, we conducted a comparative analysis. Specifically, we examine changes in feature importance for the Decision Trees model as an illustrative example, as demonstrated in Appendix C. Moreover, to statistically validate any observed disparities in performance, we employed the Mann-Whitney U test on F1 scores. This statistical approach was selected due to potential deviations from normality in the distribution of the F1 scores.

\subsubsection{MO-OBAM}\label{sec:ml perfor-MOOBAM}
Before discussing F1 scores, it is important to address feature importance and how the values of hyperparameters in our model impact feature importance. We have assessed feature importance from the original datasets and observed it under two scenarios: one that promotes higher protection against homogeneity attacks and another that promotes lower protection.
In the scenario promoting higher protection against homogeneity attacks, our model generates fewer clusters and reduces the diversity of QI values. Consequently, feature importance of QIs significantly deviates from the baseline, generally diminishing their importance. This trend is particularly noticeable for the most important QI among QIs. Conversely, when promoting lower protection against homogeneity attacks, our model leads to an increase in the number of clusters. In this scenario, the importance levels of QIs may decrease but approach the baseline. This pattern is consistently observed across all datasets. Detailed information on feature importance is presented in Appendix C.

Given how our model influences the importance of QIs, it is essential to explore F1 scores for each ML model.
\begin{itemize}
    \item \textbf{German credit:} Table \ref{tab: datasets ML with p-value}(a) presents the F1 scores corresponding to ML performance using both the original German credit dataset and the dataset anonymized by our model at different levels. Across DT, NB, NN, RF, and SVM, negligible variances in F1 scores are observed compared to the baseline. Only LR exhibits decreased F1 scores for certain hyperparameter configurations relative to the baseline. This implies that our model maintains ML performance across most models tested under the listed two conditions for the German credit dataset.
    \item \textbf{Adult:} Table~\ref{tab: datasets ML with p-value}(b) presents a comprehensive view of the performance of ML models on the Adult dataset under varying levels of anonymization generated by our model. We observe statistically significant differences between our model and the baseline in DT, LR, and RF. Specifically, when hyperparameter values are chosen to promote higher defense against homogeneity attacks, DT and LR display decreases in F1 scores relative to the baseline. Additionally, our RF consistently exhibits lower F1 scores compared to the baseline across all scenarios. Moreover, within Table~\ref{tab: datasets ML with p-value}(b), it is observed that F1 scores in the block of $n_C=4,\lambda=1$ are lower than those in the block of $n_C=100,\lambda=0.0001$ for some ML models. This underscores the potential influence of anonymization levels on ML model performance in the Adult dataset.
    \item \textbf{Sepsis Patient:} As previously mentioned, the original Sepsis patient dataset is highly imbalanced, with only 3\% of patients diagnosed with sepsis. To address this imbalance, we applied PSM to reduce the ratio. Table~\ref{tab: datasets ML with p-value}(c) illustrates the F1 scores obtained using the PSM-adjusted Sepsis patient dataset. Upon comparing F1 scores between our model and the baseline, it is noteworthy that three out of six ML models—DT, NN, and RF—exhibit lower F1 scores than the baseline, and none of them occur when $n_C=3240$, which is the largest number of clusters we selected. This observation suggests that a sufficient number of clusters may mitigate the impact of anonymization on ML performance. 
\end{itemize}

Our model has the capability to maintain ML performance. In scenarios where higher protection against homogeneity attacks is prioritized, our model tends to reduce feature importance for QIs but the F1 score analysis indicates that while there are some performance trade-offs in certain ML models and datasets, the overall impact on ML performance is manageable. Additionally, in scenarios emphasizing lower protection, our model retains higher feature importance levels closer to the baseline, so our model maintains adequate ML performance across various models. We can mitigate the impact of our model on ML performance by adjusting the number of clusters. This adaptability makes our model a robust choice for applications requiring a balance between privacy and predictive accuracy.

\begin{table}
\caption{Comparison of average F1 scores in the three datasets between our model with selected hyperparameter values and the baseline. Color-coded cells indicate that the p-value of the Mann-Whitney U test, compared to the baseline, is less than 0.05.}
\begin{tabular}{c}
(a) German Credit Dataset \\
\begin{tabular}{lccccccccc}
\toprule
Model &
  $n_C$ &
  $\lambda$ &
  $k$ &
  DT &
  LR &
  NB &
  NN &
  RF &
  SVM \\ \hline
Baseline &
  310 &
   &
   &
  0.7702 &
  0.8124 &
  0.7938 &
  0.7109 &
  0.8381 &
  0.8266 \\ \hline
\multirow{8}{*}{MO-OBAM} &
  \multirow{4}{*}{30} &
  \multirow{4}{*}{0.0001} &
  5 &
  0.7708 &
  0.8079 &
  0.7901 &
  0.7210 &
  0.8396 &
  0.8273 \\ \cmidrule(lr){4-10} 
 &
   &
   &
  10 &
  0.7683 &
  \cellcolor[HTML]{F1A983}0.8049 &
  0.7928 &
  0.7208 &
  0.8400 &
  0.8244 \\ \cmidrule(lr){4-10} 
 &
   &
   &
  15 &
  0.7690 &
  \cellcolor[HTML]{F1A983}0.8038 &
  0.7960 &
  0.7156 &
  0.8341 &
  0.8245 \\ \cmidrule(lr){4-10} 
 &
   &
   &
  20 &
  0.7699 &
  0.8073 &
  0.7939 &
  0.7239 &
  0.8373 &
  0.8265 \\ \cmidrule(lr){2-10} 
 &
  \multirow{4}{*}{4} &
  \multirow{4}{*}{1} &
  5 &
  0.7701 &
  0.8123 &
  0.7910 &
  0.7153 &
  0.8382 &
  0.8239 \\ \cmidrule(lr){4-10} 
 &
   &
   &
  10 &
  0.7714 &
  0.8077 &
  0.7863 &
  0.7148 &
  0.8398 &
  0.8251 \\ \cmidrule(lr){4-10} 
 &
   &
   &
  15 &
  0.7698 &
  0.8090 &
  0.7923 &
  0.7218 &
  0.8387 &
  0.8241 \\ \cmidrule(lr){4-10} 
 &
   &
   &
  20 &
  0.7696 &
  \cellcolor[HTML]{F1A983}0.8057 &
  0.7878 &
  0.7189 &
  0.8351 &
  0.8264 \\ \bottomrule
\end{tabular} \\ \\
(b) Adult Dataset \\
\begin{tabular}{lccccccccc}
\toprule
Model &
  $n_C$ &
  $\lambda$ &
  $k$ &
  DT &
  LR &
  NB &
  NN &
  RF &
  SVM \\ \hline
Baseline &
  1900 &
   &
   &
  0.6203 &
  0.4052 &
  0.4183 &
  0.3475 &
  0.6787 &
  0.2757 \\ \hline
\multirow{8}{*}{MO-OBAM} &
  \multirow{4}{*}{100} &
  \multirow{4}{*}{0.0001} &
  5 &
  0.6202 &
  0.4077 &
  0.4192 &
  0.3481 &
  \cellcolor[HTML]{F1A983}0.6684 &
  0.2729 \\ \cmidrule(lr){4-10} 
 &
   &
   &
  10 &
  0.6203 &
  0.4076 &
  0.4190 &
  0.3495 &
  \cellcolor[HTML]{F1A983}0.6695 &
  0.2746 \\ \cmidrule(lr){4-10} 
 &
   &
   &
  15 &
  0.6203 &
  0.4052 &
  0.4190 &
  0.3477 &
  \cellcolor[HTML]{F1A983}0.6674 &
  0.2741 \\ \cmidrule(lr){4-10} 
 &
   &
   &
  20 &
  0.6203 &
  0.4073 &
  0.4197 &
  0.3478 &
  \cellcolor[HTML]{F1A983}0.6688 &
  0.2746 \\ \cmidrule(lr){2-10} 
 &
  \multirow{4}{*}{4} &
  \multirow{4}{*}{1} &
  5 &
  \cellcolor[HTML]{F1A983}0.6186 &
  \cellcolor[HTML]{F1A983}0.4014 &
  0.4193 &
  0.3566 &
  \cellcolor[HTML]{F1A983}0.6600 &
  0.2727 \\ \cmidrule(lr){4-10} 
 &
   &
   &
  10 &
  0.6194 &
  \cellcolor[HTML]{F1A983}0.4022 &
  0.4189 &
  0.3492 &
  \cellcolor[HTML]{F1A983}0.6607 &
  0.2743 \\ \cmidrule(lr){4-10} 
 &
   &
   &
  15 &
  \cellcolor[HTML]{F1A983}0.6180 &
  \cellcolor[HTML]{F1A983}0.4011 &
  0.4196 &
  0.3530 &
  \cellcolor[HTML]{F1A983}0.6608 &
  0.2741 \\ \cmidrule(lr){4-10} 
 &
   &
   &
  20 &
  \cellcolor[HTML]{F1A983}0.6183 &
  \cellcolor[HTML]{F1A983}0.4002 &
  0.4182 &
  0.3504 &
  \cellcolor[HTML]{F1A983}0.6603 &
  0.2746 \\ \bottomrule
\end{tabular} \\ \\
(c) PSM-adjusted Sepsis Patient Dataset \\
\begin{tabular}{lccccccccc}
\toprule
Model &
  $n_C$ &
  $\lambda$ &
  $k$ &
  DT &
  LR &
  NB &
  NN &
  RF &
  SVM \\ \hline
Baseline &
  23553 &
   &
   &
  0.5778 &
  0.6607 &
  0.5122 &
  0.6207 &
  0.6494 &
  0.5463 \\ \hline

\multirow{4}{*}{MO-OBAM} &
  3240 &
  1 &
  5 &
  0.5775 &
  0.6587 &
  0.5130 &
  0.6189 &
  0.6481 &
  0.5369 \\ \cmidrule(lr){2-10}
 
  &
  2310 &
  1 &
  10 &
  0.5757 &
  0.6599 &
  0.5117 &
  \cellcolor[HTML]{F1A983}0.6162 &
  \cellcolor[HTML]{F1A983}0.6437 &
  0.5446 \\ \cmidrule(lr){2-10}

  &
  1740 &
  1 &
  15 &
  0.5768 &
  0.6610 &
  0.5093 &
  \cellcolor[HTML]{F1A983}0.6150 &
  0.6478 &
  0.5326 \\ \cmidrule(lr){2-10}

  &
  820 &
  1 &
  20 &
  \cellcolor[HTML]{F1A983}0.5749 &
  0.6605 &
  0.5131 &
  \cellcolor[HTML]{F1A983}0.6117 &
  \cellcolor[HTML]{F1A983}0.6416 &
  0.5304 \\ \bottomrule
\end{tabular}  
\end{tabular}
\label{tab: datasets ML with p-value}
\end{table}

\subsubsection{Comparison Analysis of Machine Learning Performance}\label{sec:ml perfor-comparison}

\begin{itemize}
    \item \textbf{German credit:} From Table \ref{tab: datasets ML with p-value}(a), it is noted that LR F1 scores decrease in three instances in our model: 1) $n_C=30,\lambda=0.0001,k=10$; 2) $n_C=30,\lambda=0.0001,k=15$; 3) $n_C=4,\lambda=1,k=20$. Table~\ref{tab: German credit ML F1 scores compared with Wei} reveals that \cite{zheng2019effective} did not demonstrate statistically significant differences compared to our model in these three instances. However, the $k$-anonymity algorithm indicates statistically significant superior results in these situations. Analyzing the shifting importance of QIs between our model and the $k$-anonymity algorithm in Appendix C, we demonstrate that the $k$-anonymity algorithm maintains QI importance consistently during changes in $k$, while substantial shifts occur in our model as $n_C$ increases from 4 to 30. Additionally, \cite{zheng2019effective} demonstrate statistically significant improvements at $k=10$ and $k=20$ in NB when compared to our model, as evidenced in Table~\ref{tab: German credit ML F1 scores compared with Wei}.
    \item \textbf{Adult:} According to Table~\ref{tab: Adult ML F1 scores compared with Wei}, comparing the ML model performances between our model and two alternative algorithms using the Adult dataset, we observe that the alternative algorithms outperform in DT, LR, and RF—the ML models that our model exhibits statistically significant decreases in F1 scores compared to the baseline in Table~\ref{tab: datasets ML with p-value}(b) in the most scenarios. When \cite{zheng2019effective} utilize $k=20,l=14$, their SVM F1 score is lower than ours. Apart from the aforementioned cases, for other ML models such as NB, NN, and SVM, our model maintains comparable performance levels to the other algorithms. 
    \item \textbf{Sepsis patient:} Examining Table~\ref{tab: PSM-adjusted sepsis ML compared to Wei}, we observe that among the eight highlighted cells, only three of them indicate our model has statistically significantly lower F1 scores compared to alternative algorithms, while the remaining five of them show that our model has statistically significantly higher F1 scores. Thus, based on Table~\ref{tab: PSM-adjusted sepsis ML compared to Wei}, we demonstrate that our model maintains comparable ML performance to other algorithms. 
\end{itemize}

Overall, our model demonstrates competitive performance in terms of ML effectiveness when compared to alternative algorithms across different datasets. Although there are instances where alternative algorithms outperform our model, particularly in specific configurations and ML models, our model generally maintains comparable or superior performance. 

\begin{table}
\caption{Comparison of average F1 scores for the German credit dataset among different models. Highlighted cells indicate statistical significance (p-value$<$0.05, Mann-Whitney U test) when comparing F1 scores after applying our model or another model.}
\label{tab: German credit ML F1 scores compared with Wei}
\begin{tabular}{c}
(a) Hyperparameter values that promote lower protection against homogeneity attacks \\
\resizebox{\textwidth}{!}{\begin{tabular}{llcccccc}
\toprule
\multicolumn{1}{l}{Model} &
  \multicolumn{1}{l}{Hyperparameter values} &
  \multicolumn{1}{l}{DT} &
  \multicolumn{1}{l}{LR} &
  \multicolumn{1}{l}{NB} &
  \multicolumn{1}{l}{NN} &
  \multicolumn{1}{l}{RF} &
  SVM \\ \hline
\multicolumn{1}{l}{MO-OBAM} &
  \multicolumn{1}{l}{$k$=5, $n_C$=30, $\lambda$=0.0001} &
  \multicolumn{1}{l}{0.7708} &
  \multicolumn{1}{l}{0.8079} &
  \multicolumn{1}{l}{0.7901} &
  \multicolumn{1}{l}{0.7210} &
  \multicolumn{1}{l}{0.8396} &
  0.8273 \\ \hline
\multicolumn{1}{l}{$k$-anonymity} &
  \multicolumn{1}{l}{$k$=5} &
  \multicolumn{1}{l}{0.7690} &
  \multicolumn{1}{l}{\cellcolor[HTML]{F1A983}0.8156} &
  \multicolumn{1}{l}{0.7934} &
  \multicolumn{1}{l}{0.7147} &
  \multicolumn{1}{l}{0.8412} &
  0.8252 \\
\multicolumn{1}{l}{Zheng et al} &
  \multicolumn{1}{l}{$k$=5, $l$=2} &
  \multicolumn{1}{l}{0.7704} &
  \multicolumn{1}{l}{0.8121} &
  \multicolumn{1}{l}{0.7950} &
  \multicolumn{1}{l}{0.7227} &
  \multicolumn{1}{l}{0.8390} &
  0.8235 \\ \hline
 &
   &
   &
   &
   &
   &
   &
   \\ \hline
\multicolumn{1}{l}{MO-OBAM} &
  \multicolumn{1}{l}{$k$=10, $n_C$=30, $\lambda$=0.0001} &
  \multicolumn{1}{l}{0.7683} &
  \multicolumn{1}{l}{0.8049} &
  \multicolumn{1}{l}{0.7928} &
  \multicolumn{1}{l}{0.7208} &
  \multicolumn{1}{l}{0.8400} &
  0.8244 \\ \hline
\multicolumn{1}{l}{$k$-anonymity} &
  \multicolumn{1}{l}{$k$=10} &
  \multicolumn{1}{l}{0.7707} &
  \multicolumn{1}{l}{\cellcolor[HTML]{F1A983}0.8129} &
  \multicolumn{1}{l}{0.7903} &
  \multicolumn{1}{l}{0.7047} &
  \multicolumn{1}{l}{0.8428} &
  0.8252 \\
\multicolumn{1}{l}{Zheng et al} &
  \multicolumn{1}{l}{$k$=10, $l$=2} &
  \multicolumn{1}{l}{0.7700} &
  \multicolumn{1}{l}{0.8094} &
  \multicolumn{1}{l}{0.7972} &
  \multicolumn{1}{l}{0.7170} &
  \multicolumn{1}{l}{0.8376} &
  0.8244 \\ \hline
 &
   &
   &
   &
   &
   &
   &
   \\ \hline
\multicolumn{1}{l}{MO-OBAM} &
  \multicolumn{1}{l}{$k$=15, $n_C$=30, $\lambda$=0.0001} &
  \multicolumn{1}{l}{0.7690} &
  \multicolumn{1}{l}{0.8038} &
  \multicolumn{1}{l}{0.7960} &
  \multicolumn{1}{l}{0.7156} &
  \multicolumn{1}{l}{0.8341} &
  0.8245 \\ \hline
\multicolumn{1}{l}{$k$-anonymity} &
  \multicolumn{1}{l}{$k$=15} &
  \multicolumn{1}{l}{0.7719} &
  \multicolumn{1}{l}{\cellcolor[HTML]{F1A983}0.8146} &
  \multicolumn{1}{l}{0.7912} &
  \multicolumn{1}{l}{0.7086} &
  \multicolumn{1}{l}{0.8381} &
  0.8257 \\
\multicolumn{1}{l}{Zheng et al} &
  \multicolumn{1}{l}{$k$=15, $l$=2} &
  \multicolumn{1}{l}{0.7693} &
  \multicolumn{1}{l}{0.8085} &
  \multicolumn{1}{l}{0.7938} &
  \multicolumn{1}{l}{0.7167} &
  \multicolumn{1}{l}{0.8373} &
  0.8284 \\ \hline
 &
   &
   &
   &
   &
   &
   &
   \\ \hline
\multicolumn{1}{l}{MO-OBAM} &
  \multicolumn{1}{l}{$k$=20, $n_C$=30, $\lambda$=0.0001} &
  \multicolumn{1}{l}{0.7699} &
  \multicolumn{1}{l}{0.8073} &
  \multicolumn{1}{l}{0.7939} &
  \multicolumn{1}{l}{0.7239} &
  \multicolumn{1}{l}{0.8373} &
  0.8265 \\ \hline
\multicolumn{1}{l}{$k$-anonymity} &
  \multicolumn{1}{l}{$k$=20} &
  \multicolumn{1}{l}{0.7703} &
  \multicolumn{1}{l}{0.8123} &
  \multicolumn{1}{l}{0.7916} &
  \multicolumn{1}{l}{0.7108} &
  \multicolumn{1}{l}{0.8361} &
  0.8262 \\
\multicolumn{1}{l}{Zheng et al} &
  \multicolumn{1}{l}{$k$=20, $l$=2} &
  \multicolumn{1}{l}{0.7659} &
  \multicolumn{1}{l}{0.8072} &
  \multicolumn{1}{l}{0.7943} &
  \multicolumn{1}{l}{0.7172} &
  \multicolumn{1}{l}{0.8384} &
  0.8256 \\ \hline
\end{tabular}} \\ \\
(b) Hyperparameter values that promote higher protection against homogeneity attacks \\
\resizebox{\textwidth}{!}{\begin{tabular}{llcccccc}
\toprule
\multicolumn{1}{l}{Model} &
  \multicolumn{1}{l}{Hyperparameter values} &
  \multicolumn{1}{l}{DT} &
  \multicolumn{1}{l}{LR} &
  \multicolumn{1}{l}{NB} &
  \multicolumn{1}{l}{NN} &
  \multicolumn{1}{l}{RF} &
  SVM \\ \hline
\multicolumn{1}{l}{MO-OBAM} &
  \multicolumn{1}{l}{$k$=5, $n_C$=4, $\lambda$=1} &
  \multicolumn{1}{l}{0.7701} &
  \multicolumn{1}{l}{0.8123} &
  \multicolumn{1}{l}{0.7910} &
  \multicolumn{1}{l}{0.7153} &
  \multicolumn{1}{l}{0.8382} &
  0.8239 \\ \hline
\multicolumn{1}{l}{$k$-anonymity} &
  \multicolumn{1}{l}{$k$=5} &
  \multicolumn{1}{l}{0.7690} &
  \multicolumn{1}{l}{0.8156} &
  \multicolumn{1}{l}{0.7934} &
  \multicolumn{1}{l}{0.7147} &
  \multicolumn{1}{l}{0.8412} &
  0.8252 \\
\multicolumn{1}{l}{Zheng et al} &
  \multicolumn{1}{l}{$k$=5, $l$=4} &
  \multicolumn{1}{l}{0.7690} &
  \multicolumn{1}{l}{0.8092} &
  \multicolumn{1}{l}{0.7969} &
  \multicolumn{1}{l}{0.7151} &
  \multicolumn{1}{l}{0.8358} &
  0.8226 \\ \hline
 &
   &
   &
   &
   &
   &
   &
   \\ \hline
\multicolumn{1}{l}{MO-OBAM} &
  \multicolumn{1}{l}{$k$=10, $n_C$=4, $\lambda$=1} &
  \multicolumn{1}{l}{0.7714} &
  \multicolumn{1}{l}{0.8077} &
  \multicolumn{1}{l}{0.7863} &
  \multicolumn{1}{l}{0.7148} &
  \multicolumn{1}{l}{0.8398} &
  0.8251 \\ \hline
\multicolumn{1}{l}{$k$-anonymity} &
  \multicolumn{1}{l}{$k$=10} &
  \multicolumn{1}{l}{0.7707} &
  \multicolumn{1}{l}{0.8129} &
  \multicolumn{1}{l}{0.7903} &
  \multicolumn{1}{l}{0.7047} &
  \multicolumn{1}{l}{0.8428} &
  0.8252 \\
\multicolumn{1}{l}{Zheng et al} &
  \multicolumn{1}{l}{$k$=10, $l$=4} &
  \multicolumn{1}{l}{0.7696} &
  \multicolumn{1}{l}{0.8141} &
  \multicolumn{1}{l}{\cellcolor[HTML]{F1A983}0.8027} &
  \multicolumn{1}{l}{0.7202} &
  \multicolumn{1}{l}{0.8399} &
  0.8249 \\ \hline
 &
   &
   &
   &
   &
   &
   &
   \\ \hline
\multicolumn{1}{l}{MO-OBAM} &
  \multicolumn{1}{l}{$k$=15, $n_C$=4, $\lambda$=1} &
  \multicolumn{1}{l}{0.7698} &
  \multicolumn{1}{l}{0.8090} &
  \multicolumn{1}{l}{0.7923} &
  \multicolumn{1}{l}{0.7218} &
  \multicolumn{1}{l}{0.8387} &
  0.8241 \\ \hline
\multicolumn{1}{l}{$k$-anonymity} &
  \multicolumn{1}{l}{$k$=15} &
  \multicolumn{1}{l}{0.7719} &
  \multicolumn{1}{l}{0.8146} &
  \multicolumn{1}{l}{0.7912} &
  \multicolumn{1}{l}{0.7086} &
  \multicolumn{1}{l}{0.8381} &
  0.8257 \\
\multicolumn{1}{l}{Zheng et al} &
  \multicolumn{1}{l}{$k$=15, $l$=4} &
  \multicolumn{1}{l}{0.7708} &
  \multicolumn{1}{l}{0.8065} &
  \multicolumn{1}{l}{0.7944} &
  \multicolumn{1}{l}{0.7175} &
  \multicolumn{1}{l}{0.8412} &
  0.8230 \\ \hline
 &
   &
   &
   &
   &
   &
   &
   \\ \hline
\multicolumn{1}{l}{MO-OBAM} &
  \multicolumn{1}{l}{$k$=20, $n_C$=4, $\lambda$=1} &
  \multicolumn{1}{l}{0.7696} &
  \multicolumn{1}{l}{0.8057} &
  \multicolumn{1}{l}{0.7878} &
  \multicolumn{1}{l}{0.7189} &
  \multicolumn{1}{l}{0.8351} &
  0.8264 \\ \hline
\multicolumn{1}{l}{$k$-anonymity} &
  \multicolumn{1}{l}{$k$=20} &
  \multicolumn{1}{l}{0.7703} &
  \multicolumn{1}{l}{\cellcolor[HTML]{F1A983}0.8123} &
  \multicolumn{1}{l}{0.7916} &
  \multicolumn{1}{l}{0.7108} &
  \multicolumn{1}{l}{0.8361} &
  0.8262 \\
\multicolumn{1}{l}{Zheng et al} &
  \multicolumn{1}{l}{$k$=20, $l$=4} &
  \multicolumn{1}{l}{0.7681} &
  \multicolumn{1}{l}{0.8082} &
  \multicolumn{1}{l}{\cellcolor[HTML]{F1A983}0.7981} &
  \multicolumn{1}{l}{0.7217} &
  \multicolumn{1}{l}{0.8409} &
  0.8274 \\ \bottomrule
\end{tabular}}
\end{tabular}
\end{table}

\begin{table}
\caption{Comparison of average F1 scores for the Adult dataset among different models. Highlighted cells indicate statistical significance (p-value$<$0.05, Mann-Whitney U test) when comparing F1 scores after applying our model or another model.}
\label{tab: Adult ML F1 scores compared with Wei}
\begin{tabular}{c}
(a) Hyperparameter values that promote lower protection against homogeneity attacks \\
\resizebox{\textwidth}{!}{\begin{tabular}{llcccccc}
\toprule
\multicolumn{1}{l}{Model} &
  \multicolumn{1}{l}{Hyperparameter values} &
  \multicolumn{1}{l}{DT} &
  \multicolumn{1}{l}{LR} &
  \multicolumn{1}{l}{NB} &
  \multicolumn{1}{l}{NN} &
  \multicolumn{1}{l}{RF} &
  SVM \\ \hline
\multicolumn{1}{l}{MO-OBAM} &
  \multicolumn{1}{l}{$k$=5, $n_C$=100, $\lambda$=0.0001} &
  \multicolumn{1}{l}{0.6202} &
  \multicolumn{1}{l}{0.4077} &
  \multicolumn{1}{l}{0.4192} &
  \multicolumn{1}{l}{0.3481} &
  \multicolumn{1}{l}{0.6684} &
  0.2729 \\ \hline
\multicolumn{1}{l}{$k$-anonymity} &
  \multicolumn{1}{l}{$k$=5} &
  \multicolumn{1}{l}{0.6208} &
  \multicolumn{1}{l}{0.4070} &
  \multicolumn{1}{l}{0.4202} &
  \multicolumn{1}{l}{0.3517} &
  \multicolumn{1}{l}{\cellcolor[HTML]{F1A983}0.6786} &
  0.2743 \\
\multicolumn{1}{l}{Zheng et al} &
  \multicolumn{1}{l}{$k$=5, $l$=2} &
  \multicolumn{1}{l}{\cellcolor[HTML]{F1A983}0.6192} &
  \multicolumn{1}{l}{0.4083} &
  \multicolumn{1}{l}{0.4183} &
  \multicolumn{1}{l}{0.3558} &
  \multicolumn{1}{l}{\cellcolor[HTML]{F1A983}0.6783} &
  0.2734 \\ \hline
 &
   &
   &
   &
   &
   &
   &
   \\ \hline
\multicolumn{1}{l}{MO-OBAM} &
  \multicolumn{1}{l}{$k$=10, $n_C$=100, $\lambda$=0.0001} &
  \multicolumn{1}{l}{0.6203} &
  \multicolumn{1}{l}{0.4076} &
  \multicolumn{1}{l}{0.4190} &
  \multicolumn{1}{l}{0.3495} &
  \multicolumn{1}{l}{0.6695} &
  0.2746 \\ \hline
\multicolumn{1}{l}{$k$-anonymity} &
  \multicolumn{1}{l}{$k$=10} &
  \multicolumn{1}{l}{0.6209} &
  \multicolumn{1}{l}{\cellcolor[HTML]{F1A983}0.4087} &
  \multicolumn{1}{l}{0.4210} &
  \multicolumn{1}{l}{0.3564} &
  \multicolumn{1}{l}{\cellcolor[HTML]{F1A983}0.6794} &
  0.2733 \\
\multicolumn{1}{l}{Zheng et al} &
  \multicolumn{1}{l}{$k$=10, $l$=2} &
  \multicolumn{1}{l}{0.6202} &
  \multicolumn{1}{l}{\cellcolor[HTML]{F1A983}0.4069} &
  \multicolumn{1}{l}{0.4181} &
  \multicolumn{1}{l}{0.3545} &
  \multicolumn{1}{l}{\cellcolor[HTML]{F1A983}0.6791} &
  0.2729 \\ \hline
 &
   &
   &
   &
   &
   &
   &
   \\ \hline
\multicolumn{1}{l}{MO-OBAM} &
  \multicolumn{1}{l}{$k$=15, $n_C$=100, $\lambda$=0.0001} &
  \multicolumn{1}{l}{0.6203} &
  \multicolumn{1}{l}{0.4052} &
  \multicolumn{1}{l}{0.4190} &
  \multicolumn{1}{l}{0.3477} &
  \multicolumn{1}{l}{0.6674} &
  0.2741 \\ \hline
\multicolumn{1}{l}{$k$-anonymity} &
  \multicolumn{1}{l}{$k$=15} &
  \multicolumn{1}{l}{\cellcolor[HTML]{F1A983}0.6214} &
  \multicolumn{1}{l}{0.4059} &
  \multicolumn{1}{l}{0.4199} &
  \multicolumn{1}{l}{0.3513} &
  \multicolumn{1}{l}{\cellcolor[HTML]{F1A983}0.6787} &
  0.2732 \\
\multicolumn{1}{l}{Zheng et al} &
  \multicolumn{1}{l}{$k$=15, $l$=2} &
  \multicolumn{1}{l}{0.6204} &
  \multicolumn{1}{l}{0.4047} &
  \multicolumn{1}{l}{0.4187} &
  \multicolumn{1}{l}{0.3523} &
  \multicolumn{1}{l}{\cellcolor[HTML]{F1A983}0.6758} &
  0.2753 \\ \hline
 &
   &
   &
   &
   &
   &
   &
   \\ \hline
\multicolumn{1}{l}{MO-OBAM} &
  \multicolumn{1}{l}{$k$=20, $n_C$=100, $\lambda$=0.0001} &
  \multicolumn{1}{l}{0.6203} &
  \multicolumn{1}{l}{0.4073} &
  \multicolumn{1}{l}{0.4197} &
  \multicolumn{1}{l}{0.3478} &
  \multicolumn{1}{l}{0.6688} &
  0.2746 \\ \hline
\multicolumn{1}{l}{$k$-anonymity} &
  \multicolumn{1}{l}{$k$=20} &
  \multicolumn{1}{l}{0.6203} &
  \multicolumn{1}{l}{0.4079} &
  \multicolumn{1}{l}{0.4195} &
  \multicolumn{1}{l}{0.3535} &
  \multicolumn{1}{l}{\cellcolor[HTML]{F1A983}0.6770} &
  0.2738 \\
\multicolumn{1}{l}{Zheng et al} &
  \multicolumn{1}{l}{$k$=20, $l$=2} &
  \multicolumn{1}{l}{\cellcolor[HTML]{F1A983}0.6214} &
  \multicolumn{1}{l}{0.4081} &
  \multicolumn{1}{l}{0.4200} &
  \multicolumn{1}{l}{0.3496} &
  \multicolumn{1}{l}{\cellcolor[HTML]{F1A983}0.6767} &
  0.2742 \\ \bottomrule
\end{tabular}} \\ \\ 
(b) Hyperparameter values that promote higher protection against homogeneity attacks \\
\resizebox{\textwidth}{!}{\begin{tabular}{llcccccc}
\toprule
\multicolumn{1}{l}{Model} &
  \multicolumn{1}{l}{Hyperparameter values} &
  \multicolumn{1}{l}{DT} &
  \multicolumn{1}{l}{LR} &
  \multicolumn{1}{l}{NB} &
  \multicolumn{1}{l}{NN} &
  \multicolumn{1}{l}{RF} &
  SVM \\ \hline
\multicolumn{1}{l}{MO-OBAM} &
  \multicolumn{1}{l}{$k$=5, $n_C$=4, $\lambda$=1} &
  \multicolumn{1}{l}{0.6186} &
  \multicolumn{1}{l}{0.4014} &
  \multicolumn{1}{l}{0.4193} &
  \multicolumn{1}{l}{0.3566} &
  \multicolumn{1}{l}{0.6600} &
  0.2727 \\ \hline
\multicolumn{1}{l}{$k$-anonymity} &
  \multicolumn{1}{l}{$k$=5} &
  \multicolumn{1}{l}{\cellcolor[HTML]{F1A983}0.6208} &
  \multicolumn{1}{l}{\cellcolor[HTML]{F1A983}0.4070} &
  \multicolumn{1}{l}{0.4202} &
  \multicolumn{1}{l}{0.3517} &
  \multicolumn{1}{l}{\cellcolor[HTML]{F1A983}0.6786} &
  0.2743 \\
\multicolumn{1}{l}{Zheng et al} &
  \multicolumn{1}{l}{$k$=5, $l$=14} &
  \multicolumn{1}{l}{\cellcolor[HTML]{F1A983}0.6238} &
  \multicolumn{1}{l}{\cellcolor[HTML]{F1A983}0.4050} &
  \multicolumn{1}{l}{0.4186} &
  \multicolumn{1}{l}{0.3494} &
  \multicolumn{1}{l}{\cellcolor[HTML]{F1A983}0.6689} &
  0.2732 \\ \hline
 &
   &
   &
   &
   &
   &
   &
   \\ \hline
\multicolumn{1}{l}{MO-OBAM} &
  \multicolumn{1}{l}{$k$=10, $n_C$=4, $\lambda$=1} &
  \multicolumn{1}{l}{0.6194} &
  \multicolumn{1}{l}{0.4022} &
  \multicolumn{1}{l}{0.4189} &
  \multicolumn{1}{l}{0.3492} &
  \multicolumn{1}{l}{0.6607} &
  0.2743 \\ \hline
\multicolumn{1}{l}{$k$-anonymity} &
  \multicolumn{1}{l}{$k$=10} &
  \multicolumn{1}{l}{\cellcolor[HTML]{F1A983}0.6209} &
  \multicolumn{1}{l}{\cellcolor[HTML]{F1A983}0.4087} &
  \multicolumn{1}{l}{0.4210} &
  \multicolumn{1}{l}{0.3564} &
  \multicolumn{1}{l}{\cellcolor[HTML]{F1A983}0.6794} &
  0.2733 \\
\multicolumn{1}{l}{Zheng et al} &
  \multicolumn{1}{l}{$k$=10, $l$=14} &
  \multicolumn{1}{l}{\cellcolor[HTML]{F1A983}0.6233} &
  \multicolumn{1}{l}{0.4018} &
  \multicolumn{1}{l}{0.4208} &
  \multicolumn{1}{l}{0.3477} &
  \multicolumn{1}{l}{\cellcolor[HTML]{F1A983}0.6685} &
  0.2744 \\ \hline
 &
   &
   &
   &
   &
   &
   &
   \\ \hline
\multicolumn{1}{l}{MO-OBAM} &
  \multicolumn{1}{l}{$k$=15, $n_C$=4, $\lambda$=1} &
  \multicolumn{1}{l}{0.6180} &
  \multicolumn{1}{l}{0.4011} &
  \multicolumn{1}{l}{0.4196} &
  \multicolumn{1}{l}{0.3530} &
  \multicolumn{1}{l}{0.6608} &
  0.2741 \\ \hline
\multicolumn{1}{l}{$k$-anonymity} &
  \multicolumn{1}{l}{$k$=15} &
  \multicolumn{1}{l}{\cellcolor[HTML]{F1A983}0.6214} &
  \multicolumn{1}{l}{\cellcolor[HTML]{F1A983}0.4059} &
  \multicolumn{1}{l}{0.4199} &
  \multicolumn{1}{l}{0.3513} &
  \multicolumn{1}{l}{\cellcolor[HTML]{F1A983}0.6787} &
  0.2732 \\
\multicolumn{1}{l}{Zheng et al} &
  \multicolumn{1}{l}{$k$=15, $l$=14} &
  \multicolumn{1}{l}{\cellcolor[HTML]{F1A983}0.6223} &
  \multicolumn{1}{l}{\cellcolor[HTML]{F1A983}0.4049} &
  \multicolumn{1}{l}{0.4185} &
  \multicolumn{1}{l}{0.3530} &
  \multicolumn{1}{l}{\cellcolor[HTML]{F1A983}0.6675} &
  0.2740 \\ \hline
 &
   &
   &
   &
   &
   &
   &
   \\ \hline
\multicolumn{1}{l}{MO-OBAM} &
  \multicolumn{1}{l}{$k$=20, $n_C$=4, $\lambda$=1} &
  \multicolumn{1}{l}{0.6183} &
  \multicolumn{1}{l}{0.4002} &
  \multicolumn{1}{l}{0.4182} &
  \multicolumn{1}{l}{0.3504} &
  \multicolumn{1}{l}{0.6603} &
  0.2746 \\ \hline
\multicolumn{1}{l}{$k$-anonymity} &
  \multicolumn{1}{l}{$k$=20} &
  \multicolumn{1}{l}{\cellcolor[HTML]{F1A983}0.6203} &
  \multicolumn{1}{l}{\cellcolor[HTML]{F1A983}0.4079} &
  \multicolumn{1}{l}{0.4195} &
  \multicolumn{1}{l}{0.3535} &
  \multicolumn{1}{l}{\cellcolor[HTML]{F1A983}0.6770} &
  0.2738 \\
\multicolumn{1}{l}{Zheng et al} &
  \multicolumn{1}{l}{$k$=20, $l$=14} &
  \multicolumn{1}{l}{\cellcolor[HTML]{F1A983}0.6218} &
  \multicolumn{1}{l}{\cellcolor[HTML]{F1A983}0.4045} &
  \multicolumn{1}{l}{0.4170} &
  \multicolumn{1}{l}{0.3497} &
  \multicolumn{1}{l}{\cellcolor[HTML]{F1A983}0.6678} &
  \cellcolor[HTML]{F1A983}0.2725 \\ \bottomrule
\end{tabular}}
\end{tabular}
\end{table}

\begin{table}
\caption{Comparison of average F1 scores for the Sepsis dataset among different models. Highlighted cells indicate statistical significance (p-value$<$0.05, Mann-Whitney U test) when comparing F1 scores after applying our model or another model.}
\label{tab: PSM-adjusted sepsis ML compared to Wei}
\begin{tabular}{llllllll}
\hline
\multicolumn{1}{l}{Model} &
  \multicolumn{1}{l}{Hyperparameter values} &
  \multicolumn{1}{l}{DT} &
  \multicolumn{1}{l}{LR} &
  \multicolumn{1}{l}{NB} &
  \multicolumn{1}{l}{NN} &
  \multicolumn{1}{l}{RF} &
  SVM \\ \hline
\multicolumn{1}{l}{MO-OBAM} &
  \multicolumn{1}{l}{$k$=5, $n_C$=3240, $\lambda$=1} &
  \multicolumn{1}{l}{0.5775} &
  \multicolumn{1}{l}{0.6587} &
  \multicolumn{1}{l}{0.5130} &
  \multicolumn{1}{l}{0.6189} &
  \multicolumn{1}{l}{0.6481} &
  0.5369 \\ \hline
\multicolumn{1}{l}{$k$-anonymity} &
  \multicolumn{1}{l}{$k$=5} &
  \multicolumn{1}{l}{0.5768} &
  \multicolumn{1}{l}{0.6588} &
  \multicolumn{1}{l}{0.5125} &
  \multicolumn{1}{l}{\cellcolor[HTML]{F1A983}0.6139} &
  \multicolumn{1}{l}{0.6462} &
  0.5383 \\
\multicolumn{1}{l}{Zheng et al} &
  \multicolumn{1}{l}{$k$=5, $l$=2} &
  \multicolumn{1}{l}{0.5781} &
  \multicolumn{1}{l}{0.6588} &
  \multicolumn{1}{l}{0.5106} &
  \multicolumn{1}{l}{\cellcolor[HTML]{F1A983}0.6142} &
  \multicolumn{1}{l}{0.6475} &
  0.5430 \\ \hline
 &
   &
   &
   &
   &
   &
   &
   \\ \hline
\multicolumn{1}{l}{MO-OBAM} &
  \multicolumn{1}{l}{$k$=10, $n_C$=2310, $\lambda$=1} &
  \multicolumn{1}{l}{0.5757} &
  \multicolumn{1}{l}{0.6599} &
  \multicolumn{1}{l}{0.5117} &
  \multicolumn{1}{l}{0.6162} &
  \multicolumn{1}{l}{0.6437} &
  0.5446 \\ \hline
\multicolumn{1}{l}{$k$-anonymity} &
  \multicolumn{1}{l}{$k$=10} &
  \multicolumn{1}{l}{0.5771} &
  \multicolumn{1}{l}{0.6579} &
  \multicolumn{1}{l}{0.5106} &
  \multicolumn{1}{l}{0.6153} &
  \multicolumn{1}{l}{\cellcolor[HTML]{F1A983}0.6486} &
  0.5356 \\
\multicolumn{1}{l}{Zheng et al} &
  \multicolumn{1}{l}{$k$=10, $l$=2} &
  \multicolumn{1}{l}{0.5761} &
  \multicolumn{1}{l}{0.6582} &
  \multicolumn{1}{l}{0.5087} &
  \multicolumn{1}{l}{\cellcolor[HTML]{F1A983}0.6127} &
  \multicolumn{1}{l}{0.6423} &
  0.5391 \\ \hline
 &
   &
   &
   &
   &
   &
   &
   \\ \hline
\multicolumn{1}{l}{MO-OBAM} &
  \multicolumn{1}{l}{$k$=15, $n_C$=1740, $\lambda$=1} &
  \multicolumn{1}{l}{0.5768} &
  \multicolumn{1}{l}{0.6610} &
  \multicolumn{1}{l}{0.5093} &
  \multicolumn{1}{l}{0.6150} &
  \multicolumn{1}{l}{0.6478} &
  0.5326 \\ \hline
\multicolumn{1}{l}{$k$-anonymity} &
  \multicolumn{1}{l}{$k$=15} &
  \multicolumn{1}{l}{\cellcolor[HTML]{F1A983}0.5731} &
  \multicolumn{1}{l}{\cellcolor[HTML]{F1A983}0.6559} &
  \multicolumn{1}{l}{0.5098} &
  \multicolumn{1}{l}{0.6143} &
  \multicolumn{1}{l}{0.6474} &
  0.5332 \\
\multicolumn{1}{l}{Zheng et al} &
  \multicolumn{1}{l}{$k$=15, $l$=2} &
  \multicolumn{1}{l}{0.5776} &
  \multicolumn{1}{l}{0.6597} &
  \multicolumn{1}{l}{0.5112} &
  \multicolumn{1}{l}{0.6129} &
  \multicolumn{1}{l}{0.6486} &
  0.5413 \\ \hline
 &
   &
   &
   &
   &
   &
   &
   \\ \hline
\multicolumn{1}{l}{MO-OBAM} &
  \multicolumn{1}{l}{$k$=20, $n_C$=820, $\lambda$=1} &
  \multicolumn{1}{l}{0.5749} &
  \multicolumn{1}{l}{0.6605} &
  \multicolumn{1}{l}{0.5131} &
  \multicolumn{1}{l}{0.6117} &
  \multicolumn{1}{l}{0.6416} &
  0.5304 \\ \hline
\multicolumn{1}{l}{$k$-anonymity} &
  \multicolumn{1}{l}{$k$=20} &
  \multicolumn{1}{l}{\cellcolor[HTML]{F1A983}0.5784} &
  \multicolumn{1}{l}{0.6596} &
  \multicolumn{1}{l}{0.5124} &
  \multicolumn{1}{l}{0.6138} &
  \multicolumn{1}{l}{\cellcolor[HTML]{F1A983}0.6493} &
  0.5380 \\
\multicolumn{1}{l}{Zheng et al} &
  \multicolumn{1}{l}{$k$=20, $l$=2} &
  \multicolumn{1}{l}{0.5760} &
  \multicolumn{1}{l}{0.6597} &
  \multicolumn{1}{l}{0.5127} &
  \multicolumn{1}{l}{0.6130} &
  \multicolumn{1}{l}{0.6439} &
  0.5443 \\\hline
\end{tabular}%

\end{table}

\section{DISCUSSION and CONCLUSION}\label{sec:conclusion}
In this study, we propose a novel model, MO-OBAM, designed to handle both categorical and numerical variables while simultaneously addressing information loss and protection against attacks. We formulate three hypotheses to evaluate the performance of our proposed model. To assess the efficiency, privacy preservation capabilities, and impact on ML performance of MO-OBAM, we conduct experiments using datasets of varying sizesfrom census, finance, and healthcare domains. Additionally, we compare the results of our model with those of two others from literature. This comprehensive evaluation provides an in-depth understanding of MO-OBAM’s strengths and limitations across different contexts.

Our empirical results align with conclusions summarized in Section~\ref{sec:related work}, validating the insights from existing research. These findings also provide a foundation for understanding how our model performs under various conditions. Building on these insights, our experiments highlight the crucial role of the number of clusters, not only in terms of privacy preservation but also regarding ML performance. When the number of clusters is small, resulting in a higher level of anonymization but greater information loss, we observe robust protection against attacks but also a significant decrease in the importance of QIs. Conversely, when the number of clusters is large, leading to a lower level of anonymization but less information loss, we observe a small number of individuals still vulnerable to attacks but better retention of the importance of QIs. Notably, if a QI holds paramount importance for ML tasks, it is advisable to limit anonymization to preserve its feature importance. Fung et al.\cite{fung2007anonymizing} and Pitoglou et al.\cite{pitoglou2022exploring} also mentioned this point in their study. These findings underscore the need to carefully select the number of clusters. It is critical for maintaining optimal ML performance while ensuring adequate privacy protection. Therefore, we provide detailed instructions for tuning the number of clusters to alleviate this challenge.

Our comparative analysis supports three key hypotheses regarding our model's performance in specific scenarios. Firstly, our model is able to have lower information loss compared to the two alternative algorithms. Secondly, it offers enhanced protection against both linkage and homogeneity attacks by reducing the number of vulnerable people. Thirdly, our model does not negatively impact the performance of ML models. In addition, our model strikes a superior balance between data utility and robust protection against attacks compared to the other two algorithms. Specifically, despite some variations in performance for specific ML models and configurations, our model generally maintains comparable ML performance across most scenarios, and our model demonstrates a significant reduction in the number of individuals at risk of linkage attacks, achieving approximately 96\% to 98\% decreases in susceptibility. It shows our model's effectiveness in preserving privacy without compromising the quality of the data.

This study explores the application of optimization techniques in privacy protection, thereby addressing a specific gap in the current literature. In addition, our optimization-based approach allows for more precise control over data modifications, making it particularly suitable for scenarios where maintaining high data utility is critical. This new perspective offers an innovative solution to privacy challenges and contributes to the diversity of methods in this field. Leveraging the flexibility of our multi-objective framework, we can accommodate various requirements, including enhancing ML performance on anonymized data. Because our model is flexible and extensible, we aim to incorporate additional types of privacy attacks, such as skewness attacks and similarity attacks \cite{li2006t}, to further enhance the robustness of the framework. Additionally, in our numerical results, we demonstrate that feasibility for the original problem has been attained, which is sufficient to prevent the considered privacy attacks. Nevertheless, in future work, we will investigate alternative solution methods that are provably globally convergent.

\bibliography{references}


\begin{thebibliography}{72}
\ifx \bisbn   \undefined \def \bisbn  #1{ISBN #1}\fi
\ifx \binits  \undefined \def \binits#1{#1}\fi
\ifx \bauthor  \undefined \def \bauthor#1{#1}\fi
\ifx \batitle  \undefined \def \batitle#1{#1}\fi
\ifx \bjtitle  \undefined \def \bjtitle#1{#1}\fi
\ifx \bvolume  \undefined \def \bvolume#1{\textbf{#1}}\fi
\ifx \byear  \undefined \def \byear#1{#1}\fi
\ifx \bissue  \undefined \def \bissue#1{#1}\fi
\ifx \bfpage  \undefined \def \bfpage#1{#1}\fi
\ifx \blpage  \undefined \def \blpage #1{#1}\fi
\ifx \burl  \undefined \def \burl#1{\textsf{#1}}\fi
\ifx \doiurl  \undefined \def \doiurl#1{\url{https://doi.org/#1}}\fi
\ifx \betal  \undefined \def \betal{\textit{et al.}}\fi
\ifx \binstitute  \undefined \def \binstitute#1{#1}\fi
\ifx \binstitutionaled  \undefined \def \binstitutionaled#1{#1}\fi
\ifx \bctitle  \undefined \def \bctitle#1{#1}\fi
\ifx \beditor  \undefined \def \beditor#1{#1}\fi
\ifx \bpublisher  \undefined \def \bpublisher#1{#1}\fi
\ifx \bbtitle  \undefined \def \bbtitle#1{#1}\fi
\ifx \bedition  \undefined \def \bedition#1{#1}\fi
\ifx \bseriesno  \undefined \def \bseriesno#1{#1}\fi
\ifx \blocation  \undefined \def \blocation#1{#1}\fi
\ifx \bsertitle  \undefined \def \bsertitle#1{#1}\fi
\ifx \bsnm \undefined \def \bsnm#1{#1}\fi
\ifx \bsuffix \undefined \def \bsuffix#1{#1}\fi
\ifx \bparticle \undefined \def \bparticle#1{#1}\fi
\ifx \barticle \undefined \def \barticle#1{#1}\fi
\bibcommenthead
\ifx \bconfdate \undefined \def \bconfdate #1{#1}\fi
\ifx \botherref \undefined \def \botherref #1{#1}\fi
\ifx \url \undefined \def \url#1{\textsf{#1}}\fi
\ifx \bchapter \undefined \def \bchapter#1{#1}\fi
\ifx \bbook \undefined \def \bbook#1{#1}\fi
\ifx \bcomment \undefined \def \bcomment#1{#1}\fi
\ifx \oauthor \undefined \def \oauthor#1{#1}\fi
\ifx \citeauthoryear \undefined \def \citeauthoryear#1{#1}\fi
\ifx \endbibitem  \undefined \def \endbibitem {}\fi
\ifx \bconflocation  \undefined \def \bconflocation#1{#1}\fi
\ifx \arxivurl  \undefined \def \arxivurl#1{\textsf{#1}}\fi
\csname PreBibitemsHook\endcsname

\bibitem[\protect\citeauthoryear{News}{2018}]{insideainews2018netflix}
\begin{botherref}
\oauthor{\bsnm{News}, \binits{I.A.}}:
Netflix Uses Big Data to Drive Success.
\url{https://insideainews.com/2018/01/20/netflix-uses-big-data-drive-success/}.
Accessed: 2025-04-17
(2018)
\end{botherref}
\endbibitem

\bibitem[\protect\citeauthoryear{Küpper et~al.}{2020}]{bcg2020share}
\begin{botherref}
\oauthor{\bsnm{Küpper}, \binits{D.}},
\oauthor{\bsnm{Okur}, \binits{A.}},
\oauthor{\bsnm{Betti}, \binits{F.}},
\oauthor{\bsnm{Bezamat}, \binits{F.}},
\oauthor{\bsnm{Fendri}, \binits{M.}},
\oauthor{\bsnm{Fernandez}, \binits{B.}}:
Share to Gain: Unlocking Data Value in Manufacturing.
\url{https://www.bcg.com/publications/2020/manufacturers-unlock-value-from-data-sharing}.
Accessed: 2025-04-18
(2020)
\end{botherref}
\endbibitem

\bibitem[\protect\citeauthoryear{Stark et~al.}{2025}]{stark2025call}
\begin{barticle}
\bauthor{\bsnm{Stark}, \binits{Z.}},
\bauthor{\bsnm{Glazer}, \binits{D.}},
\bauthor{\bsnm{Hofmann}, \binits{O.}},
\bauthor{\bsnm{Rendon}, \binits{A.}},
\bauthor{\bsnm{Marshall}, \binits{C.R.}},
\bauthor{\bsnm{Ginsburg}, \binits{G.S.}},
\bauthor{\bsnm{Lunt}, \binits{C.}},
\bauthor{\bsnm{Allen}, \binits{N.}},
\bauthor{\bsnm{Effingham}, \binits{M.}},
\bauthor{\bsnm{Hastings~Ward}, \binits{J.}}, \betal:
\batitle{A call to action to scale up research and clinical genomic data sharing}.
\bjtitle{Nature Reviews Genetics}
\bvolume{26}(\bissue{2}),
\bfpage{141}--\blpage{147}
(\byear{2025})
\end{barticle}
\endbibitem

\bibitem[\protect\citeauthoryear{{Identity Theft Resource Center}}{2024}]{idtheft2024}
\begin{botherref}
\oauthor{\bsnm{{Identity Theft Resource Center}}}:
2024 Data Breach Report.
\url{https://www.idtheftcenter.org/publication/2024-data-breach-report/}, (accessed Apr 10, 2025)
(2024)
\end{botherref}
\endbibitem

\bibitem[\protect\citeauthoryear{GlobeNewswire}{}]{dataprotection2024}
\begin{botherref}
\oauthor{\bsnm{GlobeNewswire}}:
Data Protection Business Research Report 2024: Market to Reach \$129.6 Billion by 2030 from \$77.9 Billion in 2023, Fueled by Growing Global Data Regulations.
\url{https://www.globenewswire.com/news-release/2024/09/02/2939151/0/en/Data-Protection-Business-Research-Report-2024-Market-to-Reach-129-6-Billion-by-2030-from-77-9-Billion-in-2023-Fueled-by-Growing-Global-Data-Regulations.html?utm_source=chatgpt.com} (accessed Apr 11, 2025)
\end{botherref}
\endbibitem

\bibitem[\protect\citeauthoryear{{American Hospital Association}}{2024}]{aha2024}
\begin{botherref}
\oauthor{\bsnm{{American Hospital Association}}}:
A Look at 2024’s Health Care Cybersecurity Challenges.
\url{https://www.aha.org/news/aha-cyber-intel/2024-10-07-look-2024s-health-care-cybersecurity-challenges?utm_source=chatgpt.com} (accessed Apr 10, 2025)
(2024)
\end{botherref}
\endbibitem

\bibitem[\protect\citeauthoryear{Arbuckle and El~Emam}{2020}]{arbuckle2020building}
\begin{bbook}
\bauthor{\bsnm{Arbuckle}, \binits{L.}},
\bauthor{\bsnm{El~Emam}, \binits{K.}}:
\bbtitle{Building an Anonymization Pipeline: Creating Safe Data}.
\bpublisher{O'Reilly Media}, \blocation{???}
(\byear{2020})
\end{bbook}
\endbibitem

\bibitem[\protect\citeauthoryear{data~protection Regulation~({GDPR})}{2016}]{gdpr}
\begin{botherref}
\oauthor{\bsnm{Regulation~({GDPR})}, \binits{G.}}
\url{https://gdpr-info.eu/}.
Accessed: 2024-12-12
(2016)
\end{botherref}
\endbibitem

\bibitem[\protect\citeauthoryear{({CCPA})}{2018}]{ccpa}
\begin{botherref}
\oauthor{\bsnm{({CCPA})}, \binits{C.C.P.A.}}
\url{https://oag.ca.gov/privacy/ccpa}.
Accessed: 2024-12-12
(2018)
\end{botherref}
\endbibitem

\bibitem[\protect\citeauthoryear{for De-identification of Protected Health Information in Accordance with~the Health Insurance~Portability and Rule}{1996}]{hipaa}
\begin{botherref}
\oauthor{\bsnm{Health Insurance~Portability}, \binits{G.R.M.}},
\oauthor{\bsnm{Rule}, \binits{A.A.H.P.}}
\url{https://www.hhs.gov/hipaa/for-professionals/special-topics/de-identification/index.html}.
Accessed: 2024-12-12
(1996)
\end{botherref}
\endbibitem

\bibitem[\protect\citeauthoryear{Brough et~al.}{2022}]{brough2022bulletproof}
\begin{barticle}
\bauthor{\bsnm{Brough}, \binits{A.R.}},
\bauthor{\bsnm{Norton}, \binits{D.A.}},
\bauthor{\bsnm{Sciarappa}, \binits{S.L.}},
\bauthor{\bsnm{John}, \binits{L.K.}}:
\batitle{The bulletproof glass effect: Unintended consequences of privacy notices}.
\bjtitle{Journal of Marketing Research}
\bvolume{59}(\bissue{4}),
\bfpage{739}--\blpage{754}
(\byear{2022})
\end{barticle}
\endbibitem

\bibitem[\protect\citeauthoryear{El~Emam et~al.}{2011}]{el2011systematic}
\begin{barticle}
\bauthor{\bsnm{El~Emam}, \binits{K.}},
\bauthor{\bsnm{Jonker}, \binits{E.}},
\bauthor{\bsnm{Arbuckle}, \binits{L.}},
\bauthor{\bsnm{Malin}, \binits{B.}}:
\batitle{A systematic review of re-identification attacks on health data}.
\bjtitle{PloS one}
\bvolume{6}(\bissue{12}),
\bfpage{28071}
(\byear{2011})
\end{barticle}
\endbibitem

\bibitem[\protect\citeauthoryear{Benitez and Malin}{2010}]{benitez2010evaluating}
\begin{barticle}
\bauthor{\bsnm{Benitez}, \binits{K.}},
\bauthor{\bsnm{Malin}, \binits{B.}}:
\batitle{Evaluating re-identification risks with respect to the {HIPAA} privacy rule}.
\bjtitle{Journal of the American Medical Informatics Association}
\bvolume{17}(\bissue{2}),
\bfpage{169}--\blpage{177}
(\byear{2010})
\end{barticle}
\endbibitem

\bibitem[\protect\citeauthoryear{Kwok et~al.}{2011}]{kwok2011harder}
\begin{botherref}
\oauthor{\bsnm{Kwok}, \binits{P.}},
\oauthor{\bsnm{Davern}, \binits{M.}},
\oauthor{\bsnm{Hair}, \binits{E.}},
\oauthor{\bsnm{Lafky}, \binits{D.}}:
Harder than you think: A case study of re-identification risk of {HIPAA}-compliant records.
Chicago: NORC at The University of Chicago. Abstract
\textbf{302255}
(2011)
\end{botherref}
\endbibitem

\bibitem[\protect\citeauthoryear{Janmey and Elkin}{2018}]{janmey2018re}
\begin{bchapter}
\bauthor{\bsnm{Janmey}, \binits{V.}},
\bauthor{\bsnm{Elkin}, \binits{P.L.}}:
\bctitle{Re-identification risk in {HIPAA} de-identified datasets: The mva attack}.
In: \bbtitle{AMIA Annual Symposium Proceedings},
vol. \bseriesno{2018},
p. \bfpage{1329}
(\byear{2018}).
\bcomment{American Medical Informatics Association}
\end{bchapter}
\endbibitem

\bibitem[\protect\citeauthoryear{Sweeney}{2002}]{sweeney2002k}
\begin{barticle}
\bauthor{\bsnm{Sweeney}, \binits{L.}}:
\batitle{$k$-anonymity: A model for protecting privacy}.
\bjtitle{International Journal of Uncertainty, Fuzziness and Knowledge-based Systems}
\bvolume{10}(\bissue{05}),
\bfpage{557}--\blpage{570}
(\byear{2002})
\end{barticle}
\endbibitem

\bibitem[\protect\citeauthoryear{Pifer}{2019}]{healthcaredive2019breaches}
\begin{botherref}
\oauthor{\bsnm{Pifer}, \binits{R.}}:
More than 70\% of hospital data breaches include sensitive info.
\url{https://www.healthcaredive.com/news/more-than-70-of-hospital-data-breaches-include-sensitive-info/563517/}.
Accessed: 2025-04-11
(2019)
\end{botherref}
\endbibitem

\bibitem[\protect\citeauthoryear{Machanavajjhala et~al.}{2007}]{machanavajjhala2007diversity}
\begin{barticle}
\bauthor{\bsnm{Machanavajjhala}, \binits{A.}},
\bauthor{\bsnm{Kifer}, \binits{D.}},
\bauthor{\bsnm{Gehrke}, \binits{J.}},
\bauthor{\bsnm{Venkitasubramaniam}, \binits{M.}}:
\batitle{$l$-diversity: Privacy beyond $k$-anonymity}.
\bjtitle{ACM Transactions on Knowledge Discovery from Data (TKDD)}
\bvolume{1}(\bissue{1}),
\bfpage{3}
(\byear{2007})
\end{barticle}
\endbibitem

\bibitem[\protect\citeauthoryear{Li et~al.}{2006}]{li2006t}
\begin{bchapter}
\bauthor{\bsnm{Li}, \binits{N.}},
\bauthor{\bsnm{Li}, \binits{T.}},
\bauthor{\bsnm{Venkatasubramanian}, \binits{S.}}:
\bctitle{$t$-closeness: Privacy beyond $k$-anonymity and $l$-diversity}.
In: \bbtitle{2007 IEEE 23rd International Conference on Data Engineering},
pp. \bfpage{106}--\blpage{115}
(\byear{2006}).
\bcomment{IEEE}
\end{bchapter}
\endbibitem

\bibitem[\protect\citeauthoryear{Zaki}{2024}]{zaki2024securing}
\begin{botherref}
\oauthor{\bsnm{Zaki}, \binits{H.}}:
Securing insights: Safeguarding sensitive data in machine learning through privacy-preserving techniques.
Technical report,
EasyChair
(2024)
\end{botherref}
\endbibitem

\bibitem[\protect\citeauthoryear{Turgay et~al.}{2023}]{turgay2023perturbation}
\begin{barticle}
\bauthor{\bsnm{Turgay}, \binits{S.}},
\bauthor{\bsnm{{\.I}lter}, \binits{{\. I}.}}, \betal:
\batitle{Perturbation methods for protecting data privacy: A review of techniques and applications}.
\bjtitle{Automation and Machine Learning}
\bvolume{4}(\bissue{2}),
\bfpage{31}--\blpage{41}
(\byear{2023})
\end{barticle}
\endbibitem

\bibitem[\protect\citeauthoryear{Liang and Samavi}{2020}]{liang2020optimization}
\begin{barticle}
\bauthor{\bsnm{Liang}, \binits{Y.}},
\bauthor{\bsnm{Samavi}, \binits{R.}}:
\batitle{Optimization-based $k$-anonymity algorithms}.
\bjtitle{Computers \& Security}
\bvolume{93},
\bfpage{101753}
(\byear{2020})
\end{barticle}
\endbibitem

\bibitem[\protect\citeauthoryear{Ranjan}{2025}]{ranjan2025behavioural}
\begin{barticle}
\bauthor{\bsnm{Ranjan}, \binits{R.}}:
\batitle{Behavioural finance in banking and management: A study on the trends and challenges in the banking industry}.
\bjtitle{Asian Journal of Economics, Business and Accounting}
\bvolume{25}(\bissue{1}),
\bfpage{374}--\blpage{386}
(\byear{2025})
\end{barticle}
\endbibitem

\bibitem[\protect\citeauthoryear{Al~Zaabi and Alhashmi}{2024}]{al2024big}
\begin{botherref}
\oauthor{\bsnm{Al~Zaabi}, \binits{M.}},
\oauthor{\bsnm{Alhashmi}, \binits{S.M.}}:
Big data security and privacy in healthcare: A systematic review and future research directions.
Information Development,
02666669241247781
(2024)
\end{botherref}
\endbibitem

\bibitem[\protect\citeauthoryear{Martin and Palmatier}{2020}]{MARTIN2020449}
\begin{barticle}
\bauthor{\bsnm{Martin}, \binits{K.D.}},
\bauthor{\bsnm{Palmatier}, \binits{R.W.}}:
\batitle{Data privacy in retail: Navigating tensions and directing future research}.
\bjtitle{Journal of Retailing}
\bvolume{96}(\bissue{4}),
\bfpage{449}--\blpage{457}
(\byear{2020})
\doiurl{10.1016/j.jretai.2020.10.002}
\end{barticle}
\endbibitem

\bibitem[\protect\citeauthoryear{Reidenberg and Schaub}{2018}]{reidenberg2018achieving}
\begin{barticle}
\bauthor{\bsnm{Reidenberg}, \binits{J.R.}},
\bauthor{\bsnm{Schaub}, \binits{F.}}:
\batitle{Achieving big data privacy in education}.
\bjtitle{Theory and Research in Education}
\bvolume{16}(\bissue{3}),
\bfpage{263}--\blpage{279}
(\byear{2018})
\end{barticle}
\endbibitem

\bibitem[\protect\citeauthoryear{Slijep{\v{c}}evi{\'c} et~al.}{2021}]{slijepvcevic2021k}
\begin{barticle}
\bauthor{\bsnm{Slijep{\v{c}}evi{\'c}}, \binits{D.}},
\bauthor{\bsnm{Henzl}, \binits{M.}},
\bauthor{\bsnm{Klausner}, \binits{L.D.}},
\bauthor{\bsnm{Dam}, \binits{T.}},
\bauthor{\bsnm{Kieseberg}, \binits{P.}},
\bauthor{\bsnm{Zeppelzauer}, \binits{M.}}:
\batitle{$k$-anonymity in practice: How generalisation and suppression affect machine learning classifiers}.
\bjtitle{Computers \& Security}
\bvolume{111},
\bfpage{102488}
(\byear{2021})
\end{barticle}
\endbibitem

\bibitem[\protect\citeauthoryear{Fung et~al.}{2010}]{fung2010privacy}
\begin{barticle}
\bauthor{\bsnm{Fung}, \binits{B.C.}},
\bauthor{\bsnm{Wang}, \binits{K.}},
\bauthor{\bsnm{Chen}, \binits{R.}},
\bauthor{\bsnm{Yu}, \binits{P.S.}}:
\batitle{Privacy-preserving data publishing: A survey of recent developments}.
\bjtitle{ACM Computing Surveys (Csur)}
\bvolume{42}(\bissue{4}),
\bfpage{1}--\blpage{53}
(\byear{2010})
\end{barticle}
\endbibitem

\bibitem[\protect\citeauthoryear{El~Emam}{2013}]{ElEmamKhaled2013MtPo}
\begin{bchapter}
\bauthor{\bsnm{El~Emam}, \binits{K.}}:
\bctitle{Measuring the {P}robability of {R}e-{I}dentification}.
In: \bbtitle{Guide to the De-Identification of Personal Health Information},
pp. \bfpage{196}--\blpage{215}.
\bpublisher{Auerbach Publications}, \blocation{???}
(\byear{2013}).
\doiurl{10.1201/b14764-20}
\end{bchapter}
\endbibitem

\bibitem[\protect\citeauthoryear{Willenborg and De~Waal}{2001}]{willenborg-2001}
\begin{bbook}
\bauthor{\bsnm{Willenborg}, \binits{L.}},
\bauthor{\bsnm{De~Waal}, \binits{T.}}:
\bbtitle{{Elements of Statistical Disclosure Control}},
(\byear{2001}).
\doiurl{10.1007/978-1-4613-0121-9} .
\burl{https://doi.org/10.1007/978-1-4613-0121-9}
\end{bbook}
\endbibitem

\bibitem[\protect\citeauthoryear{Domingo-Ferrer and Torra}{2001}]{domingo2001disclosure}
\begin{botherref}
\oauthor{\bsnm{Domingo-Ferrer}, \binits{J.}},
\oauthor{\bsnm{Torra}, \binits{V.}}:
Disclosure control methods and information loss for microdata.
Confidentiality, Disclosure, and Data Access: Theory and Practical Applications for Statistical Agencies,
91--110
(2001)
\end{botherref}
\endbibitem

\bibitem[\protect\citeauthoryear{Khan et~al.}{2022}]{khan2022tau}
\begin{barticle}
\bauthor{\bsnm{Khan}, \binits{R.}},
\bauthor{\bsnm{Tao}, \binits{X.}},
\bauthor{\bsnm{Anjum}, \binits{A.}},
\bauthor{\bsnm{Malik}, \binits{S.R.}},
\bauthor{\bsnm{Yu}, \binits{S.}},
\bauthor{\bsnm{Khan}, \binits{A.}},
\bauthor{\bsnm{Rehman}, \binits{W.}},
\bauthor{\bsnm{Malik}, \binits{H.}}:
\batitle{($\tau$, m)-slicedbucket privacy model for sequential anonymization for improving privacy and utility}.
\bjtitle{Transactions on Emerging Telecommunications Technologies}
\bvolume{33}(\bissue{6}),
\bfpage{4130}
(\byear{2022})
\end{barticle}
\endbibitem

\bibitem[\protect\citeauthoryear{Sweeney}{2002}]{sweeney2002achieving}
\begin{barticle}
\bauthor{\bsnm{Sweeney}, \binits{L.}}:
\batitle{Achieving k-anonymity privacy protection using generalization and suppression}.
\bjtitle{International Journal of Uncertainty, Fuzziness and Knowledge-Based Systems}
\bvolume{10}(\bissue{05}),
\bfpage{571}--\blpage{588}
(\byear{2002})
\end{barticle}
\endbibitem

\bibitem[\protect\citeauthoryear{Cao and Karras}{2012}]{cao2012publishing}
\begin{botherref}
\oauthor{\bsnm{Cao}, \binits{J.}},
\oauthor{\bsnm{Karras}, \binits{P.}}:
Publishing microdata with a robust privacy guarantee.
arXiv preprint arXiv:1208.0220
(2012)
\end{botherref}
\endbibitem

\bibitem[\protect\citeauthoryear{Khan et~al.}{2020}]{khan2020theta}
\begin{barticle}
\bauthor{\bsnm{Khan}, \binits{R.}},
\bauthor{\bsnm{Tao}, \binits{X.}},
\bauthor{\bsnm{Anjum}, \binits{A.}},
\bauthor{\bsnm{Kanwal}, \binits{T.}},
\bauthor{\bsnm{Malik}, \binits{S.U.R.}},
\bauthor{\bsnm{Khan}, \binits{A.}},
\bauthor{\bsnm{Rehman}, \binits{W.U.}},
\bauthor{\bsnm{Maple}, \binits{C.}}:
\batitle{$\theta$-sensitive $k$-anonymity: An anonymization model for iot based electronic health records}.
\bjtitle{Electronics}
\bvolume{9}(\bissue{5}),
\bfpage{716}
(\byear{2020})
\end{barticle}
\endbibitem

\bibitem[\protect\citeauthoryear{Domingo-Ferrer and Torra}{2005}]{domingo2005ordinal}
\begin{barticle}
\bauthor{\bsnm{Domingo-Ferrer}, \binits{J.}},
\bauthor{\bsnm{Torra}, \binits{V.}}:
\batitle{Ordinal, continuous and heterogeneous $k$-anonymity through microaggregation}.
\bjtitle{Data Mining and Knowledge Discovery}
\bvolume{11},
\bfpage{195}--\blpage{212}
(\byear{2005})
\end{barticle}
\endbibitem

\bibitem[\protect\citeauthoryear{Fung et~al.}{2005}]{fung2005top}
\begin{bchapter}
\bauthor{\bsnm{Fung}, \binits{B.C.}},
\bauthor{\bsnm{Wang}, \binits{K.}},
\bauthor{\bsnm{Yu}, \binits{P.S.}}:
\bctitle{Top-down specialization for information and privacy preservation}.
In: \bbtitle{21st International Conference on Data Engineering (ICDE'05)},
pp. \bfpage{205}--\blpage{216}
(\byear{2005}).
\bcomment{IEEE}
\end{bchapter}
\endbibitem

\bibitem[\protect\citeauthoryear{Wang et~al.}{2004}]{wang2004bottom}
\begin{bchapter}
\bauthor{\bsnm{Wang}, \binits{K.}},
\bauthor{\bsnm{Yu}, \binits{P.S.}},
\bauthor{\bsnm{Chakraborty}, \binits{S.}}:
\bctitle{Bottom-up generalization: A data mining solution to privacy protection}.
In: \bbtitle{Fourth IEEE International Conference on Data Mining (ICDM'04)},
pp. \bfpage{249}--\blpage{256}
(\byear{2004}).
\bcomment{IEEE}
\end{bchapter}
\endbibitem

\bibitem[\protect\citeauthoryear{Chen et~al.}{2023}]{chen2023sensitivity}
\begin{bchapter}
\bauthor{\bsnm{Chen}, \binits{S.}},
\bauthor{\bsnm{Wang}, \binits{B.}},
\bauthor{\bsnm{Chen}, \binits{Y.}},
\bauthor{\bsnm{Ma}, \binits{Y.}},
\bauthor{\bsnm{Xing}, \binits{T.}},
\bauthor{\bsnm{Zhao}, \binits{J.}}:
\bctitle{Sensitivity-based (p, $\alpha$, k)-anonymity privacy protection algorithm}.
In: \bbtitle{2023 IEEE 3rd International Conference on Computer Communication and Artificial Intelligence (CCAI)},
pp. \bfpage{140}--\blpage{146}
(\byear{2023}).
\bcomment{IEEE}
\end{bchapter}
\endbibitem

\bibitem[\protect\citeauthoryear{Wang et~al.}{2020}]{wang2020enhanced}
\begin{bchapter}
\bauthor{\bsnm{Wang}, \binits{N.}},
\bauthor{\bsnm{Song}, \binits{H.}},
\bauthor{\bsnm{Luo}, \binits{T.}},
\bauthor{\bsnm{Sun}, \binits{J.}},
\bauthor{\bsnm{Li}, \binits{J.}}:
\bctitle{Enhanced p-sensitive k-anonymity models for achieving better privacy}.
In: \bbtitle{2020 IEEE/CIC International Conference on Communications in China (ICCC)},
pp. \bfpage{148}--\blpage{153}
(\byear{2020}).
\bcomment{IEEE}
\end{bchapter}
\endbibitem

\bibitem[\protect\citeauthoryear{Amiri et~al.}{2023}]{amiri2023enhancing}
\begin{barticle}
\bauthor{\bsnm{Amiri}, \binits{F.}},
\bauthor{\bsnm{Khan}, \binits{R.}},
\bauthor{\bsnm{Anjum}, \binits{A.}},
\bauthor{\bsnm{Syed}, \binits{M.H.}},
\bauthor{\bsnm{Rehman}, \binits{S.}}:
\batitle{Enhancing utility in anonymized data against the adversary’s background knowledge}.
\bjtitle{Applied Sciences}
\bvolume{13}(\bissue{7}),
\bfpage{4091}
(\byear{2023})
\end{barticle}
\endbibitem

\bibitem[\protect\citeauthoryear{Doka et~al.}{2015}]{doka2015k}
\begin{bchapter}
\bauthor{\bsnm{Doka}, \binits{K.}},
\bauthor{\bsnm{Xue}, \binits{M.}},
\bauthor{\bsnm{Tsoumakos}, \binits{D.}},
\bauthor{\bsnm{Karras}, \binits{P.}}:
\bctitle{$k$-anonymization by freeform generalization}.
In: \bbtitle{Proceedings of the 10th ACM Symposium on Information, Computer and Communications Security},
pp. \bfpage{519}--\blpage{530}
(\byear{2015})
\end{bchapter}
\endbibitem

\bibitem[\protect\citeauthoryear{Iyengar}{2002}]{iyengar2002transforming}
\begin{bchapter}
\bauthor{\bsnm{Iyengar}, \binits{V.S.}}:
\bctitle{Transforming data to satisfy privacy constraints}.
In: \bbtitle{Proceedings of the Eighth ACM SIGKDD International Conference on Knowledge Discovery and Data Mining},
pp. \bfpage{279}--\blpage{288}
(\byear{2002})
\end{bchapter}
\endbibitem

\bibitem[\protect\citeauthoryear{Batista et~al.}{2022}]{batista2022privacy}
\begin{barticle}
\bauthor{\bsnm{Batista}, \binits{E.}},
\bauthor{\bsnm{Mart{\'\i}nez-Ballest{\'e}}, \binits{A.}},
\bauthor{\bsnm{Solanas}, \binits{A.}}:
\batitle{Privacy-preserving process mining: A microaggregation-based approach}.
\bjtitle{Journal of Information Security and Applications}
\bvolume{68},
\bfpage{103235}
(\byear{2022})
\end{barticle}
\endbibitem

\bibitem[\protect\citeauthoryear{Singh and Singh}{2022}]{singh2022social}
\begin{botherref}
\oauthor{\bsnm{Singh}, \binits{A.}},
\oauthor{\bsnm{Singh}, \binits{M.}}:
Social networks privacy preservation: A novel framework.
Cybernetics and Systems,
1--32
(2022)
\end{botherref}
\endbibitem

\bibitem[\protect\citeauthoryear{Aleroud et~al.}{2024}]{aleroud2024privacy}
\begin{barticle}
\bauthor{\bsnm{Aleroud}, \binits{A.}},
\bauthor{\bsnm{Shariah}, \binits{M.}},
\bauthor{\bsnm{Malkawi}, \binits{R.}},
\bauthor{\bsnm{Khamaiseh}, \binits{S.Y.}},
\bauthor{\bsnm{Al-Alaj}, \binits{A.}}:
\batitle{A privacy-enhanced human activity recognition using gan \& entropy ranking of microaggregated data}.
\bjtitle{Cluster Computing}
\bvolume{27}(\bissue{2}),
\bfpage{2117}--\blpage{2132}
(\byear{2024})
\end{barticle}
\endbibitem

\bibitem[\protect\citeauthoryear{Abidi et~al.}{2018}]{abidi2018hybrid}
\begin{botherref}
\oauthor{\bsnm{Abidi}, \binits{B.}},
\oauthor{\bsnm{Ben~Yahia}, \binits{S.}},
\oauthor{\bsnm{Perera}, \binits{C.}}:
Hybrid microaggregation for privacy preserving data mining. J Ambient Intell Human Comput
(2018)
\end{botherref}
\endbibitem

\bibitem[\protect\citeauthoryear{Wu et~al.}{2019}]{wu2019micro}
\begin{barticle}
\bauthor{\bsnm{Wu}, \binits{X.}},
\bauthor{\bsnm{Wei}, \binits{Y.}},
\bauthor{\bsnm{Jiang}, \binits{T.}},
\bauthor{\bsnm{Wang}, \binits{Y.}},
\bauthor{\bsnm{Jiang}, \binits{S.}}:
\batitle{A micro-aggregation algorithm based on density partition method for anonymizing biomedical data}.
\bjtitle{Current Bioinformatics}
\bvolume{14}(\bissue{7}),
\bfpage{667}--\blpage{675}
(\byear{2019})
\end{barticle}
\endbibitem

\bibitem[\protect\citeauthoryear{Aminifar et~al.}{2021}]{aminifar2021diversity}
\begin{bchapter}
\bauthor{\bsnm{Aminifar}, \binits{A.}},
\bauthor{\bsnm{Rabbi}, \binits{F.}},
\bauthor{\bsnm{Pun}, \binits{V.K.I.}},
\bauthor{\bsnm{Lamo}, \binits{Y.}}:
\bctitle{Diversity-aware anonymization for structured health data}.
In: \bbtitle{2021 43rd Annual International Conference of the IEEE Engineering in Medicine \& Biology Society (EMBC)},
pp. \bfpage{2148}--\blpage{2154}
(\byear{2021}).
\bcomment{IEEE}
\end{bchapter}
\endbibitem

\bibitem[\protect\citeauthoryear{Dewri et~al.}{2011}]{dewri2011exploring}
\begin{barticle}
\bauthor{\bsnm{Dewri}, \binits{R.}},
\bauthor{\bsnm{Ray}, \binits{I.}},
\bauthor{\bsnm{Ray}, \binits{I.}},
\bauthor{\bsnm{Whitley}, \binits{D.}}:
\batitle{Exploring privacy versus data quality trade-offs in anonymization techniques using multi-objective optimization}.
\bjtitle{Journal of Computer Security}
\bvolume{19}(\bissue{5}),
\bfpage{935}--\blpage{974}
(\byear{2011})
\end{barticle}
\endbibitem

\bibitem[\protect\citeauthoryear{Lin and Xiao}{2024}]{lin2024exploring}
\begin{botherref}
\oauthor{\bsnm{Lin}, \binits{Y.}},
\oauthor{\bsnm{Xiao}, \binits{N.}}:
Exploring the tradeoff between privacy and utility of complete-count census data using a multiobjective optimization approach.
Geographical Analysis
(2024)
\end{botherref}
\endbibitem

\bibitem[\protect\citeauthoryear{Halawi et~al.}{2023}]{halawi2023multi}
\begin{barticle}
\bauthor{\bsnm{Halawi}, \binits{O.N.}},
\bauthor{\bsnm{Abu-Khzam}, \binits{F.N.}},
\bauthor{\bsnm{Thoumi}, \binits{S.}}:
\batitle{A multi-objective degree-based network anonymization method}.
\bjtitle{Algorithms}
\bvolume{16}(\bissue{9}),
\bfpage{436}
(\byear{2023})
\end{barticle}
\endbibitem

\bibitem[\protect\citeauthoryear{Sugitha}{2024}]{sugitha2024multi}
\begin{barticle}
\bauthor{\bsnm{Sugitha}, \binits{G.}}:
\batitle{A multi-objective privacy preservation model for cloud security using hunter prey optimization algorithm}.
\bjtitle{Peer-to-Peer Networking and Applications}
\bvolume{17}(\bissue{2}),
\bfpage{911}--\blpage{923}
(\byear{2024})
\end{barticle}
\endbibitem

\bibitem[\protect\citeauthoryear{Ahamad et~al.}{2022}]{ahamad2022multi}
\begin{barticle}
\bauthor{\bsnm{Ahamad}, \binits{D.}},
\bauthor{\bsnm{Hameed}, \binits{S.A.}},
\bauthor{\bsnm{Akhtar}, \binits{M.}}:
\batitle{A multi-objective privacy preservation model for cloud security using hybrid jaya-based shark smell optimization}.
\bjtitle{Journal of King Saud University-Computer and Information Sciences}
\bvolume{34}(\bissue{6}),
\bfpage{2343}--\blpage{2358}
(\byear{2022})
\end{barticle}
\endbibitem

\bibitem[\protect\citeauthoryear{Jahan et~al.}{2025}]{jahan2025analysis}
\begin{botherref}
\oauthor{\bsnm{Jahan}, \binits{S.}},
\oauthor{\bsnm{Ge}, \binits{Y.-F.}},
\oauthor{\bsnm{Kabir}, \binits{E.}},
\oauthor{\bsnm{Wang}, \binits{K.}}:
Analysis and multi-objective protection of public medical datasets from privacy and utility perspectives.
Data Science and Engineering,
1--14
(2025)
\end{botherref}
\endbibitem

\bibitem[\protect\citeauthoryear{Sadeghi-Nasab and Rahmani}{2025}]{sadeghi2025optimizing}
\begin{barticle}
\bauthor{\bsnm{Sadeghi-Nasab}, \binits{A.}},
\bauthor{\bsnm{Rahmani}, \binits{M.}}:
\batitle{Optimizing data privacy: an rfd-based approach to anonymization strategy selection}.
\bjtitle{The Journal of Supercomputing}
\bvolume{81}(\bissue{1}),
\bfpage{1}--\blpage{27}
(\byear{2025})
\end{barticle}
\endbibitem

\bibitem[\protect\citeauthoryear{Jahan et~al.}{2024}]{jahan2024dynamic}
\begin{bchapter}
\bauthor{\bsnm{Jahan}, \binits{S.}},
\bauthor{\bsnm{Ge}, \binits{Y.-F.}},
\bauthor{\bsnm{Wang}, \binits{H.}},
\bauthor{\bsnm{Kabir}, \binits{E.}}:
\bctitle{Dynamic-parameter genetic algorithm for multi-objective privacy-preserving trajectory data publishing}.
In: \bbtitle{International Conference on Web Information Systems Engineering},
pp. \bfpage{46}--\blpage{57}
(\byear{2024}).
\bcomment{Springer}
\end{bchapter}
\endbibitem

\bibitem[\protect\citeauthoryear{Oprescu et~al.}{2022}]{oprescu2022energy}
\begin{bchapter}
\bauthor{\bsnm{Oprescu}, \binits{A.}},
\bauthor{\bsnm{Misdorp}, \binits{S.}},
\bauthor{\bsnm{Elsen}, \binits{K.}}:
\bctitle{Energy cost and accuracy impact of $k$-anonymity}.
In: \bbtitle{2022 International Conference on ICT for Sustainability (ICT4S)},
pp. \bfpage{65}--\blpage{76}
(\byear{2022}).
\bcomment{IEEE}
\end{bchapter}
\endbibitem

\bibitem[\protect\citeauthoryear{Senavirathne and Torra}{2020}]{senavirathne2020role}
\begin{bchapter}
\bauthor{\bsnm{Senavirathne}, \binits{N.}},
\bauthor{\bsnm{Torra}, \binits{V.}}:
\bctitle{On the role of data anonymization in machine learning privacy}.
In: \bbtitle{2020 IEEE 19th International Conference on Trust, Security and Privacy in Computing and Communications (TrustCom)},
pp. \bfpage{664}--\blpage{675}
(\byear{2020}).
\bcomment{IEEE}
\end{bchapter}
\endbibitem

\bibitem[\protect\citeauthoryear{Pitoglou et~al.}{2022}]{pitoglou2022exploring}
\begin{barticle}
\bauthor{\bsnm{Pitoglou}, \binits{S.}},
\bauthor{\bsnm{Filntisi}, \binits{A.}},
\bauthor{\bsnm{Anastasiou}, \binits{A.}},
\bauthor{\bsnm{Matsopoulos}, \binits{G.K.}},
\bauthor{\bsnm{Koutsouris}, \binits{D.}}:
\batitle{Exploring the utility of anonymized ehr datasets in machine learning experiments in the context of the modelhealth project}.
\bjtitle{Applied Sciences}
\bvolume{12}(\bissue{12}),
\bfpage{5942}
(\byear{2022})
\end{barticle}
\endbibitem

\bibitem[\protect\citeauthoryear{Mauger et~al.}{2020}]{mauger2020multi}
\begin{bchapter}
\bauthor{\bsnm{Mauger}, \binits{C.}},
\bauthor{\bsnm{Mahec}, \binits{G.L.}},
\bauthor{\bsnm{Dequen}, \binits{G.}}:
\bctitle{Multi-criteria optimization using $l$-diversity and $t$-closeness for $k$-anonymization}.
In: \bbtitle{Data Privacy Management, Cryptocurrencies and Blockchain Technology: ESORICS 2020 International Workshops, DPM 2020 and CBT 2020, Guildford, UK, September 17--18, 2020, Revised Selected Papers 15},
pp. \bfpage{73}--\blpage{88}
(\byear{2020}).
\bcomment{Springer}
\end{bchapter}
\endbibitem

\bibitem[\protect\citeauthoryear{Domingo-Ferrer and Mateo-Sanz}{2002}]{domingo2002practical}
\begin{barticle}
\bauthor{\bsnm{Domingo-Ferrer}, \binits{J.}},
\bauthor{\bsnm{Mateo-Sanz}, \binits{J.M.}}:
\batitle{Practical data-oriented microaggregation for statistical disclosure control}.
\bjtitle{IEEE Transactions on Knowledge and Data Engineering}
\bvolume{14}(\bissue{1}),
\bfpage{189}--\blpage{201}
(\byear{2002})
\end{barticle}
\endbibitem

\bibitem[\protect\citeauthoryear{Zheng et~al.}{2019}]{zheng2019effective}
\begin{bchapter}
\bauthor{\bsnm{Zheng}, \binits{W.}},
\bauthor{\bsnm{Ma}, \binits{Y.}},
\bauthor{\bsnm{Wang}, \binits{Z.}},
\bauthor{\bsnm{Jia}, \binits{C.}},
\bauthor{\bsnm{Li}, \binits{P.}}:
\bctitle{Effective $l$-diversity anonymization algorithm based on improved clustering}.
In: \bbtitle{Cyberspace Safety and Security: 11th International Symposium, CSS 2019, Guangzhou, China, December 1--3, 2019, Proceedings, Part II 11},
pp. \bfpage{318}--\blpage{329}
(\byear{2019}).
\bcomment{Springer}
\end{bchapter}
\endbibitem

\bibitem[\protect\citeauthoryear{Doka et~al.}{2015}]{10.1145/2714576.2714590}
\begin{bchapter}
\bauthor{\bsnm{Doka}, \binits{K.}},
\bauthor{\bsnm{Xue}, \binits{M.}},
\bauthor{\bsnm{Tsoumakos}, \binits{D.}},
\bauthor{\bsnm{Karras}, \binits{P.}}:
\bctitle{k-anonymization by freeform generalization}.
In: \bbtitle{Proceedings of the 10th ACM Symposium on Information, Computer and Communications Security}.
\bsertitle{ASIA CCS '15},
pp. \bfpage{519}--\blpage{530}.
\bpublisher{Association for Computing Machinery},
\blocation{New York, NY, USA}
(\byear{2015}).
\doiurl{10.1145/2714576.2714590} .
\burl{https://doi.org/10.1145/2714576.2714590}
\end{bchapter}
\endbibitem

\bibitem[\protect\citeauthoryear{Mauger et~al.}{2020}]{10.1007/978-3-030-66172-4_5}
\begin{bchapter}
\bauthor{\bsnm{Mauger}, \binits{C.}},
\bauthor{\bsnm{Mahec}, \binits{G.L.}},
\bauthor{\bsnm{Dequen}, \binits{G.}}:
\bctitle{Multi-criteria optimization using l-diversity and t-closeness for k-anonymization}.
In: \beditor{\bsnm{Garcia-Alfaro}, \binits{J.}},
\beditor{\bsnm{Navarro-Arribas}, \binits{G.}},
\beditor{\bsnm{Herrera-Joancomarti}, \binits{J.}} (eds.)
\bbtitle{Data Privacy Management, Cryptocurrencies and Blockchain Technology},
pp. \bfpage{73}--\blpage{88}.
\bpublisher{Springer},
\blocation{Cham}
(\byear{2020})
\end{bchapter}
\endbibitem

\bibitem[\protect\citeauthoryear{Kennedy and Eberhart}{1995}]{kennedy1995particle}
\begin{bchapter}
\bauthor{\bsnm{Kennedy}, \binits{J.}},
\bauthor{\bsnm{Eberhart}, \binits{R.}}:
\bctitle{Particle swarm optimization}.
In: \bbtitle{Proceedings of ICNN'95-international Conference on Neural Networks},
vol. \bseriesno{4},
pp. \bfpage{1942}--\blpage{1948}
(\byear{1995}).
\bcomment{ieee}
\end{bchapter}
\endbibitem

\bibitem[\protect\citeauthoryear{Xu and Yu}{2018}]{XU201865}
\begin{barticle}
\bauthor{\bsnm{Xu}, \binits{G.}},
\bauthor{\bsnm{Yu}, \binits{G.}}:
\batitle{On convergence analysis of particle swarm optimization algorithm}.
\bjtitle{Journal of Computational and Applied Mathematics}
\bvolume{333},
\bfpage{65}--\blpage{73}
(\byear{2018})
\doiurl{10.1016/j.cam.2017.10.026}
\end{barticle}
\endbibitem

\bibitem[\protect\citeauthoryear{Hofmann}{1994}]{statlog_(german_credit_data)_144}
\begin{botherref}
\oauthor{\bsnm{Hofmann}, \binits{H.}}:
{Statlog (German Credit Data)}.
UCI Machine Learning Repository.
{DOI}: https://doi.org/10.24432/C5NC77
(1994)
\end{botherref}
\endbibitem

\bibitem[\protect\citeauthoryear{Becker and Kohavi}{1996}]{adult_2}
\begin{botherref}
\oauthor{\bsnm{Becker}, \binits{B.}},
\oauthor{\bsnm{Kohavi}, \binits{R.}}:
{Adult}.
UCI Machine Learning Repository.
{DOI}: https://doi.org/10.24432/C5XW20
(1996)
\end{botherref}
\endbibitem

\bibitem[\protect\citeauthoryear{El~Emam}{2013}]{ElEmamKhaled2013CMT}
\begin{bchapter}
\bauthor{\bsnm{El~Emam}, \binits{K.}}:
\bctitle{Choosing metric thresholds}.
In: \bbtitle{Guide to the De-Identification of Personal Health Information},
pp. \bfpage{242}--\blpage{251}.
\bpublisher{Auerbach Publications}, \blocation{???}
(\byear{2013})
\end{bchapter}
\endbibitem

\bibitem[\protect\citeauthoryear{Yan}{2016}]{yan2016rbayesianoptimization}
\begin{botherref}
\oauthor{\bsnm{Yan}, \binits{Y.}}:
rbayesianoptimization: bayesian optimization of hyperparameters.
R package version
\textbf{1}(0)
(2016)
\end{botherref}
\endbibitem

\bibitem[\protect\citeauthoryear{Fung et~al.}{2007}]{fung2007anonymizing}
\begin{barticle}
\bauthor{\bsnm{Fung}, \binits{B.C.}},
\bauthor{\bsnm{Wang}, \binits{K.}},
\bauthor{\bsnm{Philip}, \binits{S.Y.}}:
\batitle{Anonymizing classification data for privacy preservation}.
\bjtitle{IEEE Transactions on Knowledge and Data Engineering}
\bvolume{19}(\bissue{5}),
\bfpage{711}--\blpage{725}
(\byear{2007})
\end{barticle}
\endbibitem

\end{thebibliography}

\pagebreak
\begin{appendices}
\section{Hyperparameter Values}\label{app: hyperparameter values}
\begin{table}[!ht]
  \centering
  \caption{The anonymization models and the corresponding hyperparameter values applied in the context of this paper}
    \begin{tabular}{m{8em}m{9em}m{22em}}\hline
      Anonymization Model            & Dataset                         & Hyperparameters                 \\\hline
      \multirow{3}{*}{$k$-anonymity} & German credit                   & \multirow{3}{*}{$k=5,10,15,20$} \\
                                     & Adult                           &                                 \\
                                     & Sepsis patient                  &                                 \\\hline
      \multirow{6}{*}{Zheng et al.}  & \multirow{2}{*}{German credit}  & $k=5,10,15,20$,                 \\ & & $l=2,3,4$\\\cmidrule(lr){2-3}
                                     & \multirow{2}{*}{Adult}          & $k=5,10,15,20$,                 \\
                                     &                                 & $l=2,4,6,8,10,12,14$            \\\cmidrule(lr){2-3}
                                     & \multirow{2}{*}{Sepsis patient} & $k=5,10,15,20$,                 \\
                                     &                                 & $l=2$                           \\\hline
      \multirow{9}{*}{MO-OBAM}       & \multirow{3}{*}{German credit}  & $k=5,10,15,20$,                 \\&&$n_C=2,4,6,8,10,12,14,16,18,20,22,24,26,28,30$,\\&& $\lambda=0.0001,0.001,0.01,0.1,1$ \\\cmidrule(lr){2-3}
                                     & \multirow{3}{*}{Adult}          & $k=5,10,15,20$,                 \\ &&$n_C=4,10,20,30,40,50,60,70,80,90,100$,\\ && $\lambda=0.0001,0.001,0.01,0.1,1$ \\\cmidrule(lr){2-3}
                                     & \multirow{3}{*}{Sepsis patient} & $k=5,10,15,20$,                 \\ &&$n_C=4,10,20,30,40,820,1740,2310,3240$, \\ &&$\lambda=0.0001,0.001,0.01,0.1,1$ \\\hline
    \end{tabular}
  \label{tab: hyperparameter values}
\end{table}


\section{Sensitive Analysis}\label{app: sensitive analysis}
\subsection{Information Loss}
\begin{figure}[]
\centering
\subfloat[German credit]{
\includegraphics[scale=0.3]{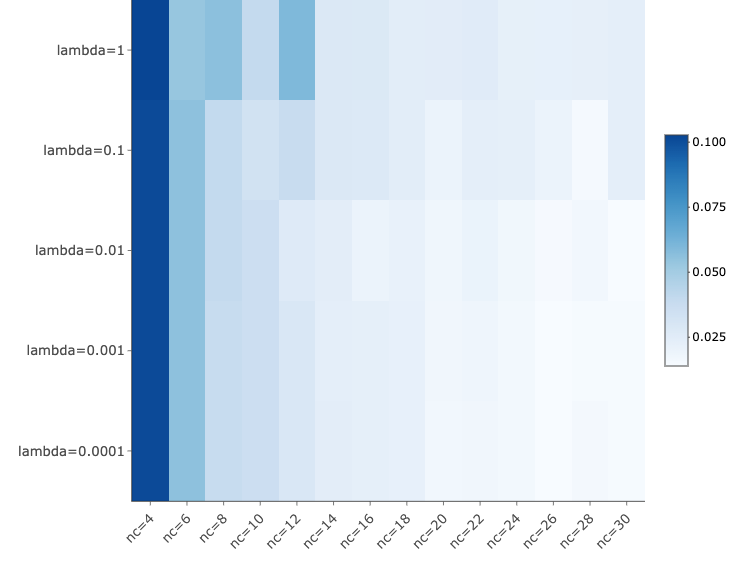}
\label{fig: German-credit-infoloss-our-model}}
\hfill
\subfloat[Adult]{
\includegraphics[scale=0.3]{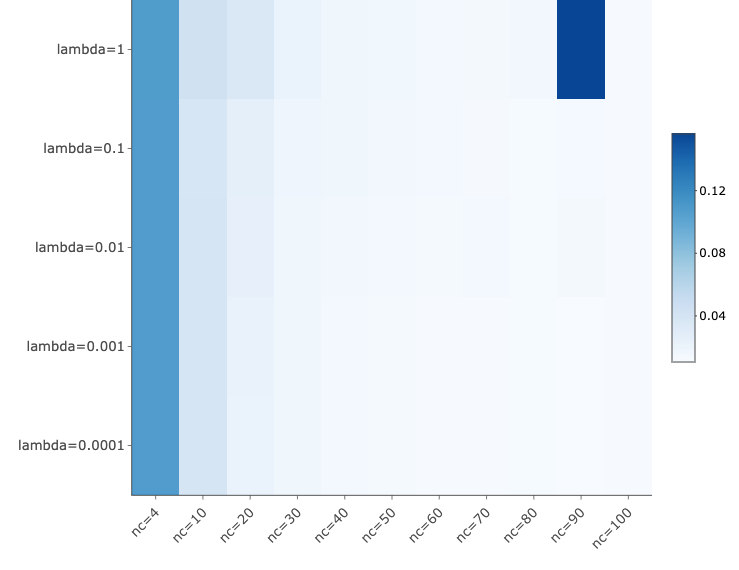}
\label{fig: Adult-infoloss-our-model}}
\hfil
\subfloat[Sepsis patient]{
\includegraphics[scale=0.3]{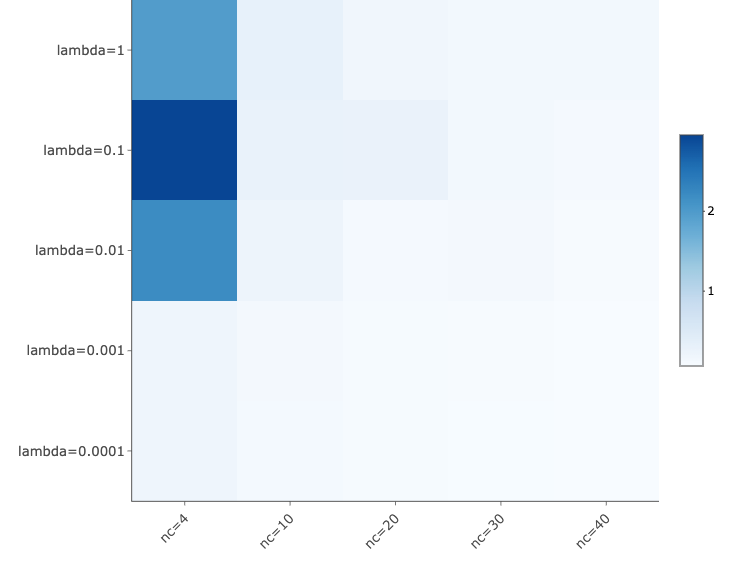}
\label{fig: Sepsis-infoloss-our-model}}
\caption{Information loss ($k=5$) caused by MO-OBAM}
\label{fig: infoloss Wei}
\end{figure}

Figure~\ref{fig: infoloss Wei} illustrates the relationship between information loss, $n_C$, and $\lambda$ for $k=5$. The x-axis represents the values of $n_C$, while the y-axis depicts the values of $\lambda$. Observing the figure, it becomes apparent that as $n_C$ increases while holding $\lambda$ constant, there is a consistent decrease in information loss across all three datasets. Similarly, when $n_C$ is fixed, increasing $\lambda$ results in higher information loss. This trend remains consistent across different values of $k$.

\subsection{Protection against Linkage Attacks}
Table~\ref{tab: German credit LA (our model)},~\ref{tab: Adult LA (our model)},~\ref{tab: Sepsis LA (our model)} demonstrate that our model offers sufficient protection against linkage attacks with a smaller number of clusters. As the number of clusters increases, fewer individuals belong to the same class, thereby escalating the risk of linkage attacks, especially when accompanied by a large value of $\lambda$.

\begin{table*}[tbp]
  \caption{The number of people s.t linkage attacks in the German credit dataset after anonymizing by our model. The third row is the number of people s.t linkage attacks in the original dataset}
  \resizebox{\textwidth}{!}{%
\begin{tabular}{lllllllllllllllllllllllllll}
  \hline
  \multicolumn{27}{c}{German Credit}                                                                                                                                                                                                                                                                                                                                                                         \\ \hline
  $n_C$                & $\lambda$               & $k$ & $\tau$=0.05 & $\tau$=0.075 & $\tau$=0.1 &  & $n_C$                & $\lambda$               & $k$ & $\tau$=0.05 & $\tau$=0.075 & $\tau$=0.1 &  & $n_C$                & $\lambda$               & $k$ & $\tau$=0.05 & $\tau$=0.075 & $\tau$=0.1 &  & $n_C$                & $\lambda$               & $k$ & $\tau$=0.05 & $\tau$=0.075 & $\tau$=0.1 \\ \cmidrule(lr){1-6} \cmidrule(lr){8-13} \cmidrule(lr){15-20} \cmidrule(lr){22-27} 
  310                  &                         &     & 959         & 828          & 698        &  & 310                  &                         &     & 959         & 828          & 698        &  & 310                  &                         &     & 959         & 828          & 698        &  & 310                  &                         &     & 959         & 828          & 698        \\ \cmidrule(lr){1-6} \cmidrule(lr){8-13} \cmidrule(lr){15-20} \cmidrule(lr){22-27} 
  \multirow{20}{*}{4}  & \multirow{4}{*}{0.0001} & 5   & 0           & 0            & 0          &  & \multirow{20}{*}{6}  & \multirow{4}{*}{0.0001} & 5   & 0           & 0            & 0          &  & \multirow{20}{*}{8}  & \multirow{4}{*}{0.0001} & 5   & 0           & 0            & 0          &  & \multirow{20}{*}{10} & \multirow{4}{*}{0.0001} & 5   & 0           & 0            & 0          \\
\hline
  \end{tabular}%
  }
  \label{tab: German credit LA (our model)}
  \end{table*}

\begin{table*}[tbp]
  \caption{The number of people s.t linkage attacks in the Adult dataset after anonymizing by our model. The third row is the number of people s.t linkage attacks in the original dataset}
  \resizebox{\textwidth}{!}{%
  \begin{tabular}{lllllllllllllllllllllllllll}
  \hline
  \multicolumn{27}{c}{Adult}                                                                                                                                                                                                                                                                                                                                                                                  \\ \hline
  $n_C$                & $\lambda$               & $k$ & $\tau$=0.05 & $\tau$=0.075 & $\tau$=0.1 &  & $n_C$                & $\lambda$               & $k$ & $\tau$=0.05 & $\tau$=0.075 & $\tau$=0.1 &  & $n_C$                 & $\lambda$               & $k$ & $\tau$=0.05 & $\tau$=0.075 & $\tau$=0.1 &  & $n_C$                & $\lambda$               & $k$ & $\tau$=0.05 & $\tau$=0.075 & $\tau$=0.1 \\ \cmidrule(lr){1-6} \cmidrule(lr){8-13} \cmidrule(lr){15-20} \cmidrule(lr){22-27} 
  1900                 &                         &     & 6506        & 4906         & 3910       &  & 1900                 &                         &     & 6506        & 4906         & 3910       &  & 1900                  &                         &     & 6506        & 4906         & 3910       &  & 1900                 &                         &     & 6506        & 4906         & 3910       \\ \cmidrule(lr){1-6} \cmidrule(lr){8-13} \cmidrule(lr){15-20} \cmidrule(lr){22-27} 
  \multirow{20}{*}{4}  & \multirow{4}{*}{0.0001} & 5   & 0           & 0            & 0          &  & \multirow{20}{*}{10} & \multirow{4}{*}{0.0001} & 5   & 0           & 0            & 0          &  & \multirow{20}{*}{20}  & \multirow{4}{*}{0.0001} & 5   & 18          & 0            & 0          &  & \multirow{20}{*}{30} & \multirow{4}{*}{0.0001} & 5   & 0           & 0            & 0          \\
\hline
  \end{tabular}%
  }
  \label{tab: Adult LA (our model)}
  \end{table*}

\begin{table*}[tbp]
  \caption{The number of people s.t linkage attacks in Sepsis patient dataset after anonymizing by our model. The third row is the number of people s.t linkage attacks in the original dataset}
  \centering
  \resizebox{\textwidth}{!}{%
\begin{tabular}{lllllllllllllllllllllllllll}
\hline
\multicolumn{27}{c}{Adult}                                                                                                                                                                                                                                                                                                                                                                                  \\ \hline
$n_C$                & $\lambda$               & $k$ & $\tau$=0.05 & $\tau$=0.075 & $\tau$=0.1 &  & $n_C$                & $\lambda$               & $k$ & $\tau$=0.05 & $\tau$=0.075 & $\tau$=0.1 &  & $n_C$                 & $\lambda$               & $k$ & $\tau$=0.05 & $\tau$=0.075 & $\tau$=0.1 &  & $n_C$                & $\lambda$               & $k$ & $\tau$=0.05 & $\tau$=0.075 & $\tau$=0.1 \\ \cmidrule(lr){1-6} \cmidrule(lr){8-13} \cmidrule(lr){15-20} \cmidrule(lr){22-27} 
1900                 &                         &     & 6506        & 4906         & 3910       &  & 1900                 &                         &     & 6506        & 4906         & 3910       &  & 1900                  &                         &     & 6506        & 4906         & 3910       &  & 1900                 &                         &     & 6506        & 4906         & 3910       \\ \cmidrule(lr){1-6} \cmidrule(lr){8-13} \cmidrule(lr){15-20} \cmidrule(lr){22-27} 
\multirow{20}{*}{4}  & \multirow{4}{*}{0.0001} & 5   & 0           & 0            & 0          &  & \multirow{20}{*}{10} & \multirow{4}{*}{0.0001} & 5   & 0           & 0            & 0          &  & \multirow{20}{*}{20}  & \multirow{4}{*}{0.0001} & 5   & 18          & 0            & 0          &  & \multirow{20}{*}{30} & \multirow{4}{*}{0.0001} & 5   & 0           & 0            & 0          \\
\hline
\end{tabular}%
}
  \label{tab: Sepsis LA (our model)}
\end{table*}

\subsection{Protection against Homogeneity Attacks}
Table~\ref{tab: German credit HA (our model)},~\ref{tab: Adult HA (our model)}, and~\ref{tab: Sepsis HA (our model)} underscore the effectiveness of our model in mitigating homogeneity attacks. However, when $\lambda$ is set to a small value, indicative of prioritizing the minimization of the objective function over information loss, certain individuals remain vulnerable to homogeneity attacks, particularly evident in scenarios with larger values of $n_C$. This highlights the delicate balance between minimizing information loss and safeguarding against homogeneity attacks, necessitating careful consideration of the interplay between $\lambda$, the number of clusters, and the underlying risk of privacy breaches.

\begin{table*}
  \caption{The number of people s.t homogeneity attacks in the German credit dataset after anonymizing by our model. HA stands for homogeneity attacks, and the third row is the number of people s.t homogeneity attacks in the original dataset}
  \centering
  \resizebox{\linewidth}{!}{
\begin{tabular}{cccccccccccccccccccccccc}\hline
      \multicolumn{24}{c}{German Credit}                                                                                                                                                                                                                                                                                     \\\midrule
      $n_C$                & $\lambda$               & $k$ & HA  &  &
      $n_C$                & $\lambda$               & $k$ & HA  &  &
      $n_C$                & $\lambda$               & $k$ & HA  &  &
      $n_C$                & $\lambda$               & $k$ & HA  &  &
      $n_C$                & $\lambda$               & $k$ & HA                                                                                                                                                                                                                                                              \\ \cmidrule(lr){1-4} \cmidrule(lr){6-9} \cmidrule(lr){11-14} \cmidrule(lr){16-19} \cmidrule(lr){21-24}
      1900                 &                         &     & 634 &  & 1900                 &                         &    & 634 &  & 1900                 &                         &    & 634 &  & 1900                 &                         &    & 634 &  & 1900                 &                         &    & 634 \\ \cmidrule(lr){1-4} \cmidrule(lr){6-9} \cmidrule(lr){11-14} \cmidrule(lr){16-19} \cmidrule(lr){21-24}
      \multirow{20}{*}{4}  & \multirow{4}{*}{0.0001} & 5   & 0   &  & \multirow{20}{*}{6}  & \multirow{4}{*}{0.0001} & 5  & 0   &  & \multirow{20}{*}{8}  & \multirow{4}{*}{0.0001} & 5  & 0   &  & \multirow{20}{*}{10} & \multirow{4}{*}{0.0001} & 5  & 0   &  & \multirow{20}{*}{12} & \multirow{4}{*}{0.0001} & 5  & 0   \\
\hline
    \end{tabular}
  }
  \label{tab: German credit HA (our model)}
\end{table*}

\begin{table*}
  \caption{The number of people s.t homogeneity attacks in the Adult dataset after anonymizing by our model. HA stands for homogeneity attacks, and the third row is the number of people s.t homogeneity attacks in the original dataset}
  \centering
  \resizebox{0.859\linewidth}{!}{
    \begin{tabular}{ccccccccccccccccccc}\hline
      \multicolumn{19}{c}{Adult}                                                                                                                                                                                                                               \\\midrule
      $n_C$                & $\lambda$               & $k$ & HA  &  &
      $n_C$                & $\lambda$               & $k$ & HA  &  &
      $n_C$                & $\lambda$               & $k$ & HA  &  &
      $n_C$                & $\lambda$               & $k$ & HA                                                                                                                                                                                                \\ \cmidrule(lr){1-4} \cmidrule(lr){6-9} \cmidrule(lr){11-14} \cmidrule(lr){16-19}
      1900                 &                         &     & 634 &  & 1900                 &                         &    & 634 &  & 1900                  &                         &    & 634 &  & 1900                 &                         &    & 634 \\ \cmidrule(lr){1-4} \cmidrule(lr){6-9} \cmidrule(lr){11-14} \cmidrule(lr){16-19}
      \multirow{20}{*}{4}  & \multirow{4}{*}{0.0001} & 5   & 0   &  & \multirow{20}{*}{10} & \multirow{4}{*}{0.0001} & 5  & 0   &  & \multirow{20}{*}{20}  & \multirow{4}{*}{0.0001} & 5  & 0   &  & \multirow{20}{*}{30} & \multirow{4}{*}{0.0001} & 5  & 0   \\ \cmidrule(lr){3-4} \cmidrule(lr){8-9} \cmidrule(lr){13-14} \cmidrule(lr){18-19}
                           &                         & 10  & 0   &  &                      &                         & 10 & 0   &  &                       &                         & 10 & 0   &  &                      &                         & 10 & 0   \\
\hline
    \end{tabular}
  }
  \label{tab: Adult HA (our model)}
\end{table*}

\begin{table*}
  \caption{The number of people s.t homogeneity attacks in Sepsis patient dataset after anonymizing by our model HA stands for homogeneity attacks, and the third row is the number of people s.t homogeneity attacks in the original dataset}
  \centering 
    \begin{tabular}{cccccccccccccc}\hline
      \multicolumn{14}{c}{Sepsis Patient}                                                                                                                                                            \\\midrule
      $n_C$                & $\lambda$               & $k$ & HA    &  &
      $n_C$                & $\lambda$               & $k$ & HA    &  &
      $n_C$                & $\lambda$               & $k$ & HA
      \\ \cmidrule(lr){1-4} \cmidrule(lr){6-9} \cmidrule(lr){11-14}
      23553                &                         &     & 14937 &  & 23553                &                         &    & 14937 &  & 23553                &                         &    & 14937 \\ \cmidrule(lr){1-4} \cmidrule(lr){6-9} \cmidrule(lr){11-14}
      \multirow{20}{*}{4}  & \multirow{4}{*}{0.0001} & 5   & 0     &  & \multirow{20}{*}{10} & \multirow{4}{*}{0.0001} & 5  & 0     &  & \multirow{20}{*}{20} & \multirow{4}{*}{0.0001} & 5  & 0     \\ \cmidrule(lr){3-4} \cmidrule(lr){8-9} \cmidrule(lr){13-14}
                           &                         & 10  & 0     &  &                      &                         & 10 & 0     &  &                      &                         & 10 & 0     \\ \cmidrule(lr){3-4} \cmidrule(lr){8-9} \cmidrule(lr){13-14}
                           &                         & 15  & 0     &  &                      &                         & 15 & 0     &  &                      &                         & 15 & 0     \\ \cmidrule(lr){3-4} \cmidrule(lr){8-9} \cmidrule(lr){13-14}
                           &                         & 20  & 0     &  &                      &                         & 20 & 0     &  &                      &                         & 20 & 0     \\
\hline
    \end{tabular}
  
  \label{tab: Sepsis HA (our model)}
\end{table*}

\section{Feature Importance}\label{app:feature import}

  \subsection{German Credit}
  Table~\ref{tab: German credit FI} presents the feature importance of Decision Trees using the German credit dataset. Table~\ref{tab: German credit FI}(a) shows the changes in feature importance after applying the $k$-anonymity algorithm. Table~\ref{tab: German credit FI}(b) displays the changes in feature importance after applying the algorithm proposed by Zheng et al. Finally, Table~\ref{tab: German credit FI}(c) illustrates the changes in feature importance after applying our model.

  \begin{table*}[!ht]
    \centering
    \caption{Variation of feature importances in the German credit dataset with different anonymization models. Cells highlighted with color coding denote QIs.}
    \label{tab: German credit FI}

    \label{tab: German credit FI k-anonymity}
  \end{table*}

  \begin{table*}[!ht]
    \label{tab: German credit FI Zheng et al}
    \centering
    %
                          &
          0.00\%
        \end{tabular}%
      }
    \end{tabular}
    \label{tab: German credit FI our model}
  \end{table*}

  \subsection{Adult}
  Table~\ref{tab: Adult FI} presents the feature importance of Decision Trees using the Adult dataset. Table~\ref{tab: Adult FI}(a) shows the changes in feature importance after applying the $k$-anonymity algorithm. Table~\ref{tab: Adult FI}(b) displays the changes in feature importance after applying the algorithm proposed by Zheng et al. Finally, Table~\ref{tab: Adult FI}(c) illustrates the changes in feature importance after applying our model.

  \begin{table*}[!ht]
    \centering
    \caption{Variation of feature importances in the Adult dataset with different anonymization models. Cells highlighted with color coding denote QIs.}
    \label{tab: Adult FI}
                        &
          7.23\%                                                                               \\
          occupation                                                                         &
          6.30\%                                                                             &
          \cellcolor[HTML]{83CCEB}age                                                        &
          \multicolumn{1}{c|}{5.08\%}                                                        &
          \cellcolor[HTML]{83CCEB}age                                                        &
          \multicolumn{1}{c|}{5.25\%}                                                        &
          \cellcolor[HTML]{83CCEB}age                                                        &
          \multicolumn{1}{c|}{5.47\%}                                                        &
          \cellcolor[HTML]{83CCEB}age                                                        &
          \multicolumn{1}{c|}{5.66\%}                                                        &
          occupation                                                                         &
          \multicolumn{1}{c|}{5.99\%}                                                        &
          occupation                                                                         &
          \multicolumn{1}{c|}{6.20\%}                                                        &
          occupation                                                                         &
          \multicolumn{1}{c|}{5.58\%}                                                        &
          occupation                                                                         &
          5.46\%                                                                               \\
          capital\_loss                                                                      &
          3.68\%                                                                             &
          workclass                                                                          &
          \multicolumn{1}{c|}{3.79\%}                                                        &
          capital\_loss                                                                      &
          \multicolumn{1}{c|}{3.89\%}                                                        &
          workclass                                                                          &
          \multicolumn{1}{c|}{3.87\%}                                                        &
          capital\_loss                                                                      &
          \multicolumn{1}{c|}{3.88\%}                                                        &
          capital\_loss                                                                      &
          \multicolumn{1}{c|}{3.83\%}                                                        &
          capital\_loss                                                                      &
          \multicolumn{1}{c|}{3.79\%}                                                        &
          capital\_loss                                                                      &
          \multicolumn{1}{c|}{3.75\%}                                                        &
          capital\_loss                                                                      &
          3.70\%                                                                               \\
          workclass                                                                          &
          3.34\%                                                                             &
          capital\_loss                                                                      &
          \multicolumn{1}{c|}{3.71\%}                                                        &
          workclass                                                                          &
          \multicolumn{1}{c|}{3.75\%}                                                        &
          capital\_loss                                                                      &
          \multicolumn{1}{c|}{3.59\%}                                                        &
          workclass                                                                          &
          \multicolumn{1}{c|}{3.62\%}                                                        &
          workclass                                                                          &
          \multicolumn{1}{c|}{3.01\%}                                                        &
          workclass                                                                          &
          \multicolumn{1}{c|}{3.07\%}                                                        &
          workclass                                                                          &
          \multicolumn{1}{c|}{3.31\%}                                                        &
          workclass                                                                          &
          2.90\%                                                                               \\
          \cellcolor[HTML]{83E28E}race                                                       &
          1.21\%                                                                             &
          \begin{tabular}[c]{@{}l@{}}native\\ \_country\end{tabular}                         &
          \multicolumn{1}{c|}{1.52\%}                                                        &
          \begin{tabular}[c]{@{}l@{}}native\\ \_country\end{tabular}                         &
          \multicolumn{1}{c|}{1.62\%}                                                        &
          \begin{tabular}[c]{@{}l@{}}native\\ \_country\end{tabular}                         &
          \multicolumn{1}{c|}{1.68\%}                                                        &
          \begin{tabular}[c]{@{}l@{}}native\\ \_country\end{tabular}                         &
          \multicolumn{1}{c|}{1.58\%}                                                        &
          \cellcolor[HTML]{83E28E}race                                                       &
          \multicolumn{1}{c|}{1.40\%}                                                        &
          \begin{tabular}[c]{@{}l@{}}native\\ \_country\end{tabular}                         &
          \multicolumn{1}{c|}{1.46\%}                                                        &
          \begin{tabular}[c]{@{}l@{}}native\\ \_country\end{tabular}                         &
          \multicolumn{1}{c|}{1.48\%}                                                        &
          \cellcolor[HTML]{83E28E}race                                                       &
          1.61\%                                                                               \\
          \begin{tabular}[c]{@{}l@{}}native\\ \_country\end{tabular}                         &
          1.16\%                                                                             &
          education                                                                          &
          \multicolumn{1}{c|}{1.52\%}                                                        &
          education                                                                          &
          \multicolumn{1}{c|}{1.51\%}                                                        &
          education                                                                          &
          \multicolumn{1}{c|}{1.39\%}                                                        &
          education                                                                          &
          \multicolumn{1}{c|}{1.37\%}                                                        &
          education                                                                          &
          \multicolumn{1}{c|}{1.30\%}                                                        &
          \cellcolor[HTML]{83E28E}race                                                       &
          \multicolumn{1}{c|}{1.34\%}                                                        &
          education                                                                          &
          \multicolumn{1}{c|}{1.43\%}                                                        &
          \begin{tabular}[c]{@{}l@{}}native\\ \_country\end{tabular}                         &
          1.35\%                                                                               \\
          education                                                                          &
          1.03\%                                                                             &
          \cellcolor[HTML]{83E28E}race                                                       &
          \multicolumn{1}{c|}{1.08\%}                                                        &
          \cellcolor[HTML]{E49EDD}\begin{tabular}[c]{@{}l@{}}marital\\ \_status\end{tabular} &
          \multicolumn{1}{c|}{1.24\%}                                                        &
          \cellcolor[HTML]{E49EDD}\begin{tabular}[c]{@{}l@{}}marital\\ \_status\end{tabular} &
          \multicolumn{1}{c|}{1.15\%}                                                        &
          \cellcolor[HTML]{E49EDD}\begin{tabular}[c]{@{}l@{}}marital\\ \_status\end{tabular} &
          \multicolumn{1}{c|}{1.31\%}                                                        &
          \begin{tabular}[c]{@{}l@{}}native\\ \_country\end{tabular}                         &
          \multicolumn{1}{c|}{1.29\%}                                                        &
          education                                                                          &
          \multicolumn{1}{c|}{1.26\%}                                                        &
          \cellcolor[HTML]{83E28E}race                                                       &
          \multicolumn{1}{c|}{1.28\%}                                                        &
          education                                                                          &
          1.26\%                                                                               \\
          \cellcolor[HTML]{E49EDD}\begin{tabular}[c]{@{}l@{}}marital\\ \_status\end{tabular} &
          0.77\%                                                                             &
          \cellcolor[HTML]{E49EDD}\begin{tabular}[c]{@{}l@{}}marital\\ \_status\end{tabular} &
          \multicolumn{1}{c|}{1.00\%}                                                        &
          \cellcolor[HTML]{83E28E}race                                                       &
          \multicolumn{1}{c|}{0.88\%}                                                        &
          \cellcolor[HTML]{83E28E}race                                                       &
          \multicolumn{1}{c|}{0.79\%}                                                        &
          \cellcolor[HTML]{83E28E}race                                                       &
          \multicolumn{1}{c|}{0.85\%}                                                        &
          \cellcolor[HTML]{E49EDD}\begin{tabular}[c]{@{}l@{}}marital\\ \_status\end{tabular} &
          \multicolumn{1}{c|}{0.94\%}                                                        &
          \cellcolor[HTML]{E49EDD}\begin{tabular}[c]{@{}l@{}}marital\\ \_status\end{tabular} &
          \multicolumn{1}{c|}{1.12\%}                                                        &
          \cellcolor[HTML]{E49EDD}\begin{tabular}[c]{@{}l@{}}marital\\ \_status\end{tabular} &
          \multicolumn{1}{c|}{1.09\%}                                                        &
          \cellcolor[HTML]{E49EDD}\begin{tabular}[c]{@{}l@{}}marital\\ \_status\end{tabular} &
          1.26\%                                                                               \\
          \cellcolor[HTML]{F1A983}sex                                                        &
          0.39\%                                                                             &
          \cellcolor[HTML]{F1A983}sex                                                        &
          \multicolumn{1}{c|}{0.67\%}                                                        &
          \cellcolor[HTML]{F1A983}sex                                                        &
          \multicolumn{1}{c|}{0.49\%}                                                        &
          \cellcolor[HTML]{F1A983}sex                                                        &
          \multicolumn{1}{c|}{0.33\%}                                                        &
          \cellcolor[HTML]{F1A983}sex                                                        &
          \multicolumn{1}{c|}{0.34\%}                                                        &
          \cellcolor[HTML]{F1A983}sex                                                        &
          \multicolumn{1}{c|}{0.41\%}                                                        &
          \cellcolor[HTML]{F1A983}sex                                                        &
          \multicolumn{1}{c|}{0.60\%}                                                        &
          \cellcolor[HTML]{F1A983}sex                                                        &
          \multicolumn{1}{c|}{0.73\%}                                                        &
          \cellcolor[HTML]{F1A983}sex                                                        &
          0.68\%
        \end{tabular}%
      }
    \end{tabular}
  \end{table*}

  \begin{table*}[!ht]
    \centering
    \begin{tabular}{c}
      (c) MO-OBAM \\
      \resizebox{\textwidth}{!}{%
        \begin{tabular}{m{1.5cm}c|m{1.5cm}cm{1.5cm}cm{1.5cm}cm{1.5cm}cm{1.5cm}cm{1.5cm}cm{1.5cm}cm{1.5cm}c}
          \hline
          \multicolumn{2}{c|}{Original   Data}                                               &
          \multicolumn{16}{c}{MO-OBAM}                                                         \\ \hline
          Feature                                                                            &
          Importance                                                                         &
          \begin{tabular}[c]{@{}l@{}}$n_C$=4,\\ $\lambda$=1,\\ $k$=5\end{tabular}            &
          \multicolumn{1}{c|}{Importance}                                                    &
          \begin{tabular}[c]{@{}l@{}}$n_C$=4,\\ $\lambda$=1,\\ $k$=10\end{tabular}           &
          \multicolumn{1}{c|}{Importance}                                                    &
          \begin{tabular}[c]{@{}l@{}}$n_C$=4,\\ $\lambda$=1,\\ $k$=15\end{tabular}           &
          \multicolumn{1}{c|}{Importance}                                                    &
          \begin{tabular}[c]{@{}l@{}}$n_C$=4,\\ $\lambda$=1,\\ $k$=20\end{tabular}           &
          \multicolumn{1}{c|}{Importance}                                                    &
          \begin{tabular}[c]{@{}l@{}}$n_C$=100,\\ $\lambda$=1e-04,\\ $k$=5\end{tabular}      &
          \multicolumn{1}{c|}{Importance}                                                    &
          \begin{tabular}[c]{@{}l@{}}$n_C$=100,\\ $\lambda$=1e-04,\\ $k$=10\end{tabular}     &
          \multicolumn{1}{c|}{Importance}                                                    &
          \begin{tabular}[c]{@{}l@{}}$n_C$=100,\\ $\lambda$=1e-04,\\ $k$=15\end{tabular}     &
          \multicolumn{1}{c|}{Importance}                                                    &
          \begin{tabular}[c]{@{}l@{}}$n_C$=100,\\ $\lambda$=1e-04,\\ $k$=20\end{tabular}     &
          Importance                                                                           \\ \hline
          fnlwgt                                                                             &
          21.90\%                                                                            &
          fnlwgt                                                                             &
          \multicolumn{1}{c|}{29.86\%}                                                       &
          fnlwgt                                                                             &
          \multicolumn{1}{c|}{29.91\%}                                                       &
          fnlwgt                                                                             &
          \multicolumn{1}{c|}{29.91\%}                                                       &
          fnlwgt                                                                             &
          \multicolumn{1}{c|}{29.69\%}                                                       &
          fnlwgt                                                                             &
          \multicolumn{1}{c|}{22.36\%}                                                       &
          fnlwgt                                                                             &
          \multicolumn{1}{c|}{22.58\%}                                                       &
          fnlwgt                                                                             &
          \multicolumn{1}{c|}{22.94\%}                                                       &
          fnlwgt                                                                             &
          23.23\%                                                                              \\
          relationship                                                                       &
          19.89\%                                                                            &
          relationship                                                                       &
          \multicolumn{1}{c|}{19.47\%}                                                       &
          relationship                                                                       &
          \multicolumn{1}{c|}{19.69\%}                                                       &
          relationship                                                                       &
          \multicolumn{1}{c|}{19.77\%}                                                       &
          relationship                                                                       &
          \multicolumn{1}{c|}{19.47\%}                                                       &
          relationship                                                                       &
          \multicolumn{1}{c|}{19.77\%}                                                       &
          relationship                                                                       &
          \multicolumn{1}{c|}{19.77\%}                                                       &
          relationship                                                                       &
          \multicolumn{1}{c|}{19.86\%}                                                       &
          relationship                                                                       &
          19.68\%                                                                              \\
          \cellcolor[HTML]{83CCEB}age                                                        &
          13.09\%                                                                            &
          \begin{tabular}[c]{@{}l@{}}capital\_\\ gain\end{tabular}                           &
          \multicolumn{1}{c|}{11.02\%}                                                       &
          \begin{tabular}[c]{@{}l@{}}capital\_\\ gain\end{tabular}                           &
          \multicolumn{1}{c|}{10.90\%}                                                       &
          \begin{tabular}[c]{@{}l@{}}capital\_\\ gain\end{tabular}                           &
          \multicolumn{1}{c|}{10.40\%}                                                       &
          \begin{tabular}[c]{@{}l@{}}capital\_\\ gain\end{tabular}                           &
          \multicolumn{1}{c|}{10.57\%}                                                       &
          \begin{tabular}[c]{@{}l@{}}capital\_\\ gain\end{tabular}                           &
          \multicolumn{1}{c|}{10.55\%}                                                       &
          \begin{tabular}[c]{@{}l@{}}capital\_\\ gain\end{tabular}                           &
          \multicolumn{1}{c|}{10.78\%}                                                       &
          \begin{tabular}[c]{@{}l@{}}capital\_\\ gain\end{tabular}                           &
          \multicolumn{1}{c|}{10.35\%}                                                       &
          \begin{tabular}[c]{@{}l@{}}capital\_\\ gain\end{tabular}                           &
          10.43\%                                                                              \\
          capital\_gain                                                                      &
          10.38\%                                                                            &
          \begin{tabular}[c]{@{}l@{}}education\_\\ level\end{tabular}                        &
          \multicolumn{1}{c|}{10.22\%}                                                       &
          \begin{tabular}[c]{@{}l@{}}education\_\\ level\end{tabular}                        &
          \multicolumn{1}{c|}{9.87\%}                                                        &
          \begin{tabular}[c]{@{}l@{}}education\_\\ level\end{tabular}                        &
          \multicolumn{1}{c|}{10.32\%}                                                       &
          \begin{tabular}[c]{@{}l@{}}education\_\\ level\end{tabular}                        &
          \multicolumn{1}{c|}{10.23\%}                                                       &
          \begin{tabular}[c]{@{}l@{}}education\_\\ level\end{tabular}                        &
          \multicolumn{1}{c|}{10.26\%}                                                       &
          \begin{tabular}[c]{@{}l@{}}education\_\\ level\end{tabular}                        &
          \multicolumn{1}{c|}{10.61\%}                                                       &
          \begin{tabular}[c]{@{}l@{}}education\_\\ level\end{tabular}                        &
          \multicolumn{1}{c|}{10.03\%}                                                       &
          \begin{tabular}[c]{@{}l@{}}education\_\\ level\end{tabular}                        &
          10.12\%                                                                              \\
          \begin{tabular}[c]{@{}l@{}}education\_\\ level\end{tabular}                        &
          10.14\%                                                                            &
          \begin{tabular}[c]{@{}l@{}}hours\_\\ per\_week\end{tabular}                        &
          \multicolumn{1}{c|}{7.97\%}                                                        &
          \begin{tabular}[c]{@{}l@{}}hours\_\\ per\_week\end{tabular}                        &
          \multicolumn{1}{c|}{8.40\%}                                                        &
          \begin{tabular}[c]{@{}l@{}}hours\_\\ per\_week\end{tabular}                        &
          \multicolumn{1}{c|}{7.89\%}                                                        &
          \begin{tabular}[c]{@{}l@{}}hours\_\\ per\_week\end{tabular}                        &
          \multicolumn{1}{c|}{7.99\%}                                                        &
          \cellcolor[HTML]{83CCEB}age                                                        &
          \multicolumn{1}{c|}{8.04\%}                                                        &
          \cellcolor[HTML]{83CCEB}age                                                        &
          \multicolumn{1}{c|}{7.98\%}                                                        &
          \cellcolor[HTML]{83CCEB}age                                                        &
          \multicolumn{1}{c|}{8.24\%}                                                        &
          \cellcolor[HTML]{83CCEB}age                                                        &
          8.26\%                                                                               \\
          \begin{tabular}[c]{@{}l@{}}hours\_\\ per\_week\end{tabular}                        &
          6.71\%                                                                             &
          occupation                                                                         &
          \multicolumn{1}{c|}{6.39\%}                                                        &
          occupation                                                                         &
          \multicolumn{1}{c|}{6.03\%}                                                        &
          occupation                                                                         &
          \multicolumn{1}{c|}{6.43\%}                                                        &
          occupation                                                                         &
          \multicolumn{1}{c|}{6.72\%}                                                        &
          \begin{tabular}[c]{@{}l@{}}hours\_\\ per\_week\end{tabular}                        &
          \multicolumn{1}{c|}{7.09\%}                                                        &
          \begin{tabular}[c]{@{}l@{}}hours\_\\ per\_week\end{tabular}                        &
          \multicolumn{1}{c|}{7.28\%}                                                        &
          \begin{tabular}[c]{@{}l@{}}hours\_\\ per\_week\end{tabular}                        &
          \multicolumn{1}{c|}{7.22\%}                                                        &
          \begin{tabular}[c]{@{}l@{}}hours\_\\ per\_week\end{tabular}                        &
          7.65\%                                                                               \\
          occupation                                                                         &
          6.30\%                                                                             &
          capital\_loss                                                                      &
          \multicolumn{1}{c|}{3.76\%}                                                        &
          capital\_loss                                                                      &
          \multicolumn{1}{c|}{3.73\%}                                                        &
          workclass                                                                          &
          \multicolumn{1}{c|}{3.68\%}                                                        &
          capital\_loss                                                                      &
          \multicolumn{1}{c|}{3.79\%}                                                        &
          occupation                                                                         &
          \multicolumn{1}{c|}{5.97\%}                                                        &
          occupation                                                                         &
          \multicolumn{1}{c|}{5.22\%}                                                        &
          occupation                                                                         &
          \multicolumn{1}{c|}{5.99\%}                                                        &
          occupation                                                                         &
          5.68\%                                                                               \\
          capital\_loss                                                                      &
          3.68\%                                                                             &
          workclass                                                                          &
          \multicolumn{1}{c|}{3.65\%}                                                        &
          workclass                                                                          &
          \multicolumn{1}{c|}{3.63\%}                                                        &
          capital\_loss                                                                      &
          \multicolumn{1}{c|}{3.67\%}                                                        &
          workclass                                                                          &
          \multicolumn{1}{c|}{3.55\%}                                                        &
          capital\_loss                                                                      &
          \multicolumn{1}{c|}{3.62\%}                                                        &
          capital\_loss                                                                      &
          \multicolumn{1}{c|}{3.60\%}                                                        &
          capital\_loss                                                                      &
          \multicolumn{1}{c|}{3.58\%}                                                        &
          capital\_loss                                                                      &
          3.54\%                                                                               \\
          workclass                                                                          &
          3.34\%                                                                             &
          \cellcolor[HTML]{83E28E}race                                                       &
          \multicolumn{1}{c|}{1.94\%}                                                        &
          \cellcolor[HTML]{83CCEB}age                                                        &
          \multicolumn{1}{c|}{1.80\%}                                                        &
          \cellcolor[HTML]{83E28E}race                                                       &
          \multicolumn{1}{c|}{2.25\%}                                                        &
          \cellcolor[HTML]{83CCEB}age                                                        &
          \multicolumn{1}{c|}{2.52\%}                                                        &
          workclass                                                                          &
          \multicolumn{1}{c|}{3.17\%}                                                        &
          workclass                                                                          &
          \multicolumn{1}{c|}{3.38\%}                                                        &
          workclass                                                                          &
          \multicolumn{1}{c|}{3.36\%}                                                        &
          workclass                                                                          &
          3.24\%                                                                               \\
          \cellcolor[HTML]{83E28E}race                                                       &
          1.21\%                                                                             &
          \cellcolor[HTML]{83CCEB}age                                                        &
          \multicolumn{1}{c|}{1.73\%}                                                        &
          \begin{tabular}[c]{@{}l@{}}native\_\\ country\end{tabular}                         &
          \multicolumn{1}{c|}{1.66\%}                                                        &
          \begin{tabular}[c]{@{}l@{}}native\_\\ country\end{tabular}                         &
          \multicolumn{1}{c|}{1.81\%}                                                        &
          \begin{tabular}[c]{@{}l@{}}native\_\\ country\end{tabular}                         &
          \multicolumn{1}{c|}{1.79\%}                                                        &
          \cellcolor[HTML]{E49EDD}\begin{tabular}[c]{@{}l@{}}marital\_\\ status\end{tabular} &
          \multicolumn{1}{c|}{2.97\%}                                                        &
          \cellcolor[HTML]{E49EDD}\begin{tabular}[c]{@{}l@{}}marital\_\\ status\end{tabular} &
          \multicolumn{1}{c|}{2.91\%}                                                        &
          \cellcolor[HTML]{E49EDD}\begin{tabular}[c]{@{}l@{}}marital\_\\ status\end{tabular} &
          \multicolumn{1}{c|}{2.96\%}                                                        &
          \cellcolor[HTML]{E49EDD}\begin{tabular}[c]{@{}l@{}}marital\_\\ status\end{tabular} &
          2.59\%                                                                               \\
          \begin{tabular}[c]{@{}l@{}}native\_\\ country\end{tabular}                         &
          1.16\%                                                                             &
          \begin{tabular}[c]{@{}l@{}}native\_\\ country\end{tabular}                         &
          \multicolumn{1}{c|}{1.67\%}                                                        &
          education                                                                          &
          \multicolumn{1}{c|}{1.40\%}                                                        &
          education                                                                          &
          \multicolumn{1}{c|}{1.27\%}                                                        &
          education                                                                          &
          \multicolumn{1}{c|}{1.50\%}                                                        &
          \cellcolor[HTML]{83E28E}race                                                       &
          \multicolumn{1}{c|}{2.32\%}                                                        &
          \cellcolor[HTML]{83E28E}race                                                       &
          \multicolumn{1}{c|}{2.37\%}                                                        &
          \cellcolor[HTML]{83E28E}race                                                       &
          \multicolumn{1}{c|}{2.34\%}                                                        &
          \cellcolor[HTML]{83E28E}race                                                       &
          2.18\%                                                                               \\
          education                                                                          &
          1.03\%                                                                             &
          education                                                                          &
          \multicolumn{1}{c|}{1.35\%}                                                        &
          \cellcolor[HTML]{F7C7AC}sex                                                        &
          \multicolumn{1}{c|}{1.31\%}                                                        &
          \cellcolor[HTML]{83CCEB}age                                                        &
          \multicolumn{1}{c|}{1.10\%}                                                        &
          \cellcolor[HTML]{83E28E}race                                                       &
          \multicolumn{1}{c|}{1.29\%}                                                        &
          education                                                                          &
          \multicolumn{1}{c|}{1.66\%}                                                        &
          \begin{tabular}[c]{@{}l@{}}native\_\\ country\end{tabular}                         &
          \multicolumn{1}{c|}{1.52\%}                                                        &
          \begin{tabular}[c]{@{}l@{}}native\_\\ country\end{tabular}                         &
          \multicolumn{1}{c|}{1.40\%}                                                        &
          education                                                                          &
          1.31\%                                                                               \\
          \cellcolor[HTML]{E49EDD}\begin{tabular}[c]{@{}l@{}}marital\_\\ status\end{tabular} &
          0.77\%                                                                             &
          \cellcolor[HTML]{E49EDD}\begin{tabular}[c]{@{}l@{}}marital\_\\ status\end{tabular} &
          \multicolumn{1}{c|}{0.77\%}                                                        &
          \cellcolor[HTML]{83E28E}race                                                       &
          \multicolumn{1}{c|}{0.98\%}                                                        &
          \cellcolor[HTML]{E49EDD}\begin{tabular}[c]{@{}l@{}}marital\_\\ status\end{tabular} &
          \multicolumn{1}{c|}{0.87\%}                                                        &
          \cellcolor[HTML]{E49EDD}\begin{tabular}[c]{@{}l@{}}marital\_\\ status\end{tabular} &
          \multicolumn{1}{c|}{0.80\%}                                                        &
          \begin{tabular}[c]{@{}l@{}}native\_\\ country\end{tabular}                         &
          \multicolumn{1}{c|}{1.38\%}                                                        &
          education                                                                          &
          \multicolumn{1}{c|}{1.33\%}                                                        &
          education                                                                          &
          \multicolumn{1}{c|}{1.01\%}                                                        &
          \begin{tabular}[c]{@{}l@{}}native\_\\ country\end{tabular}                         &
          1.25\%                                                                               \\
          \cellcolor[HTML]{F1A983}sex                                                        &
          0.39\%                                                                             &
          \cellcolor[HTML]{F7C7AC}sex                                                        &
          \multicolumn{1}{c|}{0.21\%}                                                        &
          \cellcolor[HTML]{E49EDD}\begin{tabular}[c]{@{}l@{}}marital\_\\ status\end{tabular} &
          \multicolumn{1}{c|}{0.71\%}                                                        &
          \cellcolor[HTML]{F7C7AC}sex                                                        &
          \multicolumn{1}{c|}{0.64\%}                                                        &
          \cellcolor[HTML]{F7C7AC}sex                                                        &
          \multicolumn{1}{c|}{0.10\%}                                                        &
          \cellcolor[HTML]{F7C7AC}sex                                                        &
          \multicolumn{1}{c|}{0.84\%}                                                        &
          \cellcolor[HTML]{F7C7AC}sex                                                        &
          \multicolumn{1}{c|}{0.70\%}                                                        &
          \cellcolor[HTML]{F7C7AC}sex                                                        &
          \multicolumn{1}{c|}{0.74\%}                                                        &
          \cellcolor[HTML]{F7C7AC}sex                                                        &
          0.84\%
        \end{tabular}%
      }
    \end{tabular}
  \end{table*}

  \subsection{Sepsis Patient}
  Table~\ref{tab: Sepsis original FI} presents the feature importance of Decision Trees using the original Sepsis patient dataset. And Table~\ref{tab: Sepsis PSM FI} presents the feature importance of Decision Trees using the PSM-adjusted Sepsis patient dataset.Tables~\ref{tab: Sepsis original FI}(a) and ~\ref{tab: Sepsis PSM FI}(a) show the changes in feature importance after applying the $k$-anonymity algorithm. Tables~\ref{tab: Sepsis original FI}(b) and~\ref{tab: Sepsis PSM FI}(b) display the changes in feature importance after applying the algorithm proposed by Zheng et al. Finally, Tables~\ref{tab: Sepsis original FI}(c) and~\ref{tab: Sepsis PSM FI}(c) illustrate the changes in feature importance after applying our model.

  \begin{table*}[!ht]
    \centering
    \caption{Variation of feature importances in the original Sepsis patient dataset with different anonymization models. Cells highlighted with color coding denote QIs}
    \label{tab: Sepsis original FI}
    \begin{tabular}{c}
      (a) $k$-anonymity \\
      \resizebox{\textwidth}{!}{%
        \begin{tabular}{m{2cm}c|m{2cm}cm{2cm}cm{2cm}cm{2cm}cm{2cm}cm{2cm}cm{2cm}c}
          \hline
          \multicolumn{2}{c|}{Original   Data}                                                      &
          \multicolumn{14}{c}{$k$-anonymity}                                                          \\ \hline
          Feature                                                                                   &
          Importance                                                                                &
          $k$=5                                                                                     &
          \multicolumn{1}{c|}{Importance}                                                           &
          $k$=10                                                                                    &
          \multicolumn{1}{c|}{Importance}                                                           &
          $k$=15                                                                                    &
          \multicolumn{1}{c|}{Importance}                                                           &
          $k$=20                                                                                    &
          \multicolumn{1}{c|}{Importance}                                                           &
          $k$=100                                                                                   &
          \multicolumn{1}{c|}{Importance}                                                           &
          $k$=300                                                                                   &
          \multicolumn{1}{c|}{Importance}                                                           &
          $k$=2000                                                                                  &
          Importance                                                                                  \\ \hline
          \begin{tabular}[c]{@{}l@{}}Antibiotic\\ \_AdminFlag\end{tabular}                          &
          30.49\%                                                                                   &
          \begin{tabular}[c]{@{}l@{}}Antibiotic\\ \_AdminFlag\end{tabular}                          &
          \multicolumn{1}{c|}{29.93\%}                                                              &
          \begin{tabular}[c]{@{}l@{}}Antibiotic\\ \_AdminFlag\end{tabular}                          &
          \multicolumn{1}{c|}{30.19\%}                                                              &
          \begin{tabular}[c]{@{}l@{}}Antibiotic\\ \_AdminFlag\end{tabular}                          &
          \multicolumn{1}{c|}{30.04\%}                                                              &
          \begin{tabular}[c]{@{}l@{}}Antibiotic\\ \_AdminFlag\end{tabular}                          &
          \multicolumn{1}{c|}{29.83\%}                                                              &
          \begin{tabular}[c]{@{}l@{}}Antibiotic\\ \_AdminFlag\end{tabular}                          &
          \multicolumn{1}{c|}{30.18\%}                                                              &
          \begin{tabular}[c]{@{}l@{}}Antibiotic\\ \_AdminFlag\end{tabular}                          &
          \multicolumn{1}{c|}{30.25\%}                                                              &
          \begin{tabular}[c]{@{}l@{}}Antibiotic\\ \_AdminFlag\end{tabular}                          &
          30.94\%                                                                                     \\
          \cellcolor[HTML]{83CCEB}AgeCategory                                                       &
          11.00\%                                                                                   &
          \cellcolor[HTML]{83CCEB}AgeCategory                                                       &
          \multicolumn{1}{c|}{10.90\%}                                                              &
          \cellcolor[HTML]{83CCEB}AgeCategory                                                       &
          \multicolumn{1}{c|}{10.25\%}                                                              &
          \cellcolor[HTML]{83CCEB}AgeCategory                                                       &
          \multicolumn{1}{c|}{10.64\%}                                                              &
          \cellcolor[HTML]{83CCEB}AgeCategory                                                       &
          \multicolumn{1}{c|}{10.36\%}                                                              &
          \cellcolor[HTML]{83CCEB}AgeCategory                                                       &
          \multicolumn{1}{c|}{9.15\%}                                                               &
          \cellcolor[HTML]{83CCEB}AgeCategory                                                       &
          \multicolumn{1}{c|}{8.23\%}                                                               &
          \cellcolor[HTML]{83CCEB}AgeCategory                                                       &
          5.06\%                                                                                      \\
          \cellcolor[HTML]{F7C7AC}LOSDays                                                           &
          8.64\%                                                                                    &
          \cellcolor[HTML]{F7C7AC}LOSDays                                                           &
          \multicolumn{1}{c|}{8.30\%}                                                               &
          \cellcolor[HTML]{F7C7AC}LOSDays                                                           &
          \multicolumn{1}{c|}{9.43\%}                                                               &
          \cellcolor[HTML]{F7C7AC}LOSDays                                                           &
          \multicolumn{1}{c|}{8.29\%}                                                               &
          \cellcolor[HTML]{F7C7AC}LOSDays                                                           &
          \multicolumn{1}{c|}{9.03\%}                                                               &
          \cellcolor[HTML]{F7C7AC}LOSDays                                                           &
          \multicolumn{1}{c|}{7.82\%}                                                               &
          \cellcolor[HTML]{F7C7AC}LOSDays                                                           &
          \multicolumn{1}{c|}{6.86\%}                                                               &
          \cellcolor[HTML]{F7C7AC}LOSDays                                                           &
          4.34\%                                                                                      \\
          LYTESFlag                                                                                 &
          3.72\%                                                                                    &
          LYTESFlag                                                                                 &
          \multicolumn{1}{c|}{3.79\%}                                                               &
          LYTESFlag                                                                                 &
          \multicolumn{1}{c|}{3.76\%}                                                               &
          LYTESFlag                                                                                 &
          \multicolumn{1}{c|}{3.69\%}                                                               &
          \begin{tabular}[c]{@{}l@{}}FirstLocation\\ TypeCodeAfter\\ Arrival\end{tabular}           &
          \multicolumn{1}{c|}{3.97\%}                                                               &
          LYTESFlag                                                                                 &
          \multicolumn{1}{c|}{3.70\%}                                                               &
          LYTESFlag                                                                                 &
          \multicolumn{1}{c|}{3.75\%}                                                               &
          LYTESFlag                                                                                 &
          3.76\%                                                                                      \\
          \begin{tabular}[c]{@{}l@{}}FirstLocation\\ TypeCodeAfter\\ Arrival\end{tabular}           &
          3.42\%                                                                                    &
          \begin{tabular}[c]{@{}l@{}}FirstLocation\\ TypeCodeAfter\\ Arrival\end{tabular}           &
          \multicolumn{1}{c|}{3.49\%}                                                               &
          \begin{tabular}[c]{@{}l@{}}FirstLocation\\ TypeCodeAfter\\ Arrival\end{tabular}           &
          \multicolumn{1}{c|}{3.29\%}                                                               &
          \begin{tabular}[c]{@{}l@{}}FirstLocation\\ TypeCodeAfter\\ Arrival\end{tabular}           &
          \multicolumn{1}{c|}{3.16\%}                                                               &
          \begin{tabular}[c]{@{}l@{}}\textgreater{}6HoursToFirst\\ AntibioticAdmin\end{tabular}     &
          \multicolumn{1}{c|}{2.12\%}                                                               &
          \begin{tabular}[c]{@{}l@{}}FirstLocation\\ TypeCodeAfter\\ Arrival\end{tabular}           &
          \multicolumn{1}{c|}{3.49\%}                                                               &
          \begin{tabular}[c]{@{}l@{}}FirstLocation\\ TypeCodeAfter\\ Arrival\end{tabular}           &
          \multicolumn{1}{c|}{3.64\%}                                                               &
          \begin{tabular}[c]{@{}l@{}}FirstLocation\\ TypeCodeAfter\\ Arrival\end{tabular}           &
          3.70\%                                                                                      \\
          \begin{tabular}[c]{@{}l@{}}\textgreater{}6HoursToFirst\\ AntibioticAdmin\end{tabular}     &
          2.28\%                                                                                    &
          \begin{tabular}[c]{@{}l@{}}\textgreater{}6HoursToFirst\\ AntibioticAdmin\end{tabular}     &
          \multicolumn{1}{c|}{2.25\%}                                                               &
          \begin{tabular}[c]{@{}l@{}}\textgreater{}6HoursToFirst\\ AntibioticAdmin\end{tabular}     &
          \multicolumn{1}{c|}{2.13\%}                                                               &
          \begin{tabular}[c]{@{}l@{}}\textgreater{}6HoursToFirst\\ AntibioticAdmin\end{tabular}     &
          \multicolumn{1}{c|}{2.19\%}                                                               &
          \cellcolor[HTML]{83E28E}\begin{tabular}[c]{@{}l@{}}Race\\ Description\end{tabular}        &
          \multicolumn{1}{c|}{2.04\%}                                                               &
          \begin{tabular}[c]{@{}l@{}}\textgreater{}6HoursToFirst\\ AntibioticAdmin\end{tabular}     &
          \multicolumn{1}{c|}{2.13\%}                                                               &
          \begin{tabular}[c]{@{}l@{}}\textgreater{}6HoursToFirst\\ AntibioticAdmin\end{tabular}     &
          \multicolumn{1}{c|}{2.19\%}                                                               &
          \begin{tabular}[c]{@{}l@{}}\textgreater{}6HoursToFirst\\ AntibioticAdmin\end{tabular}     &
          2.13\%                                                                                      \\
          \cellcolor[HTML]{83E28E}\begin{tabular}[c]{@{}l@{}}Race\\ Description\end{tabular}        &
          1.81\%                                                                                    &
          \cellcolor[HTML]{83E28E}\begin{tabular}[c]{@{}l@{}}Race\\ Description\end{tabular}        &
          \multicolumn{1}{c|}{1.73\%}                                                               &
          \cellcolor[HTML]{83E28E}\begin{tabular}[c]{@{}l@{}}Race\\ Description\end{tabular}        &
          \multicolumn{1}{c|}{1.92\%}                                                               &
          \cellcolor[HTML]{FFFD78}NumberofVisits                                                    &
          \multicolumn{1}{c|}{2.08\%}                                                               &
          LYTESFlag                                                                                 &
          \multicolumn{1}{c|}{2.02\%}                                                               &
          \cellcolor[HTML]{FFFD78}NumberofVisits                                                    &
          \multicolumn{1}{c|}{1.99\%}                                                               &
          \cellcolor[HTML]{FFFD78}NumberofVisits                                                    &
          \multicolumn{1}{c|}{1.85\%}                                                               &
          FluSeasonFlag                                                                             &
          1.91\%                                                                                      \\
          \cellcolor[HTML]{E49EDD}\begin{tabular}[c]{@{}l@{}}Gender\\ Description\end{tabular}      &
          1.61\%                                                                                    &
          \cellcolor[HTML]{FFFD78}NumberofVisits                                                    &
          \multicolumn{1}{c|}{1.70\%}                                                               &
          \cellcolor[HTML]{FFFD78}NumberofVisits                                                    &
          \multicolumn{1}{c|}{1.71\%}                                                               &
          \cellcolor[HTML]{83E28E}\begin{tabular}[c]{@{}l@{}}Race\\ Description\end{tabular}        &
          \multicolumn{1}{c|}{1.69\%}                                                               &
          \cellcolor[HTML]{FFFD78}NumberofVisits                                                    &
          \multicolumn{1}{c|}{1.86\%}                                                               &
          \cellcolor[HTML]{83E28E}\begin{tabular}[c]{@{}l@{}}Race\\ Description\end{tabular}        &
          \multicolumn{1}{c|}{1.81\%}                                                               &
          FluSeasonFlag                                                                             &
          \multicolumn{1}{c|}{1.79\%}                                                               &
          HTNFlag                                                                                   &
          1.87\%                                                                                      \\
          \cellcolor[HTML]{FFFD78}NumberofVisits                                                    &
          1.58\%                                                                                    &
          \cellcolor[HTML]{E49EDD}\begin{tabular}[c]{@{}l@{}}Gender\\ Description\end{tabular}      &
          \multicolumn{1}{c|}{1.51\%}                                                               &
          \cellcolor[HTML]{E49EDD}\begin{tabular}[c]{@{}l@{}}Gender\\ Description\end{tabular}      &
          \multicolumn{1}{c|}{1.55\%}                                                               &
          FluSeasonFlag                                                                             &
          \multicolumn{1}{c|}{1.67\%}                                                               &
          FluSeasonFlag                                                                             &
          \multicolumn{1}{c|}{1.54\%}                                                               &
          FluSeasonFlag                                                                             &
          \multicolumn{1}{c|}{1.56\%}                                                               &
          \cellcolor[HTML]{83E28E}RaceDescription                                                   &
          \multicolumn{1}{c|}{1.77\%}                                                               &
          ANEMDEFFlag                                                                               &
          1.51\%                                                                                      \\
          FluSeasonFlag                                                                             &
          1.53\%                                                                                    &
          HTNFlag                                                                                   &
          \multicolumn{1}{c|}{1.39\%}                                                               &
          FluSeasonFlag                                                                             &
          \multicolumn{1}{c|}{1.26\%}                                                               &
          \cellcolor[HTML]{E49EDD}\begin{tabular}[c]{@{}l@{}}Gender\\ Description\end{tabular}      &
          \multicolumn{1}{c|}{1.53\%}                                                               &
          \cellcolor[HTML]{E49EDD}\begin{tabular}[c]{@{}l@{}}Gender\\ Description\end{tabular}      &
          \multicolumn{1}{c|}{1.43\%}                                                               &
          \cellcolor[HTML]{E49EDD}\begin{tabular}[c]{@{}l@{}}Gender\\ Description\end{tabular}      &
          \multicolumn{1}{c|}{1.55\%}                                                               &
          HTNFlag                                                                                   &
          \multicolumn{1}{c|}{1.55\%}                                                               &
          \cellcolor[HTML]{83E28E}\begin{tabular}[c]{@{}l@{}}Race\\ Description\end{tabular}        &
          1.44\%                                                                                      \\
          HX\_BLDLOSS                                                                               &
          1.39\%                                                                                    &
          HX\_BLDLOSS                                                                               &
          \multicolumn{1}{c|}{1.37\%}                                                               &
          HX\_BLDLOSS                                                                               &
          \multicolumn{1}{c|}{1.22\%}                                                               &
          HX\_BLDLOSS                                                                               &
          \multicolumn{1}{c|}{1.40\%}                                                               &
          HTNFlag                                                                                   &
          \multicolumn{1}{c|}{1.28\%}                                                               &
          HX\_BLDLOSS                                                                               &
          \multicolumn{1}{c|}{1.35\%}                                                               &
          ANEMDEFFlag                                                                               &
          \multicolumn{1}{c|}{1.39\%}                                                               &
          HX\_ULCER                                                                                 &
          1.41\%                                                                                      \\
          \begin{tabular}[c]{@{}l@{}}CHRNLUNG\\ Flag\end{tabular}                                   &
          1.03\%                                                                                    &
          FluSeasonFlag                                                                             &
          \multicolumn{1}{c|}{1.23\%}                                                               &
          ANEMDEFFlag                                                                               &
          \multicolumn{1}{c|}{1.07\%}                                                               &
          HTNFlag                                                                                   &
          \multicolumn{1}{c|}{1.03\%}                                                               &
          ANEMDEFFlag                                                                               &
          \multicolumn{1}{c|}{1.15\%}                                                               &
          ANEMDEFFlag                                                                               &
          \multicolumn{1}{c|}{1.33\%}                                                               &
          HX\_BLDLOSS                                                                               &
          \multicolumn{1}{c|}{1.32\%}                                                               &
          CHRNLUNGFlag                                                                              &
          1.41\%                                                                                      \\
          HTNFlag                                                                                   &
          1.00\%                                                                                    &
          ANEMDEFFlag                                                                               &
          \multicolumn{1}{c|}{1.21\%}                                                               &
          \cellcolor[HTML]{53E8BF}\begin{tabular}[c]{@{}l@{}}EthnicGroup\\ Description\end{tabular} &
          \multicolumn{1}{c|}{1.01\%}                                                               &
          CHRNLUNGFlag                                                                              &
          \multicolumn{1}{c|}{1.01\%}                                                               &
          CHRNLUNGFlag                                                                              &
          \multicolumn{1}{c|}{1.07\%}                                                               &
          HTNFlag                                                                                   &
          \multicolumn{1}{c|}{1.16\%}                                                               &
          CHRNLUNGFlag                                                                              &
          \multicolumn{1}{c|}{1.22\%}                                                               &
          NEUROFlag                                                                                 &
          1.29\%                                                                                      \\
          \begin{tabular}[c]{@{}l@{}}ANEMDEF\\ Flag\end{tabular}                                    &
          0.94\%                                                                                    &
          DMFlag                                                                                    &
          \multicolumn{1}{c|}{0.98\%}                                                               &
          DMFlag                                                                                    &
          \multicolumn{1}{c|}{0.95\%}                                                               &
          ANEMDEFFlag                                                                               &
          \multicolumn{1}{c|}{0.98\%}                                                               &
          DMFlag                                                                                    &
          \multicolumn{1}{c|}{0.98\%}                                                               &
          CHRNLUNGFlag                                                                              &
          \multicolumn{1}{c|}{1.02\%}                                                               &
          \cellcolor[HTML]{E49EDD}\begin{tabular}[c]{@{}l@{}}Gender\\ Description\end{tabular}      &
          \multicolumn{1}{c|}{1.15\%}                                                               &
          \cellcolor[HTML]{FFFD78}NumberofVisits                                                    &
          1.28\%                                                                                      \\
          NEUROFlag                                                                                 &
          0.91\%                                                                                    &
          CHRNLUNGFlag                                                                              &
          \multicolumn{1}{c|}{0.92\%}                                                               &
          HTNFlag                                                                                   &
          \multicolumn{1}{c|}{0.91\%}                                                               &
          \cellcolor[HTML]{53E8BF}\begin{tabular}[c]{@{}l@{}}EthnicGroup\\ Description\end{tabular} &
          \multicolumn{1}{c|}{0.95\%}                                                               &
          OBESEFlag                                                                                 &
          \multicolumn{1}{c|}{0.91\%}                                                               &
          NEUROFlag                                                                                 &
          \multicolumn{1}{c|}{0.92\%}                                                               &
          COAGFlag                                                                                  &
          \multicolumn{1}{c|}{1.06\%}                                                               &
          OBESEFlag                                                                                 &
          1.20\%                                                                                      \\
          CHFFlag                                                                                   &
          0.84\%                                                                                    &
          \cellcolor[HTML]{53E8BF}\begin{tabular}[c]{@{}l@{}}EthnicGroup\\ Description\end{tabular} &
          \multicolumn{1}{c|}{0.84\%}                                                               &
          CHFFlag                                                                                   &
          \multicolumn{1}{c|}{0.89\%}                                                               &
          DEPRESSFlag                                                                               &
          \multicolumn{1}{c|}{0.87\%}                                                               &
          CADFlag                                                                                   &
          \multicolumn{1}{c|}{0.90\%}                                                               &
          CADFlag                                                                                   &
          \multicolumn{1}{c|}{0.90\%}                                                               &
          DMFlag                                                                                    &
          \multicolumn{1}{c|}{0.99\%}                                                               &
          CADFlag                                                                                   &
          1.10\%                                                                                      \\
          DMFlag                                                                                    &
          0.84\%                                                                                    &
          HX\_LYTES                                                                                 &
          \multicolumn{1}{c|}{0.80\%}                                                               &
          NEUROFlag                                                                                 &
          \multicolumn{1}{c|}{0.87\%}                                                               &
          DMFlag                                                                                    &
          \multicolumn{1}{c|}{0.85\%}                                                               &
          HX\_Sepsis                                                                                &
          \multicolumn{1}{c|}{0.89\%}                                                               &
          HX\_LYTES                                                                                 &
          \multicolumn{1}{c|}{0.88\%}                                                               &
          CADFlag                                                                                   &
          \multicolumn{1}{c|}{0.95\%}                                                               &
          DEPRESSFlag                                                                               &
          1.10\%                                                                                      \\
          OBESEFlag                                                                                 &
          0.79\%                                                                                    &
          CHFFlag                                                                                   &
          \multicolumn{1}{c|}{0.80\%}                                                               &
          OBESEFlag                                                                                 &
          \multicolumn{1}{c|}{0.79\%}                                                               &
          NEUROFlag                                                                                 &
          \multicolumn{1}{c|}{0.84\%}                                                               &
          DEPRESSFlag                                                                               &
          \multicolumn{1}{c|}{0.83\%}                                                               &
          DEPRESSFlag                                                                               &
          \multicolumn{1}{c|}{0.83\%}                                                               &
          DEPRESSFlag                                                                               &
          \multicolumn{1}{c|}{0.92\%}                                                               &
          DMFlag                                                                                    &
          1.09\%                                                                                      \\
          HX\_LYTES                                                                                 &
          0.77\%                                                                                    &
          NEUROFlag                                                                                 &
          \multicolumn{1}{c|}{0.79\%}                                                               &
          COAGFlag                                                                                  &
          \multicolumn{1}{c|}{0.79\%}                                                               &
          COAGFlag                                                                                  &
          \multicolumn{1}{c|}{0.83\%}                                                               &
          NEUROFlag                                                                                 &
          \multicolumn{1}{c|}{0.78\%}                                                               &
          CHFFlag                                                                                   &
          \multicolumn{1}{c|}{0.79\%}                                                               &
          OBESEFlag                                                                                 &
          \multicolumn{1}{c|}{0.88\%}                                                               &
          CHFFlag                                                                                   &
          0.99\%                                                                                      \\
          \begin{tabular}[c]{@{}l@{}}HYPOTHY\\ Flag\end{tabular}                                    &
          0.73\%                                                                                    &
          OBESEFlag                                                                                 &
          \multicolumn{1}{c|}{0.78\%}                                                               &
          DEPRESSFlag                                                                               &
          \multicolumn{1}{c|}{0.75\%}                                                               &
          CHFFlag                                                                                   &
          \multicolumn{1}{c|}{0.81\%}                                                               &
          HYPOTHYFlag                                                                               &
          \multicolumn{1}{c|}{0.76\%}                                                               &
          HX\_CHRNLUNG                                                                              &
          \multicolumn{1}{c|}{0.78\%}                                                               &
          CHFFlag                                                                                   &
          \multicolumn{1}{c|}{0.85\%}                                                               &
          HYPOTHYFlag                                                                               &
          0.98\%                                                                                      \\
          \cellcolor[HTML]{53E8BF}\begin{tabular}[c]{@{}l@{}}EthnicGroup\\ Description\end{tabular} &
          0.73\%                                                                                    &
          COAGFlag                                                                                  &
          \multicolumn{1}{c|}{0.77\%}                                                               &
          HYPOTHYFlag                                                                               &
          \multicolumn{1}{c|}{0.74\%}                                                               &
          CADFlag                                                                                   &
          \multicolumn{1}{c|}{0.76\%}                                                               &
          CHFFlag                                                                                   &
          \multicolumn{1}{c|}{0.76\%}                                                               &
          OBESEFlag                                                                                 &
          \multicolumn{1}{c|}{0.77\%}                                                               &
          HX\_HTN                                                                                   &
          \multicolumn{1}{c|}{0.85\%}                                                               &
          COAGFlag                                                                                  &
          0.96\%                                                                                      \\
          COAGFlag                                                                                  &
          0.72\%                                                                                    &
          CADFlag                                                                                   &
          \multicolumn{1}{c|}{0.70\%}                                                               &
          CHRNLUNGFlag                                                                              &
          \multicolumn{1}{c|}{0.73\%}                                                               &
          OBESEFlag                                                                                 &
          \multicolumn{1}{c|}{0.75\%}                                                               &
          COAGFlag                                                                                  &
          \multicolumn{1}{c|}{0.75\%}                                                               &
          COAGFlag                                                                                  &
          \multicolumn{1}{c|}{0.75\%}                                                               &
          HX\_LYTES                                                                                 &
          \multicolumn{1}{c|}{0.78\%}                                                               &
          \cellcolor[HTML]{E49EDD}\begin{tabular}[c]{@{}l@{}}Gender\\ Description\end{tabular}      &
          0.96\%                                                                                      \\
          CADFlag                                                                                   &
          0.72\%                                                                                    &
          DEPRESSFlag                                                                               &
          \multicolumn{1}{c|}{0.69\%}                                                               &
          HX\_HTN                                                                                   &
          \multicolumn{1}{c|}{0.72\%}                                                               &
          HX\_Uti                                                                                   &
          \multicolumn{1}{c|}{0.70\%}                                                               &
          PSYCHFlag                                                                                 &
          \multicolumn{1}{c|}{0.73\%}                                                               &
          \cellcolor[HTML]{53E8BF}\begin{tabular}[c]{@{}l@{}}EthnicGroup\\ Description\end{tabular} &
          \multicolumn{1}{c|}{0.71\%}                                                               &
          HYPOTHYFlag                                                                               &
          \multicolumn{1}{c|}{0.77\%}                                                               &
          HX\_CHRNLUNG                                                                              &
          0.89\%                                                                                      \\
        \end{tabular}%
      }
    \end{tabular}
  \end{table*}

  \begin{table*}[!ht]
    \centering
    \begin{tabular}{c}
      (b) Algorithm proposed \citet{zheng2019effective} \\
      \resizebox{\textwidth}{!}{%
        \begin{tabular}{m{2cm}c|m{2cm}cm{2cm}cm{2cm}cm{2cm}cm{2cm}cm{2cm}cm{2cm}c}
          \hline
          \multicolumn{2}{c|}{Original   Data}                                                      &
          \multicolumn{14}{c}{Zheng et al}                                                            \\ \hline
          Feature                                                                                   &
          Importance                                                                                &
          $k$=5,$l$=2                                                                               &
          \multicolumn{1}{c|}{Importance}                                                           &
          $k$=10,$l$=2                                                                              &
          \multicolumn{1}{c|}{Importance}                                                           &
          $k$=15,$l$=2                                                                              &
          \multicolumn{1}{c|}{Importance}                                                           &
          $k$=20,$l$=2                                                                              &
          \multicolumn{1}{c|}{Importance}                                                           &
          $k$=100,$l$=2                                                                             &
          \multicolumn{1}{c|}{Importance}                                                           &
          $k$=300,$l$=2                                                                             &
          \multicolumn{1}{c|}{Importance}                                                           &
          $k$=2000,$l$=2                                                                            &
          Importance                                                                                  \\ \hline
          \begin{tabular}[c]{@{}l@{}}Antibiotic\\ \_AdminFlag\end{tabular}                          &
          30.49\%                                                                                   &
          \begin{tabular}[c]{@{}l@{}}Antibiotic\\ \_AdminFlag\end{tabular}                          &
          \multicolumn{1}{c|}{30.17\%}                                                              &
          \begin{tabular}[c]{@{}l@{}}Antibiotic\\ \_AdminFlag\end{tabular}                          &
          \multicolumn{1}{c|}{30.07\%}                                                              &
          \begin{tabular}[c]{@{}l@{}}Antibiotic\\ \_AdminFlag\end{tabular}                          &
          \multicolumn{1}{c|}{30.14\%}                                                              &
          \begin{tabular}[c]{@{}l@{}}Antibiotic\\ \_AdminFlag\end{tabular}                          &
          \multicolumn{1}{c|}{30.21\%}                                                              &
          \begin{tabular}[c]{@{}l@{}}Antibiotic\\ \_AdminFlag\end{tabular}                          &
          \multicolumn{1}{c|}{30.25\%}                                                              &
          \begin{tabular}[c]{@{}l@{}}Antibiotic\\ \_AdminFlag\end{tabular}                          &
          \multicolumn{1}{c|}{30.21\%}                                                              &
          \begin{tabular}[c]{@{}l@{}}Antibiotic\\ \_AdminFlag\end{tabular}                          &
          30.61\%                                                                                     \\
          \cellcolor[HTML]{83CCEB}AgeCategory                                                       &
          11.00\%                                                                                   &
          \cellcolor[HTML]{83CCEB}AgeCategory                                                       &
          \multicolumn{1}{c|}{10.87\%}                                                              &
          \cellcolor[HTML]{83CCEB}AgeCategory                                                       &
          \multicolumn{1}{c|}{10.28\%}                                                              &
          \cellcolor[HTML]{83CCEB}AgeCategory                                                       &
          \multicolumn{1}{c|}{10.22\%}                                                              &
          \cellcolor[HTML]{83CCEB}AgeCategory                                                       &
          \multicolumn{1}{c|}{9.73\%}                                                               &
          \cellcolor[HTML]{83CCEB}AgeCategory                                                       &
          \multicolumn{1}{c|}{9.29\%}                                                               &
          \cellcolor[HTML]{83CCEB}AgeCategory                                                       &
          \multicolumn{1}{c|}{7.92\%}                                                               &
          \cellcolor[HTML]{83CCEB}AgeCategory                                                       &
          5.34\%                                                                                      \\
          \cellcolor[HTML]{F7C7AC}LOSDays                                                           &
          8.64\%                                                                                    &
          \cellcolor[HTML]{F7C7AC}LOSDays                                                           &
          \multicolumn{1}{c|}{7.89\%}                                                               &
          \cellcolor[HTML]{F7C7AC}LOSDays                                                           &
          \multicolumn{1}{c|}{8.25\%}                                                               &
          \cellcolor[HTML]{F7C7AC}LOSDays                                                           &
          \multicolumn{1}{c|}{8.91\%}                                                               &
          \cellcolor[HTML]{F7C7AC}LOSDays                                                           &
          \multicolumn{1}{c|}{7.84\%}                                                               &
          \cellcolor[HTML]{F7C7AC}LOSDays                                                           &
          \multicolumn{1}{c|}{7.43\%}                                                               &
          \cellcolor[HTML]{F7C7AC}LOSDays                                                           &
          \multicolumn{1}{c|}{6.50\%}                                                               &
          \cellcolor[HTML]{F7C7AC}LOSDays                                                           &
          4.24\%                                                                                      \\
          LYTESFlag                                                                                 &
          3.72\%                                                                                    &
          LYTESFlag                                                                                 &
          \multicolumn{1}{c|}{3.68\%}                                                               &
          LYTESFlag                                                                                 &
          \multicolumn{1}{c|}{3.66\%}                                                               &
          LYTESFlag                                                                                 &
          \multicolumn{1}{c|}{3.74\%}                                                               &
          LYTESFlag                                                                                 &
          \multicolumn{1}{c|}{3.89\%}                                                               &
          LYTESFlag                                                                                 &
          \multicolumn{1}{c|}{3.88\%}                                                               &
          LYTESFlag                                                                                 &
          \multicolumn{1}{c|}{3.77\%}                                                               &
          LYTESFlag                                                                                 &
          3.92\%                                                                                      \\
          \begin{tabular}[c]{@{}l@{}}FirstLocation\\ TypeCodeAfter\\ Arrival\end{tabular}           &
          3.42\%                                                                                    &
          \begin{tabular}[c]{@{}l@{}}FirstLocation\\ TypeCodeAfter\\ Arrival\end{tabular}           &
          \multicolumn{1}{c|}{3.32\%}                                                               &
          \begin{tabular}[c]{@{}l@{}}FirstLocation\\ TypeCodeAfter\\ Arrival\end{tabular}           &
          \multicolumn{1}{c|}{3.37\%}                                                               &
          \begin{tabular}[c]{@{}l@{}}FirstLocation\\ TypeCodeAfter\\ Arrival\end{tabular}           &
          \multicolumn{1}{c|}{3.41\%}                                                               &
          \begin{tabular}[c]{@{}l@{}}FirstLocation\\ TypeCodeAfter\\ Arrival\end{tabular}           &
          \multicolumn{1}{c|}{3.38\%}                                                               &
          \begin{tabular}[c]{@{}l@{}}FirstLocation\\ TypeCodeAfter\\ Arrival\end{tabular}           &
          \multicolumn{1}{c|}{3.51\%}                                                               &
          \begin{tabular}[c]{@{}l@{}}FirstLocation\\ TypeCodeAfter\\ Arrival\end{tabular}           &
          \multicolumn{1}{c|}{3.74\%}                                                               &
          \begin{tabular}[c]{@{}l@{}}FirstLocation\\ TypeCodeAfter\\ Arrival\end{tabular}           &
          3.39\%                                                                                      \\
          \begin{tabular}[c]{@{}l@{}}\textgreater{}6HoursToFirst\\ AntibioticAdmin\end{tabular}     &
          2.28\%                                                                                    &
          \begin{tabular}[c]{@{}l@{}}\textgreater{}6HoursToFirst\\ AntibioticAdmin\end{tabular}     &
          \multicolumn{1}{c|}{2.16\%}                                                               &
          \begin{tabular}[c]{@{}l@{}}\textgreater{}6HoursToFirst\\ AntibioticAdmin\end{tabular}     &
          \multicolumn{1}{c|}{2.40\%}                                                               &
          \begin{tabular}[c]{@{}l@{}}\textgreater{}6HoursToFirst\\ AntibioticAdmin\end{tabular}     &
          \multicolumn{1}{c|}{2.23\%}                                                               &
          \begin{tabular}[c]{@{}l@{}}\textgreater{}6HoursToFirst\\ AntibioticAdmin\end{tabular}     &
          \multicolumn{1}{c|}{2.21\%}                                                               &
          \cellcolor[HTML]{FFFD78}NumberofVisits                                                    &
          \multicolumn{1}{c|}{2.46\%}                                                               &
          \begin{tabular}[c]{@{}l@{}}\textgreater{}6HoursToFirst\\ AntibioticAdmin\end{tabular}     &
          \multicolumn{1}{c|}{2.26\%}                                                               &
          \cellcolor[HTML]{FFFD78}NumberofVisits                                                    &
          2.29\%                                                                                      \\
          \cellcolor[HTML]{83E28E}\begin{tabular}[c]{@{}l@{}}Race\\ Description\end{tabular}        &
          1.81\%                                                                                    &
          \cellcolor[HTML]{FFFD78}NumberofVisits                                                    &
          \multicolumn{1}{c|}{2.01\%}                                                               &
          \cellcolor[HTML]{FFFD78}NumberofVisits                                                    &
          \multicolumn{1}{c|}{2.01\%}                                                               &
          \cellcolor[HTML]{FFFD78}NumberofVisits                                                    &
          \multicolumn{1}{c|}{2.01\%}                                                               &
          \cellcolor[HTML]{FFFD78}NumberofVisits                                                    &
          \multicolumn{1}{c|}{2.20\%}                                                               &
          \begin{tabular}[c]{@{}l@{}}\textgreater{}6HoursToFirst\\ AntibioticAdmin\end{tabular}     &
          \multicolumn{1}{c|}{2.18\%}                                                               &
          \cellcolor[HTML]{FFFD78}NumberofVisits                                                    &
          \multicolumn{1}{c|}{2.14\%}                                                               &
          FluSeasonFlag                                                                             &
          2.22\%                                                                                      \\
          \cellcolor[HTML]{E49EDD}\begin{tabular}[c]{@{}l@{}}Gender\\ Description\end{tabular}      &
          1.61\%                                                                                    &
          FluSeasonFlag                                                                             &
          \multicolumn{1}{c|}{1.78\%}                                                               &
          \cellcolor[HTML]{83E28E}\begin{tabular}[c]{@{}l@{}}Race\\ Description\end{tabular}        &
          \multicolumn{1}{c|}{1.82\%}                                                               &
          \cellcolor[HTML]{83E28E}\begin{tabular}[c]{@{}l@{}}Race\\ Description\end{tabular}        &
          \multicolumn{1}{c|}{1.78\%}                                                               &
          FluSeasonFlag                                                                             &
          \multicolumn{1}{c|}{1.68\%}                                                               &
          FluSeasonFlag                                                                             &
          \multicolumn{1}{c|}{1.54\%}                                                               &
          FluSeasonFlag                                                                             &
          \multicolumn{1}{c|}{2.00\%}                                                               &
          \begin{tabular}[c]{@{}l@{}}\textgreater{}6HoursToFirst\\ AntibioticAdmin\end{tabular}     &
          2.02\%                                                                                      \\
          \cellcolor[HTML]{FFFD78}NumberofVisits                                                    &
          1.58\%                                                                                    &
          \cellcolor[HTML]{83E28E}\begin{tabular}[c]{@{}l@{}}Race\\ Description\end{tabular}        &
          \multicolumn{1}{c|}{1.62\%}                                                               &
          FluSeasonFlag                                                                             &
          \multicolumn{1}{c|}{1.62\%}                                                               &
          \cellcolor[HTML]{E49EDD}\begin{tabular}[c]{@{}l@{}}Gender\\ Description\end{tabular}      &
          \multicolumn{1}{c|}{1.35\%}                                                               &
          HX\_BLDLOSS                                                                               &
          \multicolumn{1}{c|}{1.44\%}                                                               &
          ANEMDEFFlag                                                                               &
          \multicolumn{1}{c|}{1.29\%}                                                               &
          \cellcolor[HTML]{83E28E}\begin{tabular}[c]{@{}l@{}}Race\\ Description\end{tabular}        &
          \multicolumn{1}{c|}{1.49\%}                                                               &
          HTNFlag                                                                                   &
          1.61\%                                                                                      \\
          FluSeasonFlag                                                                             &
          1.53\%                                                                                    &
          \cellcolor[HTML]{E49EDD}\begin{tabular}[c]{@{}l@{}}Gender\\ Description\end{tabular}      &
          \multicolumn{1}{c|}{1.41\%}                                                               &
          \cellcolor[HTML]{E49EDD}\begin{tabular}[c]{@{}l@{}}Gender\\ Description\end{tabular}      &
          \multicolumn{1}{c|}{1.35\%}                                                               &
          HTNFlag                                                                                   &
          \multicolumn{1}{c|}{1.17\%}                                                               &
          \cellcolor[HTML]{E49EDD}\begin{tabular}[c]{@{}l@{}}Gender\\ Description\end{tabular}      &
          \multicolumn{1}{c|}{1.40\%}                                                               &
          \cellcolor[HTML]{E49EDD}\begin{tabular}[c]{@{}l@{}}Gender\\ Description\end{tabular}      &
          \multicolumn{1}{c|}{1.28\%}                                                               &
          ANEMDEFFlag                                                                               &
          \multicolumn{1}{c|}{1.40\%}                                                               &
          ANEMDEFFlag                                                                               &
          1.54\%                                                                                      \\
          HX\_BLDLOSS                                                                               &
          1.39\%                                                                                    &
          HX\_BLDLOSS                                                                               &
          \multicolumn{1}{c|}{1.36\%}                                                               &
          HTNFlag                                                                                   &
          \multicolumn{1}{c|}{1.19\%}                                                               &
          FluSeasonFlag                                                                             &
          \multicolumn{1}{c|}{1.09\%}                                                               &
          \cellcolor[HTML]{83E28E}\begin{tabular}[c]{@{}l@{}}Race\\ Description\end{tabular}        &
          \multicolumn{1}{c|}{1.35\%}                                                               &
          HX\_ULCER                                                                                 &
          \multicolumn{1}{c|}{1.27\%}                                                               &
          HTNFlag                                                                                   &
          \multicolumn{1}{c|}{1.38\%}                                                               &
          HX\_BLDLOSS                                                                               &
          1.29\%                                                                                      \\
          CHRNLUNGFlag                                                                              &
          1.03\%                                                                                    &
          HTNFlag                                                                                   &
          \multicolumn{1}{c|}{1.19\%}                                                               &
          CHRNLUNGFlag                                                                              &
          \multicolumn{1}{c|}{1.15\%}                                                               &
          ANEMDEFFlag                                                                               &
          \multicolumn{1}{c|}{1.07\%}                                                               &
          HTNFlag                                                                                   &
          \multicolumn{1}{c|}{1.31\%}                                                               &
          CHRNLUNGFlag                                                                              &
          \multicolumn{1}{c|}{1.24\%}                                                               &
          HX\_BLDLOSS                                                                               &
          \multicolumn{1}{c|}{1.33\%}                                                               &
          DEPRESSFlag                                                                               &
          1.26\%                                                                                      \\
          HTNFlag                                                                                   &
          1.00\%                                                                                    &
          ANEMDEFFlag                                                                               &
          \multicolumn{1}{c|}{1.10\%}                                                               &
          ANEMDEFFlag                                                                               &
          \multicolumn{1}{c|}{1.14\%}                                                               &
          CHRNLUNGFlag                                                                              &
          \multicolumn{1}{c|}{0.97\%}                                                               &
          ANEMDEFFlag                                                                               &
          \multicolumn{1}{c|}{1.20\%}                                                               &
          HTNFlag                                                                                   &
          \multicolumn{1}{c|}{1.19\%}                                                               &
          \cellcolor[HTML]{E49EDD}\begin{tabular}[c]{@{}l@{}}Gender\\ Description\end{tabular}      &
          \multicolumn{1}{c|}{1.13\%}                                                               &
          CHRNLUNGFlag                                                                              &
          1.23\%                                                                                      \\
          ANEMDEFFlag                                                                               &
          0.94\%                                                                                    &
          DEPRESSFlag                                                                               &
          \multicolumn{1}{c|}{0.96\%}                                                               &
          DMFlag                                                                                    &
          \multicolumn{1}{c|}{0.91\%}                                                               &
          DMFlag                                                                                    &
          \multicolumn{1}{c|}{0.93\%}                                                               &
          NEUROFlag                                                                                 &
          \multicolumn{1}{c|}{0.95\%}                                                               &
          \cellcolor[HTML]{83E28E}\begin{tabular}[c]{@{}l@{}}Race\\ Description\end{tabular}        &
          \multicolumn{1}{c|}{1.15\%}                                                               &
          CADFlag                                                                                   &
          \multicolumn{1}{c|}{1.05\%}                                                               &
          NEUROFlag                                                                                 &
          1.18\%                                                                                      \\
          NEUROFlag                                                                                 &
          0.91\%                                                                                    &
          DMFlag                                                                                    &
          \multicolumn{1}{c|}{0.95\%}                                                               &
          COAGFlag                                                                                  &
          \multicolumn{1}{c|}{0.87\%}                                                               &
          HX\_HTN                                                                                   &
          \multicolumn{1}{c|}{0.90\%}                                                               &
          COAGFlag                                                                                  &
          \multicolumn{1}{c|}{0.94\%}                                                               &
          DMFlag                                                                                    &
          \multicolumn{1}{c|}{0.94\%}                                                               &
          DMFlag                                                                                    &
          \multicolumn{1}{c|}{1.04\%}                                                               &
          \cellcolor[HTML]{E49EDD}\begin{tabular}[c]{@{}l@{}}Gender\\ Description\end{tabular}      &
          1.17\%                                                                                      \\
          CHFFlag                                                                                   &
          0.84\%                                                                                    &
          COAGFlag                                                                                  &
          \multicolumn{1}{c|}{0.91\%}                                                               &
          NEUROFlag                                                                                 &
          \multicolumn{1}{c|}{0.86\%}                                                               &
          NEUROFlag                                                                                 &
          \multicolumn{1}{c|}{0.89\%}                                                               &
          CADFlag                                                                                   &
          \multicolumn{1}{c|}{0.94\%}                                                               &
          NEUROFlag                                                                                 &
          \multicolumn{1}{c|}{0.94\%}                                                               &
          CHRNLUNGFlag                                                                              &
          \multicolumn{1}{c|}{0.99\%}                                                               &
          CADFlag                                                                                   &
          1.17\%                                                                                      \\
          DMFlag                                                                                    &
          0.84\%                                                                                    &
          CADFlag                                                                                   &
          \multicolumn{1}{c|}{0.82\%}                                                               &
          CADFlag                                                                                   &
          \multicolumn{1}{c|}{0.84\%}                                                               &
          OBESEFlag                                                                                 &
          \multicolumn{1}{c|}{0.87\%}                                                               &
          CHRNLUNGFlag                                                                              &
          \multicolumn{1}{c|}{0.92\%}                                                               &
          DEPRESSFlag                                                                               &
          \multicolumn{1}{c|}{0.92\%}                                                               &
          NEUROFlag                                                                                 &
          \multicolumn{1}{c|}{0.98\%}                                                               &
          DMFlag                                                                                    &
          1.09\%                                                                                      \\
          OBESEFlag                                                                                 &
          0.79\%                                                                                    &
          NEUROFlag                                                                                 &
          \multicolumn{1}{c|}{0.78\%}                                                               &
          CHFFlag                                                                                   &
          \multicolumn{1}{c|}{0.79\%}                                                               &
          COAGFlag                                                                                  &
          \multicolumn{1}{c|}{0.83\%}                                                               &
          DEPRESSFlag                                                                               &
          \multicolumn{1}{c|}{0.92\%}                                                               &
          CADFlag                                                                                   &
          \multicolumn{1}{c|}{0.87\%}                                                               &
          WGHTLOSSFlag                                                                              &
          \multicolumn{1}{c|}{0.96\%}                                                               &
          OBESEFlag                                                                                 &
          1.08\%                                                                                      \\
          HX\_LYTES                                                                                 &
          0.77\%                                                                                    &
          CHRNLUNGFlag                                                                              &
          \multicolumn{1}{c|}{0.78\%}                                                               &
          DEPRESSFlag                                                                               &
          \multicolumn{1}{c|}{0.79\%}                                                               &
          CHFFlag                                                                                   &
          \multicolumn{1}{c|}{0.83\%}                                                               &
          DMFlag                                                                                    &
          \multicolumn{1}{c|}{0.80\%}                                                               &
          COAGFlag                                                                                  &
          \multicolumn{1}{c|}{0.78\%}                                                               &
          DEPRESSFlag                                                                               &
          \multicolumn{1}{c|}{0.92\%}                                                               &
          HYPOTHYFlag                                                                               &
          1.01\%                                                                                      \\
          HYPOTHYFlag                                                                               &
          0.73\%                                                                                    &
          HX\_OBESE                                                                                 &
          \multicolumn{1}{c|}{0.74\%}                                                               &
          HX\_Sepsis                                                                                &
          \multicolumn{1}{c|}{0.78\%}                                                               &
          DEPRESSFlag                                                                               &
          \multicolumn{1}{c|}{0.81\%}                                                               &
          HX\_Sepsis                                                                                &
          \multicolumn{1}{c|}{0.73\%}                                                               &
          OBESEFlag                                                                                 &
          \multicolumn{1}{c|}{0.76\%}                                                               &
          HYPOTHYFlag                                                                               &
          \multicolumn{1}{c|}{0.90\%}                                                               &
          CHFFlag                                                                                   &
          0.98\%                                                                                      \\
          \cellcolor[HTML]{53E8BF}\begin{tabular}[c]{@{}l@{}}EthnicGroup\\ Description\end{tabular} &
          0.73\%                                                                                    &
          HYPOTHYFlag                                                                               &
          \multicolumn{1}{c|}{0.73\%}                                                               &
          HYPOTHYFlag                                                                               &
          \multicolumn{1}{c|}{0.78\%}                                                               &
          HYPOTHYFlag                                                                               &
          \multicolumn{1}{c|}{0.74\%}                                                               &
          CHFFlag                                                                                   &
          \multicolumn{1}{c|}{0.72\%}                                                               &
          HYPOTHYFlag                                                                               &
          \multicolumn{1}{c|}{0.75\%}                                                               &
          COAGFlag                                                                                  &
          \multicolumn{1}{c|}{0.84\%}                                                               &
          COAGFlag                                                                                  &
          0.92\%                                                                                      \\
          COAGFlag                                                                                  &
          0.72\%                                                                                    &
          RENLFAILFlag                                                                              &
          \multicolumn{1}{c|}{0.72\%}                                                               &
          WGHTLOSSFlag                                                                              &
          \multicolumn{1}{c|}{0.77\%}                                                               &
          CADFlag                                                                                   &
          \multicolumn{1}{c|}{0.71\%}                                                               &
          HYPOTHYFlag                                                                               &
          \multicolumn{1}{c|}{0.71\%}                                                               &
          RENLFAILFlag                                                                              &
          \multicolumn{1}{c|}{0.73\%}                                                               &
          CHFFlag                                                                                   &
          \multicolumn{1}{c|}{0.79\%}                                                               &
          \cellcolor[HTML]{83E28E}
\begin{tabular}[c]{@{}l@{}}Race\\ Description\end{tabular}        &
          0.92\%                                                                                      \\
          CADFlag                                                                                   &
          0.72\%                                                                                    &
          CHFFlag                                                                                   &
          \multicolumn{1}{c|}{0.71\%}                                                               &
          OBESEFlag                                                                                 &
          \multicolumn{1}{c|}{0.76\%}                                                               &
          WGHTLOSSFlag                                                                              &
          \multicolumn{1}{c|}{0.69\%}                                                               &
          \cellcolor[HTML]{53E8BF}\begin{tabular}[c]{@{}l@{}}EthnicGroup\\ Description\end{tabular} &
          \multicolumn{1}{c|}{0.71\%}                                                               &
          CHFFlag                                                                                   &
          \multicolumn{1}{c|}{0.69\%}                                                               &
          HX\_LYTES                                                                                 &
          \multicolumn{1}{c|}{0.79\%}                                                               &
          HX\_ANEMDEF                                                                               &
          0.87\%                                                                                      \\
          30.49\%                                                                                   &
          \begin{tabular}[c]{@{}l@{}}Antibiotic\\ \_AdminFlag\end{tabular}                          &
          \multicolumn{1}{c|}{30.24\%}                                                              &
          \begin{tabular}[c]{@{}l@{}}Antibiotic\\ \_AdminFlag\end{tabular}                          &
          \multicolumn{1}{c|}{30.08\%}                                                              &
          \begin{tabular}[c]{@{}l@{}}Antibiotic\\ \_AdminFlag\end{tabular}                          &
          \multicolumn{1}{c|}{30.11\%}                                                              &
          \begin{tabular}[c]{@{}l@{}}Antibiotic\\ \_AdminFlag\end{tabular}                          &
          \multicolumn{1}{c|}{30.15\%}                                                              &
          \begin{tabular}[c]{@{}l@{}}Antibiotic\\ \_AdminFlag\end{tabular}                          &
          \multicolumn{1}{c|}{31.29\%}                                                              &
          \begin{tabular}[c]{@{}l@{}}Antibiotic\\ \_AdminFlag\end{tabular}                          &
          \multicolumn{1}{c|}{31.51\%}                                                              &
          \begin{tabular}[c]{@{}l@{}}Antibiotic\\ \_AdminFlag\end{tabular}                          &
          32.11\%                                                                                     \\
          \cellcolor[HTML]{83CCEB}AgeCategory                                                       &
          11.00\%                                                                                   &
          \cellcolor[HTML]{83CCEB}AgeCategory                                                       &
          \multicolumn{1}{c|}{10.70\%}                                                              &
          \cellcolor[HTML]{83CCEB}AgeCategory                                                       &
          \multicolumn{1}{c|}{10.18\%}                                                              &
          \cellcolor[HTML]{83CCEB}AgeCategory                                                       &
          \multicolumn{1}{c|}{9.81\%}                                                               &
          \cellcolor[HTML]{F7C7AC}LOSDays                                                           &
          \multicolumn{1}{c|}{9.94\%}                                                               &
          LYTESFlag                                                                                 &
          \multicolumn{1}{c|}{3.84\%}                                                               &
          \begin{tabular}[c]{@{}l@{}}FirstLocation\\ TypeCodeAfter\\ Arrival\end{tabular}           &
          \multicolumn{1}{c|}{4.28\%}                                                               &
          \begin{tabular}[c]{@{}l@{}}FirstLocation\\ TypeCodeAfter\\ Arrival\end{tabular}           &
          3.94\%                                                                                      \\
          \cellcolor[HTML]{F7C7AC}LOSDays                                                           &
          8.64\%                                                                                    &
          \cellcolor[HTML]{F7C7AC}LOSDays                                                           &
          \multicolumn{1}{c|}{8.39\%}                                                               &
          \cellcolor[HTML]{F7C7AC}LOSDays                                                           &
          \multicolumn{1}{c|}{8.44\%}                                                               &
          \cellcolor[HTML]{F7C7AC}LOSDays                                                           &
          \multicolumn{1}{c|}{8.04\%}                                                               &
          \cellcolor[HTML]{83CCEB}AgeCategory                                                       &
          \multicolumn{1}{c|}{9.12\%}                                                               &
          \begin{tabular}[c]{@{}l@{}}FirstLocation\\ TypeCodeAfter\\ Arrival\end{tabular}           &
          \multicolumn{1}{c|}{3.66\%}                                                               &
          \cellcolor[HTML]{F7C7AC}LOSDays                                                           &
          \multicolumn{1}{c|}{3.32\%}                                                               &
          LYTESFlag                                                                                 &
          3.80\%                                                                                      \\
          LYTESFlag                                                                                 &
          3.72\%                                                                                    &
          LYTESFlag                                                                                 &
          \multicolumn{1}{c|}{3.80\%}                                                               &
          LYTESFlag                                                                                 &
          \multicolumn{1}{c|}{3.70\%}                                                               &
          LYTESFlag                                                                                 &
          \multicolumn{1}{c|}{3.73\%}                                                               &
          \cellcolor[HTML]{FFFD78}NumberofVisits                                                    &
          \multicolumn{1}{c|}{5.49\%}                                                               &
          \cellcolor[HTML]{F7C7AC}LOSDays                                                           &
          \multicolumn{1}{c|}{2.81\%}                                                               &
          LYTESFlag                                                                                 &
          \multicolumn{1}{c|}{2.63\%}                                                               &
          FluSeasonFlag                                                                             &
          2.52\%                                                                                      \\
          \begin{tabular}[c]{@{}l@{}}FirstLocation\\ TypeCodeAfter\\ Arrival\end{tabular}           &
          3.42\%                                                                                    &
          \begin{tabular}[c]{@{}l@{}}FirstLocation\\ TypeCodeAfter\\ Arrival\end{tabular}           &
          \multicolumn{1}{c|}{3.40\%}                                                               &
          \begin{tabular}[c]{@{}l@{}}FirstLocation\\ TypeCodeAfter\\ Arrival\end{tabular}           &
          \multicolumn{1}{c|}{3.35\%}                                                               &
          \begin{tabular}[c]{@{}l@{}}FirstLocation\\ TypeCodeAfter\\ Arrival\end{tabular}           &
          \multicolumn{1}{c|}{3.26\%}                                                               &
          LYTESFlag                                                                                 &
          \multicolumn{1}{c|}{3.63\%}                                                               &
          FluSeasonFlag                                                                             &
          \multicolumn{1}{c|}{2.31\%}                                                               &
          \begin{tabular}[c]{@{}l@{}}\textgreater{}6HoursToFirst\\ AntibioticAdmin\end{tabular}     &
          \multicolumn{1}{c|}{2.26\%}                                                               &
          \begin{tabular}[c]{@{}l@{}}\textgreater{}6HoursToFirst\\ AntibioticAdmin\end{tabular}     &
          2.13\%                                                                                      \\
          \begin{tabular}[c]{@{}l@{}}\textgreater{}6HoursToFirst\\ AntibioticAdmin\end{tabular}     &
          2.28\%                                                                                    &
          \cellcolor[HTML]{FFFD78}NumberofVisits                                                    &
          \multicolumn{1}{c|}{2.58\%}                                                               &
          \cellcolor[HTML]{FFFD78}NumberofVisits                                                    &
          \multicolumn{1}{c|}{2.78\%}                                                               &
          \cellcolor[HTML]{FFFD78}NumberofVisits                                                    &
          \multicolumn{1}{c|}{2.94\%}                                                               &
          \begin{tabular}[c]{@{}l@{}}FirstLocation\\ TypeCodeAfter\\ Arrival\end{tabular}           &
          \multicolumn{1}{c|}{3.22\%}                                                               &
          \begin{tabular}[c]{@{}l@{}}\textgreater{}6HoursToFirst\\ AntibioticAdmin\end{tabular}     &
          \multicolumn{1}{c|}{2.22\%}                                                               &
          FluSeasonFlag                                                                             &
          \multicolumn{1}{c|}{2.18\%}                                                               &
          HTNFlag                                                                                   &
          1.69\%                                                                                      \\
          \cellcolor[HTML]{83E28E}\begin{tabular}[c]{@{}l@{}}Race\\ Description\end{tabular}        &
          1.81\%                                                                                    &
          \begin{tabular}[c]{@{}l@{}}\textgreater{}6HoursToFirst\\ AntibioticAdmin\end{tabular}     &
          \multicolumn{1}{c|}{2.22\%}                                                               &
          \begin{tabular}[c]{@{}l@{}}\textgreater{}6HoursToFirst\\ AntibioticAdmin\end{tabular}     &
          \multicolumn{1}{c|}{2.21\%}                                                               &
          \begin{tabular}[c]{@{}l@{}}\textgreater{}6HoursToFirst\\ AntibioticAdmin\end{tabular}     &
          \multicolumn{1}{c|}{2.18\%}                                                               &
          \begin{tabular}[c]{@{}l@{}}\textgreater{}6HoursToFirst\\ AntibioticAdmin\end{tabular}     &
          \multicolumn{1}{c|}{2.33\%}                                                               &
          \cellcolor[HTML]{83E28E}\begin{tabular}[c]{@{}l@{}}Race\\ Description\end{tabular}        &
          \multicolumn{1}{c|}{1.92\%}                                                               &
          HX\_BLDLOSS                                                                               &
          \multicolumn{1}{c|}{1.51\%}                                                               &
          ANEMDEFFlag                                                                               &
          1.66\%                                                                                      \\
          \cellcolor[HTML]{E49EDD}\begin{tabular}[c]{@{}l@{}}Gender\\ Description\end{tabular}      &
          1.61\%                                                                                    &
          FluSeasonFlag                                                                             &
          \multicolumn{1}{c|}{1.53\%}                                                               &
          FluSeasonFlag                                                                             &
          \multicolumn{1}{c|}{1.65\%}                                                               &
          FluSeasonFlag                                                                             &
          \multicolumn{1}{c|}{1.47\%}                                                               &
          FluSeasonFlag                                                                             &
          \multicolumn{1}{c|}{1.73\%}                                                               &
          \cellcolor[HTML]{E49EDD}\begin{tabular}[c]{@{}l@{}}Gender\\ Description\end{tabular}      &
          \multicolumn{1}{c|}{1.71\%}                                                               &
          HTNFlag                                                                                   &
          \multicolumn{1}{c|}{1.47\%}                                                               &
          CHRNLUNGFlag                                                                              &
          1.60\%                                                                                      \\
          \cellcolor[HTML]{FFFD78}NumberofVisits                                                    &
          1.58\%                                                                                    &
          HX\_BLDLOSS                                                                               &
          \multicolumn{1}{c|}{1.34\%}                                                               &
          \cellcolor[HTML]{83E28E}\begin{tabular}[c]{@{}l@{}}Race\\ Description\end{tabular}        &
          \multicolumn{1}{c|}{1.45\%}                                                               &
          \cellcolor[HTML]{E49EDD}\begin{tabular}[c]{@{}l@{}}Gender\\ Description\end{tabular}      &
          \multicolumn{1}{c|}{1.41\%}                                                               &
          HTNFlag                                                                                   &
          \multicolumn{1}{c|}{1.15\%}                                                               &
          \cellcolor[HTML]{83CCEB}AgeCategory                                                       &
          \multicolumn{1}{c|}{1.71\%}                                                               &
          HX\_CAD                                                                                   &
          \multicolumn{1}{c|}{1.43\%}                                                               &
          DMFlag                                                                                    &
          1.51\%                                                                                      \\
          FluSeasonFlag                                                                             &
          1.53\%                                                                                    &
          \cellcolor[HTML]{E49EDD}\begin{tabular}[c]{@{}l@{}}Gender\\ Description\end{tabular}      &
          \multicolumn{1}{c|}{1.26\%}                                                               &
          \cellcolor[HTML]{E49EDD}\begin{tabular}[c]{@{}l@{}}Gender\\ Description\end{tabular}      &
          \multicolumn{1}{c|}{1.26\%}                                                               &
          HX\_BLDLOSS                                                                               &
          \multicolumn{1}{c|}{1.37\%}                                                               &
          \cellcolor[HTML]{E49EDD}\begin{tabular}[c]{@{}l@{}}Gender\\ Description\end{tabular}      &
          \multicolumn{1}{c|}{1.12\%}                                                               &
          HTNFlag                                                                                   &
          \multicolumn{1}{c|}{1.70\%}                                                               &
          ANEMDEFFlag                                                                               &
          \multicolumn{1}{c|}{1.43\%}                                                               &
          NEUROFlag                                                                                 &
          1.40\%                                                                                      \\
          HX\_BLDLOSS                                                                               &
          1.39\%                                                                                    &
          \cellcolor[HTML]{83E28E}\begin{tabular}[c]{@{}l@{}}Race\\ Description\end{tabular}        &
          \multicolumn{1}{c|}{1.24\%}                                                               &
          HX\_BLDLOSS                                                                               &
          \multicolumn{1}{c|}{1.24\%}                                                               &
          HTNFlag                                                                                   &
          \multicolumn{1}{c|}{1.26\%}                                                               &
          \cellcolor[HTML]{83E28E}\begin{tabular}[c]{@{}l@{}}Race\\ Description\end{tabular}        &
          \multicolumn{1}{c|}{1.06\%}                                                               &
          DEPRESSFlag                                                                               &
          \multicolumn{1}{c|}{1.43\%}                                                               &
          CHRNLUNGFlag                                                                              &
          \multicolumn{1}{c|}{1.38\%}                                                               &
          HX\_BLDLOSS                                                                               &
          1.40\%                                                                                      \\
          CHRNLUNGFlag                                                                              &
          1.03\%                                                                                    &
          ANEMDEFFlag                                                                               &
          \multicolumn{1}{c|}{1.14\%}                                                               &
          ANEMDEFFlag                                                                               &
          \multicolumn{1}{c|}{1.13\%}                                                               &
          \cellcolor[HTML]{83E28E}\begin{tabular}[c]{@{}l@{}}Race\\ Description\end{tabular}        &
          \multicolumn{1}{c|}{1.16\%}                                                               &
          NEUROFlag                                                                                 &
          \multicolumn{1}{c|}{0.92\%}                                                               &
          ANEMDEFFlag                                                                               &
          \multicolumn{1}{c|}{1.43\%}                                                               &
          DMFlag                                                                                    &
          \multicolumn{1}{c|}{1.29\%}                                                               &
          DEPRESSFlag                                                                               &
          1.34\%                                                                                      \\
          HTNFlag                                                                                   &
          1.00\%                                                                                    &
          NEUROFlag                                                                                 &
          \multicolumn{1}{c|}{1.12\%}                                                               &
          HTNFlag                                                                                   &
          \multicolumn{1}{c|}{1.07\%}                                                               &
          NEUROFlag                                                                                 &
          \multicolumn{1}{c|}{1.09\%}                                                               &
          ANEMDEFFlag                                                                               &
          \multicolumn{1}{c|}{0.92\%}                                                               &
          CHRNLUNGFlag                                                                              &
          \multicolumn{1}{c|}{1.38\%}                                                               &
          DEPRESSFlag                                                                               &
          \multicolumn{1}{c|}{1.27\%}                                                               &
          CADFlag                                                                                   &
          1.26\%                                                                                      \\
          ANEMDEFFlag                                                                               &
          0.94\%                                                                                    &
          HTNFlag                                                                                   &
          \multicolumn{1}{c|}{1.07\%}                                                               &
          CHRNLUNGFlag                                                                              &
          \multicolumn{1}{c|}{1.06\%}                                                               &
          ANEMDEFFlag                                                                               &
          \multicolumn{1}{c|}{1.08\%}                                                               &
          CHRNLUNGFlag                                                                              &
          \multicolumn{1}{c|}{0.89\%}                                                               &
          HX\_BLDLOSS                                                                               &
          \multicolumn{1}{c|}{1.37\%}                                                               &
          CADFlag                                                                                   &
          \multicolumn{1}{c|}{1.25\%}                                                               &
          CHFFlag                                                                                   &
          1.24\%                                                                                      \\
          NEUROFlag                                                                                 &
          0.91\%                                                                                    &
          DMFlag                                                                                    &
          \multicolumn{1}{c|}{1.00\%}                                                               &
          NEUROFlag                                                                                 &
          \multicolumn{1}{c|}{1.01\%}                                                               &
          DMFlag                                                                                    &
          \multicolumn{1}{c|}{1.05\%}                                                               &
          CHFFlag                                                                                   &
          \multicolumn{1}{c|}{0.86\%}                                                               &
          NEUROFlag                                                                                 &
          \multicolumn{1}{c|}{1.29\%}                                                               &
          OBESEFlag                                                                                 &
          \multicolumn{1}{c|}{1.20\%}                                                               &
          OBESEFlag                                                                                 &
          1.17\%                                                                                      \\
          CHFFlag                                                                                   &
          0.84\%                                                                                    &
          DEPRESSFlag                                                                               &
          \multicolumn{1}{c|}{0.96\%}                                                               &
          OBESEFlag                                                                                 &
          \multicolumn{1}{c|}{0.94\%}                                                               &
          CHRNLUNGFlag                                                                              &
          \multicolumn{1}{c|}{0.99\%}                                                               &
          DEPRESSFlag                                                                               &
          \multicolumn{1}{c|}{0.84\%}                                                               &
          CADFlag                                                                                   &
          \multicolumn{1}{c|}{1.24\%}                                                               &
          COAGFlag                                                                                  &
          \multicolumn{1}{c|}{1.11\%}                                                               &
          DMCXFlag                                                                                  &
          1.04\%                                                                                      \\
          DMFlag                                                                                    &
          0.84\%                                                                                    &
          CHRNLUNGFlag                                                                              &
          \multicolumn{1}{c|}{0.95\%}                                                               &
          HYPOTHYFlag                                                                               &
          \multicolumn{1}{c|}{0.88\%}                                                               &
          DEPRESSFlag                                                                               &
          \multicolumn{1}{c|}{0.90\%}                                                               &
          CADFlag                                                                                   &
          \multicolumn{1}{c|}{0.83\%}                                                               &
          COAGFlag                                                                                  &
          \multicolumn{1}{c|}{1.19\%}                                                               &
          HYPOTHYFlag                                                                               &
          \multicolumn{1}{c|}{1.09\%}                                                               &
          WGHTLOSSFlag                                                                              &
          1.01\%                                                                                      \\
          OBESEFlag                                                                                 &
          0.79\%                                                                                    &
          HX\_Sepsis                                                                                &
          \multicolumn{1}{c|}{0.81\%}                                                               &
          CADFlag                                                                                   &
          \multicolumn{1}{c|}{0.85\%}                                                               &
          CHFFlag                                                                                   &
          \multicolumn{1}{c|}{0.84\%}                                                               &
          HX\_Sepsis                                                                                &
          \multicolumn{1}{c|}{0.80\%}                                                               &
          RENLFAILFlag                                                                              &
          \multicolumn{1}{c|}{1.12\%}                                                               &
          \cellcolor[HTML]{83CCEB}AgeCategory                                                       &
          \multicolumn{1}{c|}{1.06\%}                                                               &
          RENLFAILFlag                                                                              &
          0.96\%                                                                                      \\
          HX\_LYTES                                                                                 &
          0.77\%                                                                                    &
          RENLFAILFlag                                                                              &
          \multicolumn{1}{c|}{0.78\%}                                                               &
          WGHTLOSSFlag                                                                              &
          \multicolumn{1}{c|}{0.77\%}                                                               &
          CADFlag                                                                                   &
          \multicolumn{1}{c|}{0.84\%}                                                               &
          COAGFlag                                                                                  &
          \multicolumn{1}{c|}{0.77\%}                                                               &
          DMFlag                                                                                    &
          \multicolumn{1}{c|}{1.02\%}                                                               &
          NEUROFlag                                                                                 &
          \multicolumn{1}{c|}{1.03\%}                                                               &
          HX\_CHRNLUNG                                                                              &
          0.96\%                                                                                      \\
          HYPOTHYFlag                                                                               &
          0.73\%                                                                                    &
          COAGFlag                                                                                  &
          \multicolumn{1}{c|}{0.73\%}                                                               &
          CHFFlag                                                                                   &
          \multicolumn{1}{c|}{0.74\%}                                                               &
          WGHTLOSSFlag                                                                              &
          \multicolumn{1}{c|}{0.81\%}                                                               &
          DMFlag                                                                                    &
          \multicolumn{1}{c|}{0.74\%}                                                               &
          HYPOTHYFlag                                                                               &
          \multicolumn{1}{c|}{0.98\%}                                                               &
          CHFFlag                                                                                   &
          \multicolumn{1}{c|}{1.01\%}                                                               &
          HX\_CAD                                                                                   &
          0.96\%                                                                                      \\
          \cellcolor[HTML]{53E8BF}\begin{tabular}[c]{@{}l@{}}EthnicGroup\\ Description\end{tabular} &
          0.73\%                                                                                    &
          CHFFlag                                                                                   &
          \multicolumn{1}{c|}{0.72\%}                                                               &
          DEPRESSFlag                                                                               &
          \multicolumn{1}{c|}{0.73\%}                                                               &
          HX\_LYTES                                                                                 &
          \multicolumn{1}{c|}{0.79\%}                                                               &
          HX\_OBESE                                                                                 &
          \multicolumn{1}{c|}{0.73\%}                                                               &
          CHFFlag                                                                                   &
          \multicolumn{1}{c|}{0.95\%}                                                               &
          HX\_HTN                                                                                   &
          \multicolumn{1}{c|}{0.98\%}                                                               &
          COAGFlag                                                                                  &
          0.92\%                                                                                      \\
          COAGFlag                                                                                  &
          0.72\%                                                                                    &
          HX\_LYTES                                                                                 &
          \multicolumn{1}{c|}{0.70\%}                                                               &
          DMFlag                                                                                    &
          \multicolumn{1}{c|}{0.73\%}                                                               &
          COAGFlag                                                                                  &
          \multicolumn{1}{c|}{0.75\%}                                                               &
          OBESEFlag                                                                                 &
          \multicolumn{1}{c|}{0.72\%}                                                               &
          OBESEFlag                                                                                 &
          \multicolumn{1}{c|}{0.87\%}                                                               &
          HX\_CHRNLUNG                                                                              &
          \multicolumn{1}{c|}{0.94\%}                                                               &
          HX\_HTN                                                                                   &
          0.91\%                                                                                      \\
          CADFlag                                                                                   &
          0.72\%                                                                                    &
          WGHTLOSSFlag                                                                              &
          \multicolumn{1}{c|}{0.69\%}                                                               &
          HX\_LYTES                                                                                 &
          \multicolumn{1}{c|}{0.72\%}                                                               &
          OBESEFlag                                                                                 &
          \multicolumn{1}{c|}{0.70\%}                                                               &
          HX\_DM                                                                                    &
          \multicolumn{1}{c|}{0.70\%}                                                               &
          \cellcolor[HTML]{53E8BF}\begin{tabular}[c]{@{}l@{}}EthnicGroup\\ Description\end{tabular} &
          \multicolumn{1}{c|}{0.85\%}                                                               &
          HX\_LYTES                                                                                 &
          \multicolumn{1}{c|}{0.93\%}                                                               &
          HX\_ANEMDEF                                                                               &
          0.91\%                                                                                      \\
          HX\_HTN                                                                                   &
          0.71\%                                                                                    &
          HYPOTHYFlag                                                                               &
          \multicolumn{1}{c|}{0.66\%}                                                               &
          HX\_Sepsis                                                                                &
          \multicolumn{1}{c|}{0.68\%}                                                               &
          HYPOTHYFlag                                                                               &
          \multicolumn{1}{c|}{0.68\%}                                                               &
          HYPOTHYFlag                                                                               &
          \multicolumn{1}{c|}{0.69\%}                                                               &
          HX\_HTN                                                                                   &
          \multicolumn{1}{c|}{0.85\%}                                                               &
          \cellcolor[HTML]{FFFD78}NumberofVisits                                                    &
          \multicolumn{1}{c|}{0.92\%}                                                               &
          PERIVASCFlag                                                                              &
          0.89\%                                                                                      \\
          HX\_OBESE                                                                                 &
          0.62\%                                                                                    &
          \begin{tabular}[c]{@{}l@{}}1-3HoursToFirst\\ AntibioticAdmin\end{tabular}                 &
          \multicolumn{1}{c|}{0.65\%}                                                               &
          COAGFlag                                                                                  &
          \multicolumn{1}{c|}{0.65\%}                                                               &
          HX\_OBESE                                                                                 &
          \multicolumn{1}{c|}{0.68\%}                                                               &
          WGHTLOSSFlag                                                                              &
          \multicolumn{1}{c|}{0.61\%}                                                               &
          \cellcolor[HTML]{FFFD78}NumberofVisits                                                    &
          \multicolumn{1}{c|}{0.83\%}                                                               &
          HX\_ANEMDEF                                                                               &
          \multicolumn{1}{c|}{0.90\%}                                                               &
          HYPOTHYFlag                                                                               &
          0.89\%
        \end{tabular}%
      }
    \end{tabular}
  \end{table*}

  \begin{table*}[!ht]
    \caption{Variation of feature importances in the PSM-adjusted Sepsis patient dataset with different levels of k-anonymity. Cells highlighted with color coding denote QIs}
    \label{tab: Sepsis PSM FI}
    \begin{tabular}{c}
      (a) $k$-anonymity \\
      \resizebox{\textwidth}{!}{%
        \begin{tabular}{m{2cm}c|m{2cm}cm{2cm}cm{2cm}cm{2cm}cm{2cm}cm{2cm}cm{2cm}c}
          \hline
          \multicolumn{2}{c|}{Original   Data}                                                      &
          \multicolumn{14}{c}{k-anonymity}                                                            \\ \hline
          Feature                                                                                   &
          Importance                                                                                &
          $k$=5                                                                                     &
          \multicolumn{1}{c|}{Importance}                                                           &
          $k$=10                                                                                    &
          \multicolumn{1}{c|}{Importance}                                                           &
          $k$=15                                                                                    &
          \multicolumn{1}{c|}{Importance}                                                           &
          $k$=20                                                                                    &
          \multicolumn{1}{c|}{Importance}                                                           &
          $k$=100                                                                                   &
          \multicolumn{1}{c|}{Importance}                                                           &
          $k$=300                                                                                   &
          \multicolumn{1}{c|}{Importance}                                                           &
          $k$=2000                                                                                  &
          Importance                                                                                  \\ \hline
          \begin{tabular}[c]{@{}l@{}}Antibiotic\\ \_AdminFlag\end{tabular}                          &
          37.96\%                                                                                   &
          \begin{tabular}[c]{@{}l@{}}Antibiotic\\ \_AdminFlag\end{tabular}                          &
          \multicolumn{1}{c|}{38.38\%}                                                              &
          \begin{tabular}[c]{@{}l@{}}Antibiotic\\ \_AdminFlag\end{tabular}                          &
          \multicolumn{1}{c|}{38.56\%}                                                              &
          \begin{tabular}[c]{@{}l@{}}Antibiotic\\ \_AdminFlag\end{tabular}                          &
          \multicolumn{1}{c|}{38.33\%}                                                              &
          \begin{tabular}[c]{@{}l@{}}Antibiotic\\ \_AdminFlag\end{tabular}                          &
          \multicolumn{1}{c|}{38.30\%}                                                              &
          \begin{tabular}[c]{@{}l@{}}Antibiotic\\ \_AdminFlag\end{tabular}                          &
          \multicolumn{1}{c|}{38.34\%}                                                              &
          \begin{tabular}[c]{@{}l@{}}Antibiotic\\ \_AdminFlag\end{tabular}                          &
          \multicolumn{1}{c|}{38.08\%}                                                              &
          \begin{tabular}[c]{@{}l@{}}Antibiotic\\ \_AdminFlag\end{tabular}                          &
          38.18\%                                                                                     \\
          \cellcolor[HTML]{83CCEB}AgeCategory                                                       &
          9.95\%                                                                                    &
          \cellcolor[HTML]{F7C7AC}LOSDays                                                           &
          \multicolumn{1}{c|}{9.09\%}                                                               &
          \cellcolor[HTML]{F7C7AC}LOSDays                                                           &
          \multicolumn{1}{c|}{8.90\%}                                                               &
          \cellcolor[HTML]{83CCEB}AgeCategory                                                       &
          \multicolumn{1}{c|}{9.97\%}                                                               &
          \cellcolor[HTML]{F7C7AC}LOSDays                                                           &
          \multicolumn{1}{c|}{9.08\%}                                                               &
          \cellcolor[HTML]{83CCEB}AgeCategory                                                       &
          \multicolumn{1}{c|}{8.58\%}                                                               &
          \cellcolor[HTML]{F7C7AC}LOSDays                                                           &
          \multicolumn{1}{c|}{7.86\%}                                                               &
          \cellcolor[HTML]{83CCEB}AgeCategory                                                       &
          5.04\%                                                                                      \\
          \cellcolor[HTML]{F7C7AC}LOSDays                                                           &
          9.09\%                                                                                    &
          \cellcolor[HTML]{83CCEB}AgeCategory                                                       &
          \multicolumn{1}{c|}{8.90\%}                                                               &
          \cellcolor[HTML]{83CCEB}AgeCategory                                                       &
          \multicolumn{1}{c|}{8.79\%}                                                               &
          \cellcolor[HTML]{F7C7AC}LOSDays                                                           &
          \multicolumn{1}{c|}{8.92\%}                                                               &
          \cellcolor[HTML]{83CCEB}AgeCategory                                                       &
          \multicolumn{1}{c|}{9.07\%}                                                               &
          \cellcolor[HTML]{F7C7AC}LOSDays                                                           &
          \multicolumn{1}{c|}{8.35\%}                                                               &
          \cellcolor[HTML]{83CCEB}AgeCategory                                                       &
          \multicolumn{1}{c|}{7.05\%}                                                               &
          \cellcolor[HTML]{F7C7AC}LOSDays                                                           &
          4.99\%                                                                                      \\
          \begin{tabular}[c]{@{}l@{}}FirstLocation\\ TypeCodeAfter\\ Arrival\end{tabular}           &
          3.30\%                                                                                    &
          \begin{tabular}[c]{@{}l@{}}FirstLocation\\ TypeCodeAfter\\ Arrival\end{tabular}           &
          \multicolumn{1}{c|}{3.15\%}                                                               &
          \begin{tabular}[c]{@{}l@{}}FirstLocation\\ TypeCodeAfter\\ Arrival\end{tabular}           &
          \multicolumn{1}{c|}{3.22\%}                                                               &
          \begin{tabular}[c]{@{}l@{}}FirstLocation\\ TypeCodeAfter\\ Arrival\end{tabular}           &
          \multicolumn{1}{c|}{3.32\%}                                                               &
          \begin{tabular}[c]{@{}l@{}}FirstLocation\\ TypeCodeAfter\\ Arrival\end{tabular}           &
          \multicolumn{1}{c|}{3.35\%}                                                               &
          \begin{tabular}[c]{@{}l@{}}FirstLocation\\ TypeCodeAfter\\ Arrival\end{tabular}           &
          \multicolumn{1}{c|}{3.62\%}                                                               &
          \begin{tabular}[c]{@{}l@{}}FirstLocation\\ TypeCodeAfter\\ Arrival\end{tabular}           &
          \multicolumn{1}{c|}{3.35\%}                                                               &
          \begin{tabular}[c]{@{}l@{}}FirstLocation\\ TypeCodeAfter\\ Arrival\end{tabular}           &
          3.70\%                                                                                      \\
          \begin{tabular}[c]{@{}l@{}}\textgreater{}6HoursToFirst\\ AntibioticAdmin\end{tabular}     &
          1.93\%                                                                                    &
          \begin{tabular}[c]{@{}l@{}}\textgreater{}6HoursToFirst\\ AntibioticAdmin\end{tabular}     &
          \multicolumn{1}{c|}{1.94\%}                                                               &
          \begin{tabular}[c]{@{}l@{}}\textgreater{}6HoursToFirst\\ AntibioticAdmin\end{tabular}     &
          \multicolumn{1}{c|}{2.06\%}                                                               &
          \begin{tabular}[c]{@{}l@{}}\textgreater{}6HoursToFirst\\ AntibioticAdmin\end{tabular}     &
          \multicolumn{1}{c|}{1.98\%}                                                               &
          \begin{tabular}[c]{@{}l@{}}\textgreater{}6HoursToFirst\\ AntibioticAdmin\end{tabular}     &
          \multicolumn{1}{c|}{2.07\%}                                                               &
          \begin{tabular}[c]{@{}l@{}}\textgreater{}6HoursToFirst\\ AntibioticAdmin\end{tabular}     &
          \multicolumn{1}{c|}{2.03\%}                                                               &
          \begin{tabular}[c]{@{}l@{}}\textgreater{}6HoursToFirst\\ AntibioticAdmin\end{tabular}     &
          \multicolumn{1}{c|}{2.22\%}                                                               &
          FluSeasonFlag                                                                             &
          2.03\%                                                                                      \\
          LYTESFlag                                                                                 &
          1.75\%                                                                                    &
          \cellcolor[HTML]{FFFF00}NumberofVisits                                                    &
          \multicolumn{1}{c|}{1.72\%}                                                               &
          \cellcolor[HTML]{83E28E}\begin{tabular}[c]{@{}l@{}}Gender\\ Description\end{tabular}      &
          \multicolumn{1}{c|}{1.67\%}                                                               &
          \cellcolor[HTML]{FFFF00}NumberofVisits                                                    &
          \multicolumn{1}{c|}{1.74\%}                                                               &
          \cellcolor[HTML]{FFFF00}NumberofVisits                                                    &
          \multicolumn{1}{c|}{1.84\%}                                                               &
          LYTESFlag                                                                                 &
          \multicolumn{1}{c|}{1.78\%}                                                               &
          LYTESFlag                                                                                 &
          \multicolumn{1}{c|}{1.72\%}                                                               &
          \begin{tabular}[c]{@{}l@{}}\textgreater{}6HoursToFirst\\ AntibioticAdmin\end{tabular}     &
          1.99\%                                                                                      \\
          \cellcolor[HTML]{FFFF00}NumberofVisits                                                    &
          1.73\%                                                                                    &
          LYTESFlag                                                                                 &
          \multicolumn{1}{c|}{1.53\%}                                                               &
          \cellcolor[HTML]{E49EDD}\begin{tabular}[c]{@{}l@{}}Race\\ Description\end{tabular}        &
          \multicolumn{1}{c|}{1.65\%}                                                               &
          \cellcolor[HTML]{E49EDD}\begin{tabular}[c]{@{}l@{}}Race\\ Description\end{tabular}        &
          \multicolumn{1}{c|}{1.60\%}                                                               &
          \cellcolor[HTML]{E49EDD}\begin{tabular}[c]{@{}l@{}}Race\\ Description\end{tabular}        &
          \multicolumn{1}{c|}{1.59\%}                                                               &
          \cellcolor[HTML]{FFFF00}NumberofVisits                                                    &
          \multicolumn{1}{c|}{1.62\%}                                                               &
          FluSeasonFlag                                                                             &
          \multicolumn{1}{c|}{1.27\%}                                                               &
          LYTESFlag                                                                                 &
          1.90\%                                                                                      \\
          \cellcolor[HTML]{E49EDD}\begin{tabular}[c]{@{}l@{}}Race\\ Description\end{tabular}        &
          1.61\%                                                                                    &
          \cellcolor[HTML]{83E28E}\begin{tabular}[c]{@{}l@{}}Gender\\ Description\end{tabular}      &
          \multicolumn{1}{c|}{1.45\%}                                                               &
          \cellcolor[HTML]{FFFF00}NumberofVisits                                                    &
          \multicolumn{1}{c|}{1.63\%}                                                               &
          LYTESFlag                                                                                 &
          \multicolumn{1}{c|}{1.33\%}                                                               &
          LYTESFlag                                                                                 &
          \multicolumn{1}{c|}{1.33\%}                                                               &
          FluSeasonFlag                                                                             &
          \multicolumn{1}{c|}{1.54\%}                                                               &
          \cellcolor[HTML]{83E28E}\begin{tabular}[c]{@{}l@{}}Gender\\ Description\end{tabular}      &
          \multicolumn{1}{c|}{1.27\%}                                                               &
          \cellcolor[HTML]{E49EDD}\begin{tabular}[c]{@{}l@{}}Race\\ Description\end{tabular}        &
          1.49\%                                                                                      \\
          \cellcolor[HTML]{83E28E}\begin{tabular}[c]{@{}l@{}}Gender\\ Description\end{tabular}      &
          1.28\%                                                                                    &
          \cellcolor[HTML]{E49EDD}\begin{tabular}[c]{@{}l@{}}Race\\ Description\end{tabular}        &
          \multicolumn{1}{c|}{1.33\%}                                                               &
          LYTESFlag                                                                                 &
          \multicolumn{1}{c|}{1.46\%}                                                               &
          \cellcolor[HTML]{83E28E}\begin{tabular}[c]{@{}l@{}}Gender\\ Description\end{tabular}      &
          \multicolumn{1}{c|}{1.27\%}                                                               &
          FluSeasonFlag                                                                             &
          \multicolumn{1}{c|}{1.31\%}                                                               &
          \cellcolor[HTML]{83E28E}\begin{tabular}[c]{@{}l@{}}Gender\\ Description\end{tabular}      &
          \multicolumn{1}{c|}{1.30\%}                                                               &
          \cellcolor[HTML]{E49EDD}\begin{tabular}[c]{@{}l@{}}Race\\ Description\end{tabular}        &
          \multicolumn{1}{c|}{1.22\%}                                                               &
          ANEMDEFFlag                                                                               &
          1.39\%                                                                                      \\
          FluSeasonFlag                                                                             &
          1.21\%                                                                                    &
          FluSeasonFlag                                                                             &
          \multicolumn{1}{c|}{1.30\%}                                                               &
          ANEMDEFFlag                                                                               &
          \multicolumn{1}{c|}{1.42\%}                                                               &
          FluSeasonFlag                                                                             &
          \multicolumn{1}{c|}{1.13\%}                                                               &
          \cellcolor[HTML]{83E28E}\begin{tabular}[c]{@{}l@{}}Gender\\ Description\end{tabular}      &
          \multicolumn{1}{c|}{1.25\%}                                                               &
          \cellcolor[HTML]{E49EDD}\begin{tabular}[c]{@{}l@{}}Race\\ Description\end{tabular}        &
          \multicolumn{1}{c|}{1.22\%}                                                               &
          \cellcolor[HTML]{FFFF00}NumberofVisits                                                    &
          \multicolumn{1}{c|}{1.22\%}                                                               &
          HTNFlag                                                                                   &
          1.38\%                                                                                      \\
          ANEMDEFFlag                                                                               &
          1.14\%                                                                                    &
          COAGFlag                                                                                  &
          \multicolumn{1}{c|}{1.17\%}                                                               &
          FluSeasonFlag                                                                             &
          \multicolumn{1}{c|}{1.26\%}                                                               &
          CHRNLUNGFlag                                                                              &
          \multicolumn{1}{c|}{1.12\%}                                                               &
          ANEMDEFFlag                                                                               &
          \multicolumn{1}{c|}{1.05\%}                                                               &
          NEUROFlag                                                                                 &
          \multicolumn{1}{c|}{1.04\%}                                                               &
          DMFlag                                                                                    &
          \multicolumn{1}{c|}{1.08\%}                                                               &
          CHRNLUNGFlag                                                                              &
          1.35\%                                                                                      \\
          HTNFlag                                                                                   &
          1.02\%                                                                                    &
          HTNFlag                                                                                   &
          \multicolumn{1}{c|}{1.10\%}                                                               &
          NEUROFlag                                                                                 &
          \multicolumn{1}{c|}{0.94\%}                                                               &
          ANEMDEFFlag                                                                               &
          \multicolumn{1}{c|}{1.12\%}                                                               &
          NEUROFlag                                                                                 &
          \multicolumn{1}{c|}{0.95\%}                                                               &
          HX\_Sepsis                                                                                &
          \multicolumn{1}{c|}{1.02\%}                                                               &
          CADFlag                                                                                   &
          \multicolumn{1}{c|}{1.06\%}                                                               &
          COAGFlag                                                                                  &
          1.28\%                                                                                      \\
          HX\_DEPRESS                                                                               &
          0.87\%                                                                                    &
          ANEMDEFFlag                                                                               &
          \multicolumn{1}{c|}{1.06\%}                                                               &
          COAGFlag                                                                                  &
          \multicolumn{1}{c|}{0.93\%}                                                               &
          COAGFlag                                                                                  &
          \multicolumn{1}{c|}{1.11\%}                                                               &
          HX\_Sepsis                                                                                &
          \multicolumn{1}{c|}{0.92\%}                                                               &
          ANEMDEFFlag                                                                               &
          \multicolumn{1}{c|}{0.97\%}                                                               &
          ANEMDEFFlag                                                                               &
          \multicolumn{1}{c|}{1.03\%}                                                               &
          NEUROFlag                                                                                 &
          1.25\%                                                                                      \\
          COAGFlag                                                                                  &
          0.83\%                                                                                    &
          DEPRESSFlag                                                                               &
          \multicolumn{1}{c|}{0.96\%}                                                               &
          HTNFlag                                                                                   &
          \multicolumn{1}{c|}{0.92\%}                                                               &
          HTNFlag                                                                                   &
          \multicolumn{1}{c|}{0.97\%}                                                               &
          COAGFlag                                                                                  &
          \multicolumn{1}{c|}{0.92\%}                                                               &
          HTNFlag                                                                                   &
          \multicolumn{1}{c|}{0.94\%}                                                               &
          HTNFlag                                                                                   &
          \multicolumn{1}{c|}{1.02\%}                                                               &
          CHFFlag                                                                                   &
          1.21\%                                                                                      \\
          DMFlag                                                                                    &
          0.79\%                                                                                    &
          NEUROFlag                                                                                 &
          \multicolumn{1}{c|}{0.88\%}                                                               &
          HX\_Sepsis                                                                                &
          \multicolumn{1}{c|}{0.89\%}                                                               &
          HX\_Sepsis                                                                                &
          \multicolumn{1}{c|}{0.94\%}                                                               &
          HTNFlag                                                                                   &
          \multicolumn{1}{c|}{0.86\%}                                                               &
          COAGFlag                                                                                  &
          \multicolumn{1}{c|}{0.93\%}                                                               &
          COAGFlag                                                                                  &
          \multicolumn{1}{c|}{0.99\%}                                                               &
          HX\_Sepsis                                                                                &
          1.17\%                                                                                      \\
          DEPRESSFlag                                                                               &
          0.77\%                                                                                    &
          \cellcolor[HTML]{BFCEF0}\begin{tabular}[c]{@{}l@{}}EthnicGroup\\ Description\end{tabular} &
          \multicolumn{1}{c|}{0.82\%}                                                               &
          CHRNLUNGFlag                                                                              &
          \multicolumn{1}{c|}{0.86\%}                                                               &
          DMFlag                                                                                    &
          \multicolumn{1}{c|}{0.92\%}                                                               &
          CHRNLUNGFlag                                                                              &
          \multicolumn{1}{c|}{0.84\%}                                                               &
          CHFFlag                                                                                   &
          \multicolumn{1}{c|}{0.87\%}                                                               &
          NEUROFlag                                                                                 &
          \multicolumn{1}{c|}{0.99\%}                                                               &
          CADFlag                                                                                   &
          1.00\%                                                                                      \\
          CHRNLUNGFlag                                                                              &
          0.76\%                                                                                    &
          CHRNLUNGFlag                                                                              &
          \multicolumn{1}{c|}{0.75\%}                                                               &
          HX\_CHRNLUNG                                                                              &
          \multicolumn{1}{c|}{0.85\%}                                                               &
          CADFlag                                                                                   &
          \multicolumn{1}{c|}{0.77\%}                                                               &
          DMFlag                                                                                    &
          \multicolumn{1}{c|}{0.81\%}                                                               &
          DMFlag                                                                                    &
          \multicolumn{1}{c|}{0.87\%}                                                               &
          OBESEFlag                                                                                 &
          \multicolumn{1}{c|}{0.94\%}                                                               &
          \cellcolor[HTML]{FFFF00}NumberofVisits                                                    &
          0.94\%                                                                                      \\
          WGHTLOSSFlag                                                                              &
          0.75\%                                                                                    &
          WGHTLOSSFlag                                                                              &
          \multicolumn{1}{c|}{0.74\%}                                                               &
          DMFlag                                                                                    &
          \multicolumn{1}{c|}{0.80\%}                                                               &
          DEPRESSFlag                                                                               &
          \multicolumn{1}{c|}{0.75\%}                                                               &
          CADFlag                                                                                   &
          \multicolumn{1}{c|}{0.73\%}                                                               &
          DEPRESSFlag                                                                               &
          \multicolumn{1}{c|}{0.83\%}                                                               &
          CHFFlag                                                                                   &
          \multicolumn{1}{c|}{0.87\%}                                                               &
          DEPRESSFlag                                                                               &
          0.92\%                                                                                      \\
          HX\_Sepsis                                                                                &
          0.71\%                                                                                    &
          HYPOTHYFlag                                                                               &
          \multicolumn{1}{c|}{0.72\%}                                                               &
          OBESEFlag                                                                                 &
          \multicolumn{1}{c|}{0.79\%}                                                               &
          CHFFlag                                                                                   &
          \multicolumn{1}{c|}{0.73\%}                                                               &
          \cellcolor[HTML]{BFCEF0}\begin{tabular}[c]{@{}l@{}}EthnicGroup\\ Description\end{tabular} &
          \multicolumn{1}{c|}{0.69\%}                                                               &
          CHRNLUNGFlag                                                                              &
          \multicolumn{1}{c|}{0.75\%}                                                               &
          DMCXFlag                                                                                  &
          \multicolumn{1}{c|}{0.86\%}                                                               &
          OBESEFlag                                                                                 &
          0.88\%                                                                                      \\
          \begin{tabular}[c]{@{}l@{}}1-3HoursToFirst\\ AntibioticAdmin\end{tabular}                 &
          0.71\%                                                                                    &
          RENLFAILFlag                                                                              &
          \multicolumn{1}{c|}{0.70\%}                                                               &
          DEPRESSFlag                                                                               &
          \multicolumn{1}{c|}{0.78\%}                                                               &
          NEUROFlag                                                                                 &
          \multicolumn{1}{c|}{0.71\%}                                                               &
          CHFFlag                                                                                   &
          \multicolumn{1}{c|}{0.69\%}                                                               &
          HX\_LYTES                                                                                 &
          \multicolumn{1}{c|}{0.72\%}                                                               &
          HX\_Sepsis                                                                                &
          \multicolumn{1}{c|}{0.85\%}                                                               &
          RENLFAILFlag                                                                              &
          0.88\%                                                                                      \\
          HX\_HTN                                                                                   &
          0.70\%                                                                                    &
          CHFFlag                                                                                   &
          \multicolumn{1}{c|}{0.70\%}                                                               &
          \cellcolor[HTML]{BFCEF0}\begin{tabular}[c]{@{}l@{}}EthnicGroup\\ Description\end{tabular} &
          \multicolumn{1}{c|}{0.65\%}                                                               &
          HX\_RENLFAIL                                                                              &
          \multicolumn{1}{c|}{0.63\%}                                                               &
          DEPRESSFlag                                                                               &
          \multicolumn{1}{c|}{0.67\%}                                                               &
          PERIVASCFlag                                                                              &
          \multicolumn{1}{c|}{0.70\%}                                                               &
          HX\_CAD                                                                                   &
          \multicolumn{1}{c|}{0.85\%}                                                               &
          DMFlag                                                                                    &
          0.85\%                                                                                      \\
          NEUROFlag                                                                                 &
          0.66\%                                                                                    &
          HX\_DM                                                                                    &
          \multicolumn{1}{c|}{0.68\%}                                                               &
          HX\_DM                                                                                    &
          \multicolumn{1}{c|}{0.64\%}                                                               &
          HX\_Uti                                                                                   &
          \multicolumn{1}{c|}{0.62\%}                                                               &
          PULMCIRCFlag                                                                              &
          \multicolumn{1}{c|}{0.65\%}                                                               &
          \cellcolor[HTML]{BFCEF0}\begin{tabular}[c]{@{}l@{}}EthnicGroup\\ Description\end{tabular} &
          \multicolumn{1}{c|}{0.68\%}                                                               &
          CHRNLUNGFlag                                                                              &
          \multicolumn{1}{c|}{0.83\%}                                                               &
          DMCXFlag                                                                                  &
          0.82\%                                                                                      \\
          RENLFAILFlag                                                                              &
          0.66\%                                                                                    &
          PERIVASCFlag                                                                              &
          \multicolumn{1}{c|}{0.68\%}                                                               &
          CADFlag                                                                                   &
          \multicolumn{1}{c|}{0.62\%}                                                               &
          HYPOTHYFlag                                                                               &
          \multicolumn{1}{c|}{0.61\%}                                                               &
          PERIVASCFlag                                                                              &
          \multicolumn{1}{c|}{0.63\%}                                                               &
          HX\_HTN                                                                                   &
          \multicolumn{1}{c|}{0.67\%}                                                               &
          DEPRESSFlag                                                                               &
          \multicolumn{1}{c|}{0.81\%}                                                               &
          PERIVASCFlag                                                                              &
          0.78\%                                                                                      \\
          HX\_ANEMDEF                                                                               &
          0.64\%                                                                                    &
          HX\_Sepsis                                                                                &
          \multicolumn{1}{c|}{0.67\%}                                                               &
          Uti\_AdminFlag                                                                            &
          \multicolumn{1}{c|}{0.61\%}                                                               &
          RENLFAILFlag                                                                              &
          \multicolumn{1}{c|}{0.59\%}                                                               &
          RENLFAILFlag                                                                              &
          \multicolumn{1}{c|}{0.62\%}                                                               &
          HX\_DM                                                                                    &
          \multicolumn{1}{c|}{0.64\%}                                                               &
          HX\_DEPRESS                                                                               &
          \multicolumn{1}{c|}{0.71\%}                                                               &
          PULMCIRCFlag                                                                              &
          0.78\%                                                                                      \\
        \end{tabular}%
      }
    \end{tabular}
  \end{table*}

  \begin{table*}[!ht]
    \centering
    \begin{tabular}{c}
      (b) Algorithm proposed by \citet{zheng2019effective} \\
      \resizebox{\textwidth}{!}{%
        \begin{tabular}{m{2cm}c|m{2cm}cm{2cm}cm{2cm}cm{2cm}cm{2cm}cm{2cm}cm{2cm}c}
          \hline
          \multicolumn{2}{c|}{Original   Data}                                                      &
          \multicolumn{14}{c}{Zheng et al}                                                            \\ \hline
          Feature                                                                                   &
          Importance                                                                                &
          $k$=5,$l$=2                                                                               &
          \multicolumn{1}{c|}{Importance}                                                           &
          $k$=10,$l$=2                                                                              &
          \multicolumn{1}{c|}{Importance}                                                           &
          $k$=15,$l$=2                                                                              &
          \multicolumn{1}{c|}{Importance}                                                           &
          $k$=20,$l$=2                                                                              &
          \multicolumn{1}{c|}{Importance}                                                           &
          $k$=100,$l$=2                                                                             &
          \multicolumn{1}{c|}{Importance}                                                           &
          $k$=300,$l$=2                                                                             &
          \multicolumn{1}{c|}{Importance}                                                           &
          $k$=2000,$l$=2                                                                            &
          Importance                                                                                  \\ \hline
          \begin{tabular}[c]{@{}l@{}}Antibiotic\\ \_AdminFlag\end{tabular}                          &
          37.96\%                                                                                   &
          \begin{tabular}[c]{@{}l@{}}Antibiotic\\ \_AdminFlag\end{tabular}                          &
          \multicolumn{1}{c|}{38.24\%}                                                              &
          \begin{tabular}[c]{@{}l@{}}Antibiotic\\ \_AdminFlag\end{tabular}                          &
          \multicolumn{1}{c|}{38.24\%}                                                              &
          \begin{tabular}[c]{@{}l@{}}Antibiotic\\ \_AdminFlag\end{tabular}                          &
          \multicolumn{1}{c|}{38.19\%}                                                              &
          \begin{tabular}[c]{@{}l@{}}Antibiotic\\ \_AdminFlag\end{tabular}                          &
          \multicolumn{1}{c|}{38.32\%}                                                              &
          \begin{tabular}[c]{@{}l@{}}Antibiotic\\ \_AdminFlag\end{tabular}                          &
          \multicolumn{1}{c|}{38.63\%}                                                              &
          \begin{tabular}[c]{@{}l@{}}Antibiotic\\ \_AdminFlag\end{tabular}                          &
          \multicolumn{1}{c|}{38.09\%}                                                              &
          \begin{tabular}[c]{@{}l@{}}Antibiotic\\ \_AdminFlag\end{tabular}                          &
          38.53\%                                                                                     \\
          \cellcolor[HTML]{83CCEB}AgeCategory                                                       &
          9.95\%                                                                                    &
          \cellcolor[HTML]{83CCEB}AgeCategory                                                       &
          \multicolumn{1}{c|}{9.64\%}                                                               &
          \cellcolor[HTML]{83CCEB}AgeCategory                                                       &
          \multicolumn{1}{c|}{8.68\%}                                                               &
          \cellcolor[HTML]{F7C7AC}LOSDays                                                           &
          \multicolumn{1}{c|}{9.33\%}                                                               &
          \cellcolor[HTML]{83CCEB}AgeCategory                                                       &
          \multicolumn{1}{c|}{9.04\%}                                                               &
          \cellcolor[HTML]{83CCEB}AgeCategory                                                       &
          \multicolumn{1}{c|}{8.93\%}                                                               &
          \cellcolor[HTML]{83CCEB}AgeCategory                                                       &
          \multicolumn{1}{c|}{7.49\%}                                                               &
          \cellcolor[HTML]{83CCEB}AgeCategory                                                       &
          5.17\%                                                                                      \\
          \cellcolor[HTML]{F7C7AC}LOSDays                                                           &
          9.09\%                                                                                    &
          \cellcolor[HTML]{F7C7AC}LOSDays                                                           &
          \multicolumn{1}{c|}{8.21\%}                                                               &
          \cellcolor[HTML]{F7C7AC}LOSDays                                                           &
          \multicolumn{1}{c|}{8.45\%}                                                               &
          \cellcolor[HTML]{83CCEB}AgeCategory                                                       &
          \multicolumn{1}{c|}{9.25\%}                                                               &
          \cellcolor[HTML]{F7C7AC}LOSDays                                                           &
          \multicolumn{1}{c|}{8.53\%}                                                               &
          \cellcolor[HTML]{F7C7AC}LOSDays                                                           &
          \multicolumn{1}{c|}{7.69\%}                                                               &
          \cellcolor[HTML]{F7C7AC}LOSDays                                                           &
          \multicolumn{1}{c|}{6.37\%}                                                               &
          \cellcolor[HTML]{F7C7AC}LOSDays                                                           &
          4.61\%                                                                                      \\
          \begin{tabular}[c]{@{}l@{}}FirstLocation\\ TypeCodeAfter\\ Arrival\end{tabular}           &
          3.30\%                                                                                    &
          \begin{tabular}[c]{@{}l@{}}FirstLocation\\ TypeCodeAfter\\ Arrival\end{tabular}           &
          \multicolumn{1}{c|}{3.60\%}                                                               &
          \begin{tabular}[c]{@{}l@{}}FirstLocation\\ TypeCodeAfter\\ Arrival\end{tabular}           &
          \multicolumn{1}{c|}{3.28\%}                                                               &
          \begin{tabular}[c]{@{}l@{}}FirstLocation\\ TypeCodeAfter\\ Arrival\end{tabular}           &
          \multicolumn{1}{c|}{3.50\%}                                                               &
          \begin{tabular}[c]{@{}l@{}}FirstLocation\\ TypeCodeAfter\\ Arrival\end{tabular}           &
          \multicolumn{1}{c|}{3.62\%}                                                               &
          \begin{tabular}[c]{@{}l@{}}FirstLocation\\ TypeCodeAfter\\ Arrival\end{tabular}           &
          \multicolumn{1}{c|}{3.38\%}                                                               &
          \begin{tabular}[c]{@{}l@{}}FirstLocation\\ TypeCodeAfter\\ Arrival\end{tabular}           &
          \multicolumn{1}{c|}{3.30\%}                                                               &
          \begin{tabular}[c]{@{}l@{}}FirstLocation\\ TypeCodeAfter\\ Arrival\end{tabular}           &
          3.56\%                                                                                      \\
          \begin{tabular}[c]{@{}l@{}}\textgreater{}6HoursToFirst\\ AntibioticAdmin\end{tabular}     &
          1.93\%                                                                                    &
          \begin{tabular}[c]{@{}l@{}}\textgreater{}6HoursToFirst\\ AntibioticAdmin\end{tabular}     &
          \multicolumn{1}{c|}{2.07\%}                                                               &
          \begin{tabular}[c]{@{}l@{}}\textgreater{}6HoursToFirst\\ AntibioticAdmin\end{tabular}     &
          \multicolumn{1}{c|}{2.13\%}                                                               &
          \cellcolor[HTML]{FFFF00}NumberofVisits                                                    &
          \multicolumn{1}{c|}{1.91\%}                                                               &
          \begin{tabular}[c]{@{}l@{}}\textgreater{}6HoursToFirst\\ AntibioticAdmin\end{tabular}     &
          \multicolumn{1}{c|}{1.99\%}                                                               &
          \begin{tabular}[c]{@{}l@{}}\textgreater{}6HoursToFirst\\ AntibioticAdmin\end{tabular}     &
          \multicolumn{1}{c|}{2.17\%}                                                               &
          \begin{tabular}[c]{@{}l@{}}\textgreater{}6HoursToFirst\\ AntibioticAdmin\end{tabular}     &
          \multicolumn{1}{c|}{2.06\%}                                                               &
          \begin{tabular}[c]{@{}l@{}}\textgreater{}6HoursToFirst\\ AntibioticAdmin\end{tabular}     &
          1.95\%                                                                                      \\
          LYTESFlag                                                                                 &
          1.75\%                                                                                    &
          LYTESFlag                                                                                 &
          \multicolumn{1}{c|}{1.81\%}                                                               &
          LYTESFlag                                                                                 &
          \multicolumn{1}{c|}{1.75\%}                                                               &
          \begin{tabular}[c]{@{}l@{}}\textgreater{}6HoursToFirst\\ AntibioticAdmin\end{tabular}     &
          \multicolumn{1}{c|}{1.81\%}                                                               &
          \cellcolor[HTML]{FFFF00}NumberofVisits                                                    &
          \multicolumn{1}{c|}{1.89\%}                                                               &
          \cellcolor[HTML]{FFFF00}NumberofVisits                                                    &
          \multicolumn{1}{c|}{2.00\%}                                                               &
          \cellcolor[HTML]{FFFF00}NumberofVisits                                                    &
          \multicolumn{1}{c|}{1.98\%}                                                               &
          \cellcolor[HTML]{FFFF00}NumberofVisits                                                    &
          1.80\%                                                                                      \\
          \cellcolor[HTML]{FFFF00}NumberofVisits                                                    &
          1.73\%                                                                                    &
          \cellcolor[HTML]{FFFF00}NumberofVisits                                                    &
          \multicolumn{1}{c|}{1.48\%}                                                               &
          \cellcolor[HTML]{FFFF00}NumberofVisits                                                    &
          \multicolumn{1}{c|}{1.71\%}                                                               &
          LYTESFlag                                                                                 &
          \multicolumn{1}{c|}{1.59\%}                                                               &
          LYTESFlag                                                                                 &
          \multicolumn{1}{c|}{1.80\%}                                                               &
          LYTESFlag                                                                                 &
          \multicolumn{1}{c|}{1.83\%}                                                               &
          LYTESFlag                                                                                 &
          \multicolumn{1}{c|}{1.81\%}                                                               &
          LYTESFlag                                                                                 &
          1.74\%                                                                                      \\
          \cellcolor[HTML]{E49EDD}\begin{tabular}[c]{@{}l@{}}Race\\ Description\end{tabular}        &
          1.61\%                                                                                    &
          \cellcolor[HTML]{E49EDD}\begin{tabular}[c]{@{}l@{}}Race\\ Description\end{tabular}        &
          \multicolumn{1}{c|}{1.29\%}                                                               &
          \cellcolor[HTML]{83E28E}\begin{tabular}[c]{@{}l@{}}Gender\\ Description\end{tabular}      &
          \multicolumn{1}{c|}{1.61\%}                                                               &
          \cellcolor[HTML]{E49EDD}\begin{tabular}[c]{@{}l@{}}Race\\ Description\end{tabular}        &
          \multicolumn{1}{c|}{1.35\%}                                                               &
          ANEMDEFFlag                                                                               &
          \multicolumn{1}{c|}{1.37\%}                                                               &
          FluSeasonFlag                                                                             &
          \multicolumn{1}{c|}{1.34\%}                                                               &
          FluSeasonFlag                                                                             &
          \multicolumn{1}{c|}{1.67\%}                                                               &
          HTNFlag                                                                                   &
          1.62\%                                                                                      \\
          \cellcolor[HTML]{83E28E}\begin{tabular}[c]{@{}l@{}}Gender\\ Description\end{tabular}      &
          1.28\%                                                                                    &
          CHRNLUNGFlag                                                                              &
          \multicolumn{1}{c|}{1.24\%}                                                               &
          HTNFlag                                                                                   &
          \multicolumn{1}{c|}{1.38\%}                                                               &
          FluSeasonFlag                                                                             &
          \multicolumn{1}{c|}{1.19\%}                                                               &
          \cellcolor[HTML]{83E28E}\begin{tabular}[c]{@{}l@{}}Gender\\ Description\end{tabular}      &
          \multicolumn{1}{c|}{1.24\%}                                                               &
          COAGFlag                                                                                  &
          \multicolumn{1}{c|}{1.24\%}                                                               &
          \cellcolor[HTML]{E49EDD}\begin{tabular}[c]{@{}l@{}}Race\\ Description\end{tabular}        &
          \multicolumn{1}{c|}{1.49\%}                                                               &
          FluSeasonFlag                                                                             &
          1.61\%                                                                                      \\
          FluSeasonFlag                                                                             &
          1.21\%                                                                                    &
          \cellcolor[HTML]{83E28E}\begin{tabular}[c]{@{}l@{}}Gender\\ Description\end{tabular}      &
          \multicolumn{1}{c|}{1.23\%}                                                               &
          \cellcolor[HTML]{E49EDD}\begin{tabular}[c]{@{}l@{}}Race\\ Description\end{tabular}        &
          \multicolumn{1}{c|}{1.32\%}                                                               &
          \cellcolor[HTML]{83E28E}\begin{tabular}[c]{@{}l@{}}Gender\\ Description\end{tabular}      &
          \multicolumn{1}{c|}{1.18\%}                                                               &
          FluSeasonFlag                                                                             &
          \multicolumn{1}{c|}{1.18\%}                                                               &
          ANEMDEFFlag                                                                               &
          \multicolumn{1}{c|}{1.15\%}                                                               &
          ANEMDEFFlag                                                                               &
          \multicolumn{1}{c|}{1.37\%}                                                               &
          NEUROFlag                                                                                 &
          1.25\%                                                                                      \\
          ANEMDEFFlag                                                                               &
          1.14\%                                                                                    &
          HTNFlag                                                                                   &
          \multicolumn{1}{c|}{1.18\%}                                                               &
          FluSeasonFlag                                                                             &
          \multicolumn{1}{c|}{1.22\%}                                                               &
          COAGFlag                                                                                  &
          \multicolumn{1}{c|}{1.12\%}                                                               &
          \cellcolor[HTML]{E49EDD}\begin{tabular}[c]{@{}l@{}}Race\\ Description\end{tabular}        &
          \multicolumn{1}{c|}{1.13\%}                                                               &
          NEUROFlag                                                                                 &
          \multicolumn{1}{c|}{1.08\%}                                                               &
          HTNFlag                                                                                   &
          \multicolumn{1}{c|}{1.12\%}                                                               &
          COAGFlag                                                                                  &
          1.24\%                                                                                      \\
          HTNFlag                                                                                   &
          1.02\%                                                                                    &
          DMFlag                                                                                    &
          \multicolumn{1}{c|}{1.11\%}                                                               &
          ANEMDEFFlag                                                                               &
          \multicolumn{1}{c|}{1.15\%}                                                               &
          ANEMDEFFlag                                                                               &
          \multicolumn{1}{c|}{1.11\%}                                                               &
          NEUROFlag                                                                                 &
          \multicolumn{1}{c|}{1.03\%}                                                               &
          \cellcolor[HTML]{E49EDD}\begin{tabular}[c]{@{}l@{}}Race\\ Description\end{tabular}        &
          \multicolumn{1}{c|}{1.03\%}                                                               &
          DMFlag                                                                                    &
          \multicolumn{1}{c|}{1.06\%}                                                               &
          ANEMDEFFlag                                                                               &
          1.16\%                                                                                      \\
          HX\_DEPRESS                                                                               &
          0.87\%                                                                                    &
          FluSeasonFlag                                                                             &
          \multicolumn{1}{c|}{1.07\%}                                                               &
          COAGFlag                                                                                  &
          \multicolumn{1}{c|}{1.07\%}                                                               &
          HTNFlag                                                                                   &
          \multicolumn{1}{c|}{1.09\%}                                                               &
          CHRNLUNGFlag                                                                              &
          \multicolumn{1}{c|}{0.99\%}                                                               &
          HTNFlag                                                                                   &
          \multicolumn{1}{c|}{0.93\%}                                                               &
          NEUROFlag                                                                                 &
          \multicolumn{1}{c|}{1.03\%}                                                               &
          DEPRESSFlag                                                                               &
          1.13\%                                                                                      \\
          COAGFlag                                                                                  &
          0.83\%                                                                                    &
          ANEMDEFFlag                                                                               &
          \multicolumn{1}{c|}{0.99\%}                                                               &
          NEUROFlag                                                                                 &
          \multicolumn{1}{c|}{0.98\%}                                                               &
          NEUROFlag                                                                                 &
          \multicolumn{1}{c|}{0.91\%}                                                               &
          HTNFlag                                                                                   &
          \multicolumn{1}{c|}{0.94\%}                                                               &
          \cellcolor[HTML]{83E28E}\begin{tabular}[c]{@{}l@{}}Gender\\ Description\end{tabular}      &
          \multicolumn{1}{c|}{0.89\%}                                                               &
          CHRNLUNGFlag                                                                              &
          \multicolumn{1}{c|}{1.02\%}                                                               &
          DMFlag                                                                                    &
          1.09\%                                                                                      \\
          DMFlag                                                                                    &
          0.79\%                                                                                    &
          COAGFlag                                                                                  &
          \multicolumn{1}{c|}{0.91\%}                                                               &
          CHFFlag                                                                                   &
          \multicolumn{1}{c|}{0.96\%}                                                               &
          HX\_Sepsis                                                                                &
          \multicolumn{1}{c|}{0.82\%}                                                               &
          OBESEFlag                                                                                 &
          \multicolumn{1}{c|}{0.90\%}                                                               &
          HX\_Sepsis                                                                                &
          \multicolumn{1}{c|}{0.83\%}                                                               &
          OBESEFlag                                                                                 &
          \multicolumn{1}{c|}{0.99\%}                                                               &
          HYPOTHYFlag                                                                               &
          1.05\%                                                                                      \\
          DEPRESSFlag                                                                               &
          0.77\%                                                                                    &
          HX\_Sepsis                                                                                &
          \multicolumn{1}{c|}{0.78\%}                                                               &
          HX\_Sepsis                                                                                &
          \multicolumn{1}{c|}{0.87\%}                                                               &
          \cellcolor[HTML]{BFCEF0}\begin{tabular}[c]{@{}l@{}}EthnicGroup\\ Description\end{tabular} &
          \multicolumn{1}{c|}{0.81\%}                                                               &
          HX\_Sepsis                                                                                &
          \multicolumn{1}{c|}{0.84\%}                                                               &
          CADFlag                                                                                   &
          \multicolumn{1}{c|}{0.82\%}                                                               &
          COAGFlag                                                                                  &
          \multicolumn{1}{c|}{0.99\%}                                                               &
          CHRNLUNGFlag                                                                              &
          1.01\%                                                                                      \\
          CHRNLUNGFlag                                                                              &
          0.76\%                                                                                    &
          HX\_HTN                                                                                   &
          \multicolumn{1}{c|}{0.75\%}                                                               &
          CHRNLUNGFlag                                                                              &
          \multicolumn{1}{c|}{0.85\%}                                                               &
          HX\_CAD                                                                                   &
          \multicolumn{1}{c|}{0.80\%}                                                               &
          DMFlag                                                                                    &
          \multicolumn{1}{c|}{0.82\%}                                                               &
          CHRNLUNGFlag                                                                              &
          \multicolumn{1}{c|}{0.80\%}                                                               &
          DEPRESSFlag                                                                               &
          \multicolumn{1}{c|}{0.95\%}                                                               &
          HX\_Sepsis                                                                                &
          0.99\%                                                                                      \\
          WGHTLOSSFlag                                                                              &
          0.75\%                                                                                    &
          HX\_LYTES                                                                                 &
          \multicolumn{1}{c|}{0.74\%}                                                               &
          DMFlag                                                                                    &
          \multicolumn{1}{c|}{0.77\%}                                                               &
          DMFlag                                                                                    &
          \multicolumn{1}{c|}{0.77\%}                                                               &
          CHFFlag                                                                                   &
          \multicolumn{1}{c|}{0.82\%}                                                               &
          WGHTLOSSFlag                                                                              &
          \multicolumn{1}{c|}{0.74\%}                                                               &
          HX\_Sepsis                                                                                &
          \multicolumn{1}{c|}{0.91\%}                                                               &
          \cellcolor[HTML]{83E28E}\begin{tabular}[c]{@{}l@{}}Gender\\ Description\end{tabular}      &
          0.96\%                                                                                      \\
          HX\_Sepsis                                                                                &
          0.71\%                                                                                    &
          CHFFlag                                                                                   &
          \multicolumn{1}{c|}{0.70\%}                                                               &
          HX\_CAD                                                                                   &
          \multicolumn{1}{c|}{0.71\%}                                                               &
          HX\_HTN                                                                                   &
          \multicolumn{1}{c|}{0.76\%}                                                               &
          COAGFlag                                                                                  &
          \multicolumn{1}{c|}{0.77\%}                                                               &
          VALVEFlag                                                                                 &
          \multicolumn{1}{c|}{0.69\%}                                                               &
          \cellcolor[HTML]{83E28E}\begin{tabular}[c]{@{}l@{}}Gender\\ Description\end{tabular}      &
          \multicolumn{1}{c|}{0.89\%}                                                               &
          CADFlag                                                                                   &
          0.94\%                                                                                      \\
          \begin{tabular}[c]{@{}l@{}}1-3HoursToFirst\\ AntibioticAdmin\end{tabular}                 &
          0.71\%                                                                                    &
          HX\_DEPRESS                                                                               &
          \multicolumn{1}{c|}{0.68\%}                                                               &
          PERIVASCFlag                                                                              &
          \multicolumn{1}{c|}{0.71\%}                                                               &
          DEPRESSFlag                                                                               &
          \multicolumn{1}{c|}{0.76\%}                                                               &
          DEPRESSFlag                                                                               &
          \multicolumn{1}{c|}{0.74\%}                                                               &
          HX\_RENLFAIL                                                                              &
          \multicolumn{1}{c|}{0.64\%}                                                               &
          CADFlag                                                                                   &
          \multicolumn{1}{c|}{0.79\%}                                                               &
          OBESEFlag                                                                                 &
          0.94\%                                                                                      \\
          HX\_HTN                                                                                   &
          0.70\%                                                                                    &
          NEUROFlag                                                                                 &
          \multicolumn{1}{c|}{0.68\%}                                                               &
          DEPRESSFlag                                                                               &
          \multicolumn{1}{c|}{0.68\%}                                                               &
          CADFlag                                                                                   &
          \multicolumn{1}{c|}{0.66\%}                                                               &
          CADFlag                                                                                   &
          \multicolumn{1}{c|}{0.74\%}                                                               &
          RENLFAILFlag                                                                              &
          \multicolumn{1}{c|}{0.64\%}                                                               &
          HX\_LYTES                                                                                 &
          \multicolumn{1}{c|}{0.73\%}                                                               &
          \cellcolor[HTML]{E49EDD}\begin{tabular}[c]{@{}l@{}}Race\\ Description\end{tabular}        &
          0.88\%                                                                                      \\
          NEUROFlag                                                                                 &
          0.66\%                                                                                    &
          HX\_HYPOTHY                                                                               &
          \multicolumn{1}{c|}{0.67\%}                                                               &
          \cellcolor[HTML]{BFCEF0}\begin{tabular}[c]{@{}l@{}}EthnicGroup\\ Description\end{tabular} &
          \multicolumn{1}{c|}{0.65\%}                                                               &
          WGHTLOSSFlag                                                                              &
          \multicolumn{1}{c|}{0.66\%}                                                               &
          RENLFAILFlag                                                                              &
          \multicolumn{1}{c|}{0.68\%}                                                               &
          HYPOTHYFlag                                                                               &
          \multicolumn{1}{c|}{0.63\%}                                                               &
          WGHTLOSSFlag                                                                              &
          \multicolumn{1}{c|}{0.71\%}                                                               &
          WGHTLOSSFlag                                                                              &
          0.86\%                                                                                      \\
          RENLFAILFlag                                                                              &
          0.66\%                                                                                    &
          VALVEFlag                                                                                 &
          \multicolumn{1}{c|}{0.66\%}                                                               &
          CADFlag                                                                                   &
          \multicolumn{1}{c|}{0.64\%}                                                               &
          PERIVASCFlag                                                                              &
          \multicolumn{1}{c|}{0.64\%}                                                               &
          HX\_CHRNLUNG                                                                              &
          \multicolumn{1}{c|}{0.67\%}                                                               &
          HX\_CAD                                                                                   &
          \multicolumn{1}{c|}{0.62\%}                                                               &
          PSYCHFlag                                                                                 &
          \multicolumn{1}{c|}{0.71\%}                                                               &
          HX\_CHRNLUNG                                                                              &
          0.81\%                                                                                      \\
          HX\_ANEMDEF                                                                               &
          0.64\%                                                                                    &
          \cellcolor[HTML]{BFCEF0}\begin{tabular}[c]{@{}l@{}}EthnicGroup\\ Description\end{tabular} &
          \multicolumn{1}{c|}{0.66\%}                                                               &
          HYPOTHYFlag                                                                               &
          \multicolumn{1}{c|}{0.63\%}                                                               &
          HX\_ANEMDEF                                                                               &
          \multicolumn{1}{c|}{0.64\%}                                                               &
          \cellcolor[HTML]{BFCEF0}\begin{tabular}[c]{@{}l@{}}EthnicGroup\\ Description\end{tabular} &
          \multicolumn{1}{c|}{0.66\%}                                                               &
          HX\_CHRNLUNG                                                                              &
          \multicolumn{1}{c|}{0.61\%}                                                               &
          RENLFAILFlag                                                                              &
          \multicolumn{1}{c|}{0.69\%}                                                               &
          PULMCIRCFlag                                                                              &
          0.78\%                                                                                      \\
        \end{tabular}%
      }
    \end{tabular}
  \end{table*}

  \begin{table*}[!ht]
    \centering
    \begin{tabular}{c}
      (c) MO-OBAM \\
      \resizebox{\textwidth}{!}{%
        \begin{tabular}{m{2cm}c|m{2cm}cm{2cm}cm{2cm}cm{2cm}cm{2cm}cm{2cm}cm{2cm}c}
          \hline
          \multicolumn{2}{c|}{Original   Data}                                                      &
          \multicolumn{14}{c}{MO-OBAM}                                                                \\ \hline
          Feature                                                                                   &
          Importance                                                                                &
          $n_C$=3240,$k$=5                                                                          &
          \multicolumn{1}{c|}{Importance}                                                           &
          $n_C$=2310,$k$=10                                                                         &
          \multicolumn{1}{c|}{Importance}                                                           &
          $n_C$=1740,$k$=15                                                                         &
          \multicolumn{1}{c|}{Importance}                                                           &
          $n_C$=820,$k$=20                                                                          &
          \multicolumn{1}{c|}{Importance}                                                           &
          $n_C$=20,$k$=100                                                                          &
          \multicolumn{1}{c|}{Importance}                                                           &
          $n_C$=10,$k$=300                                                                          &
          \multicolumn{1}{c|}{Importance}                                                           &
          $n_C$=4,$k$=2000                                                                          &
          Importance                                                                                  \\ \hline
          \begin{tabular}[c]{@{}l@{}}Antibiotic\\ \_AdminFlag\end{tabular}                          &
          37.96\%                                                                                   &
          \begin{tabular}[c]{@{}l@{}}Antibiotic\\ \_AdminFlag\end{tabular}                          &
          \multicolumn{1}{c|}{38.48\%}                                                              &
          \begin{tabular}[c]{@{}l@{}}Antibiotic\\ \_AdminFlag\end{tabular}                          &
          \multicolumn{1}{c|}{38.65\%}                                                              &
          \begin{tabular}[c]{@{}l@{}}Antibiotic\\ \_AdminFlag\end{tabular}                          &
          \multicolumn{1}{c|}{38.05\%}                                                              &
          \begin{tabular}[c]{@{}l@{}}Antibiotic\\ \_AdminFlag\end{tabular}                          &
          \multicolumn{1}{c|}{38.07\%}                                                              &
          \begin{tabular}[c]{@{}l@{}}Antibiotic\\ \_AdminFlag\end{tabular}                          &
          \multicolumn{1}{c|}{31.31\%}                                                              &
          \begin{tabular}[c]{@{}l@{}}Antibiotic\\ \_AdminFlag\end{tabular}                          &
          \multicolumn{1}{c|}{31.36\%}                                                              &
          \begin{tabular}[c]{@{}l@{}}Antibiotic\\ \_AdminFlag\end{tabular}                          &
          32.11\%                                                                                     \\
          \cellcolor[HTML]{83CCEB}AgeCategory                                                       &
          9.95\%                                                                                    &
          \cellcolor[HTML]{83CCEB}AgeCategory                                                       &
          \multicolumn{1}{c|}{9.87\%}                                                               &
          \cellcolor[HTML]{83CCEB}AgeCategory                                                       &
          \multicolumn{1}{c|}{9.16\%}                                                               &
          \cellcolor[HTML]{83CCEB}AgeCategory                                                       &
          \multicolumn{1}{c|}{9.35\%}                                                               &
          \cellcolor[HTML]{F7C7AC}LOSDays                                                           &
          \multicolumn{1}{c|}{9.95\%}                                                               &
          LYTESFlag                                                                                 &
          \multicolumn{1}{c|}{3.81\%}                                                               &
          LYTESFlag                                                                                 &
          \multicolumn{1}{c|}{3.84\%}                                                               &
          LYTESFlag                                                                                 &
          3.84\%                                                                                      \\
          \cellcolor[HTML]{F7C7AC}LOSDays                                                           &
          9.09\%                                                                                    &
          \cellcolor[HTML]{F7C7AC}LOSDays                                                           &
          \multicolumn{1}{c|}{8.93\%}                                                               &
          \cellcolor[HTML]{F7C7AC}LOSDays                                                           &
          \multicolumn{1}{c|}{8.90\%}                                                               &
          \cellcolor[HTML]{F7C7AC}LOSDays                                                           &
          \multicolumn{1}{c|}{9.21\%}                                                               &
          \cellcolor[HTML]{83CCEB}AgeCategory                                                       &
          \multicolumn{1}{c|}{8.31\%}                                                               &
          \begin{tabular}[c]{@{}l@{}}FirstLocation\\ TypeCodeAfter\\ Arrival\end{tabular}           &
          \multicolumn{1}{c|}{3.68\%}                                                               &
          \begin{tabular}[c]{@{}l@{}}FirstLocation\\ TypeCodeAfter\\ Arrival\end{tabular}           &
          \multicolumn{1}{c|}{3.75\%}                                                               &
          \begin{tabular}[c]{@{}l@{}}FirstLocation\\ TypeCodeAfter\\ Arrival\end{tabular}           &
          3.81\%                                                                                      \\
          \begin{tabular}[c]{@{}l@{}}FirstLocation\\ TypeCodeAfter\\ Arrival\end{tabular}           &
          3.30\%                                                                                    &
          \begin{tabular}[c]{@{}l@{}}FirstLocation\\ TypeCodeAfter\\ Arrival\end{tabular}           &
          \multicolumn{1}{c|}{3.62\%}                                                               &
          \begin{tabular}[c]{@{}l@{}}FirstLocation\\ TypeCodeAfter\\ Arrival\end{tabular}           &
          \multicolumn{1}{c|}{3.34\%}                                                               &
          \begin{tabular}[c]{@{}l@{}}FirstLocation\\ TypeCodeAfter\\ Arrival\end{tabular}           &
          \multicolumn{1}{c|}{3.33\%}                                                               &
          \cellcolor[HTML]{FFFF00}NumberofVisits                                                    &
          \multicolumn{1}{c|}{3.54\%}                                                               &
          \cellcolor[HTML]{F7C7AC}LOSDays                                                           &
          \multicolumn{1}{c|}{3.44\%}                                                               &
          \cellcolor[HTML]{F7C7AC}LOSDays                                                           &
          \multicolumn{1}{c|}{3.15\%}                                                               &
          FluSeasonFlag                                                                             &
          2.49\%                                                                                      \\
          \begin{tabular}[c]{@{}l@{}}\textgreater{}6HoursToFirst\\ AntibioticAdmin\end{tabular}     &
          1.93\%                                                                                    &
          \cellcolor[HTML]{FFFF00}NumberofVisits                                                    &
          \multicolumn{1}{c|}{2.32\%}                                                               &
          \cellcolor[HTML]{FFFF00}NumberofVisits                                                    &
          \multicolumn{1}{c|}{2.78\%}                                                               &
          \cellcolor[HTML]{FFFF00}NumberofVisits                                                    &
          \multicolumn{1}{c|}{2.50\%}                                                               &
          \begin{tabular}[c]{@{}l@{}}FirstLocation\\ TypeCodeAfter\\ Arrival\end{tabular}           &
          \multicolumn{1}{c|}{3.49\%}                                                               &
          \begin{tabular}[c]{@{}l@{}}\textgreater{}6HoursToFirst\\ AntibioticAdmin\end{tabular}     &
          \multicolumn{1}{c|}{2.34\%}                                                               &
          FluSeasonFlag                                                                             &
          \multicolumn{1}{c|}{2.49\%}                                                               &
          \begin{tabular}[c]{@{}l@{}}\textgreater{}6HoursToFirst\\ AntibioticAdmin\end{tabular}     &
          2.39\%                                                                                      \\
          LYTESFlag                                                                                 &
          1.75\%                                                                                    &
          \begin{tabular}[c]{@{}l@{}}\textgreater{}6HoursToFirst\\ AntibioticAdmin\end{tabular}     &
          \multicolumn{1}{c|}{2.16\%}                                                               &
          \begin{tabular}[c]{@{}l@{}}\textgreater{}6HoursToFirst\\ AntibioticAdmin\end{tabular}     &
          \multicolumn{1}{c|}{2.04\%}                                                               &
          \begin{tabular}[c]{@{}l@{}}\textgreater{}6HoursToFirst\\ AntibioticAdmin\end{tabular}     &
          \multicolumn{1}{c|}{2.00\%}                                                               &
          \begin{tabular}[c]{@{}l@{}}\textgreater{}6HoursToFirst\\ AntibioticAdmin\end{tabular}     &
          \multicolumn{1}{c|}{1.92\%}                                                               &
          \cellcolor[HTML]{83CCEB}AgeCategory                                                       &
          \multicolumn{1}{c|}{2.34\%}                                                               &
          \begin{tabular}[c]{@{}l@{}}\textgreater{}6HoursToFirst\\ AntibioticAdmin\end{tabular}     &
          \multicolumn{1}{c|}{2.22\%}                                                               &
          ANEMDEFFlag                                                                               &
          1.76\%                                                                                      \\
          \cellcolor[HTML]{FFFF00}NumberofVisits                                                    &
          1.73\%                                                                                    &
          LYTESFlag                                                                                 &
          \multicolumn{1}{c|}{1.78\%}                                                               &
          LYTESFlag                                                                                 &
          \multicolumn{1}{c|}{1.48\%}                                                               &
          LYTESFlag                                                                                 &
          \multicolumn{1}{c|}{1.63\%}                                                               &
          LYTESFlag                                                                                 &
          \multicolumn{1}{c|}{1.74\%}                                                               &
          FluSeasonFlag                                                                             &
          \multicolumn{1}{c|}{2.32\%}                                                               &
          HTNFlag                                                                                   &
          \multicolumn{1}{c|}{1.69\%}                                                               &
          CHRNLUNGFlag                                                                              &
          1.52\%                                                                                      \\
          \cellcolor[HTML]{E49EDD}\begin{tabular}[c]{@{}l@{}}Race\\ Description\end{tabular}        &
          1.61\%                                                                                    &
          \cellcolor[HTML]{83E28E}\begin{tabular}[c]{@{}l@{}}Gender\\ Description\end{tabular}      &
          \multicolumn{1}{c|}{1.31\%}                                                               &
          \cellcolor[HTML]{E49EDD}\begin{tabular}[c]{@{}l@{}}Race\\ Description\end{tabular}        &
          \multicolumn{1}{c|}{1.25\%}                                                               &
          FluSeasonFlag                                                                             &
          \multicolumn{1}{c|}{1.50\%}                                                               &
          HTNFlag                                                                                   &
          \multicolumn{1}{c|}{1.25\%}                                                               &
          \cellcolor[HTML]{E49EDD}\begin{tabular}[c]{@{}l@{}}Race\\ Description\end{tabular}        &
          \multicolumn{1}{c|}{1.86\%}                                                               &
          \cellcolor[HTML]{83CCEB}AgeCategory                                                       &
          \multicolumn{1}{c|}{1.57\%}                                                               &
          DMFlag                                                                                    &
          1.50\%                                                                                      \\
          \cellcolor[HTML]{83E28E}\begin{tabular}[c]{@{}l@{}}Gender\\ Description\end{tabular}      &
          1.28\%                                                                                    &
          \cellcolor[HTML]{E49EDD}\begin{tabular}[c]{@{}l@{}}Race\\ Description\end{tabular}        &
          \multicolumn{1}{c|}{1.16\%}                                                               &
          FluSeasonFlag                                                                             &
          \multicolumn{1}{c|}{1.23\%}                                                               &
          ANEMDEFFlag                                                                               &
          \multicolumn{1}{c|}{1.40\%}                                                               &
          COAGFlag                                                                                  &
          \multicolumn{1}{c|}{1.19\%}                                                               &
          ANEMDEFFlag                                                                               &
          \multicolumn{1}{c|}{1.61\%}                                                               &
          CHRNLUNGFlag                                                                              &
          \multicolumn{1}{c|}{1.44\%}                                                               &
          DEPRESSFlag                                                                               &
          1.49\%                                                                                      \\
          FluSeasonFlag                                                                             &
          1.21\%                                                                                    &
          CHRNLUNGFlag                                                                              &
          \multicolumn{1}{c|}{1.09\%}                                                               &
          ANEMDEFFlag                                                                               &
          \multicolumn{1}{c|}{1.15\%}                                                               &
          DMFlag                                                                                    &
          \multicolumn{1}{c|}{1.23\%}                                                               &
          ANEMDEFFlag                                                                               &
          \multicolumn{1}{c|}{1.12\%}                                                               &
          CHRNLUNGFlag                                                                              &
          \multicolumn{1}{c|}{1.50\%}                                                               &
          \cellcolor[HTML]{FFFF00}NumberofVisits                                                    &
          \multicolumn{1}{c|}{1.36\%}                                                               &
          HTNFlag                                                                                   &
          1.48\%                                                                                      \\
          ANEMDEFFlag                                                                               &
          1.14\%                                                                                    &
          HTNFlag                                                                                   &
          \multicolumn{1}{c|}{1.09\%}                                                               &
          \cellcolor[HTML]{83E28E}\begin{tabular}[c]{@{}l@{}}Gender\\ Description\end{tabular}      &
          \multicolumn{1}{c|}{1.11\%}                                                               &
          HTNFlag                                                                                   &
          \multicolumn{1}{c|}{1.15\%}                                                               &
          FluSeasonFlag                                                                             &
          \multicolumn{1}{c|}{1.10\%}                                                               &
          HTNFlag                                                                                   &
          \multicolumn{1}{c|}{1.42\%}                                                               &
          NEUROFlag                                                                                 &
          \multicolumn{1}{c|}{1.35\%}                                                               &
          HX\_BLDLOSS                                                                               &
          1.42\%                                                                                      \\
          HTNFlag                                                                                   &
          1.02\%                                                                                    &
          ANEMDEFFlag                                                                               &
          \multicolumn{1}{c|}{1.04\%}                                                               &
          COAGFlag                                                                                  &
          \multicolumn{1}{c|}{1.00\%}                                                               &
          \cellcolor[HTML]{83E28E}\begin{tabular}[c]{@{}l@{}}Gender\\ Description\end{tabular}      &
          \multicolumn{1}{c|}{1.00\%}                                                               &
          CHRNLUNGFlag                                                                              &
          \multicolumn{1}{c|}{1.06\%}                                                               &
          CADFlag                                                                                   &
          \multicolumn{1}{c|}{1.32\%}                                                               &
          HX\_ULCER                                                                                 &
          \multicolumn{1}{c|}{1.35\%}                                                               &
          CADFlag                                                                                   &
          1.41\%                                                                                      \\
          HX\_DEPRESS                                                                               &
          0.87\%                                                                                    &
          FluSeasonFlag                                                                             &
          \multicolumn{1}{c|}{0.93\%}                                                               &
          CADFlag                                                                                   &
          \multicolumn{1}{c|}{0.97\%}                                                               &
          OBESEFlag                                                                                 &
          \multicolumn{1}{c|}{0.95\%}                                                               &
          NEUROFlag                                                                                 &
          \multicolumn{1}{c|}{1.06\%}                                                               &
          DMFlag                                                                                    &
          \multicolumn{1}{c|}{1.31\%}                                                               &
          DMFlag                                                                                    &
          \multicolumn{1}{c|}{1.25\%}                                                               &
          OBESEFlag                                                                                 &
          1.25\%                                                                                      \\
          COAGFlag                                                                                  &
          0.83\%                                                                                    &
          HX\_Sepsis                                                                                &
          \multicolumn{1}{c|}{0.86\%}                                                               &
          HTNFlag                                                                                   &
          \multicolumn{1}{c|}{0.97\%}                                                               &
          CHRNLUNGFlag                                                                              &
          \multicolumn{1}{c|}{0.95\%}                                                               &
          HX\_Sepsis                                                                                &
          \multicolumn{1}{c|}{0.95\%}                                                               &
          NEUROFlag                                                                                 &
          \multicolumn{1}{c|}{1.30\%}                                                               &
          HYPOTHYFlag                                                                               &
          \multicolumn{1}{c|}{1.25\%}                                                               &
          CHFFlag                                                                                   &
          1.20\%                                                                                      \\
          DMFlag                                                                                    &
          0.79\%                                                                                    &
          NEUROFlag                                                                                 &
          \multicolumn{1}{c|}{0.85\%}                                                               &
          CHRNLUNGFlag                                                                              &
          \multicolumn{1}{c|}{0.89\%}                                                               &
          HX\_Sepsis                                                                                &
          \multicolumn{1}{c|}{0.92\%}                                                               &
          \cellcolor[HTML]{83E28E}\begin{tabular}[c]{@{}l@{}}Gender\\ Description\end{tabular}      &
          \multicolumn{1}{c|}{0.77\%}                                                               &
          HX\_BLDLOSS                                                                               &
          \multicolumn{1}{c|}{1.28\%}                                                               &
          ANEMDEFFlag                                                                               &
          \multicolumn{1}{c|}{1.23\%}                                                               &
          NEUROFlag                                                                                 &
          1.17\%                                                                                      \\
          DEPRESSFlag                                                                               &
          0.77\%                                                                                    &
          DMFlag                                                                                    &
          \multicolumn{1}{c|}{0.81\%}                                                               &
          DEPRESSFlag                                                                               &
          \multicolumn{1}{c|}{0.82\%}                                                               &
          NEUROFlag                                                                                 &
          \multicolumn{1}{c|}{0.92\%}                                                               &
          DMFlag                                                                                    &
          \multicolumn{1}{c|}{0.76\%}                                                               &
          DEPRESSFlag                                                                               &
          \multicolumn{1}{c|}{1.17\%}                                                               &
          OBESEFlag                                                                                 &
          \multicolumn{1}{c|}{1.19\%}                                                               &
          \cellcolor[HTML]{F7C7AC}LOSDays                                                           &
          1.05\%                                                                                      \\
          CHRNLUNGFlag                                                                              &
          0.76\%                                                                                    &
          COAGFlag                                                                                  &
          \multicolumn{1}{c|}{0.77\%}                                                               &
          NEUROFlag                                                                                 &
          \multicolumn{1}{c|}{0.81\%}                                                               &
          COAGFlag                                                                                  &
          \multicolumn{1}{c|}{0.90\%}                                                               &
          CHFFlag                                                                                   &
          \multicolumn{1}{c|}{0.75\%}                                                               &
          OBESEFlag                                                                                 &
          \multicolumn{1}{c|}{1.16\%}                                                               &
          CADFlag                                                                                   &
          \multicolumn{1}{c|}{1.18\%}                                                               &
          HX\_CHRNLUNG                                                                              &
          1.04\%                                                                                      \\
          WGHTLOSSFlag                                                                              &
          0.75\%                                                                                    &
          CHFFlag                                                                                   &
          \multicolumn{1}{c|}{0.71\%}                                                               &
          DMFlag                                                                                    &
          \multicolumn{1}{c|}{0.80\%}                                                               &
          \cellcolor[HTML]{E49EDD}\begin{tabular}[c]{@{}l@{}}Race\\ Description\end{tabular}        &
          \multicolumn{1}{c|}{0.87\%}                                                               &
          HYPOTHYFlag                                                                               &
          \multicolumn{1}{c|}{0.74\%}                                                               &
          \cellcolor[HTML]{FFFF00}NumberofVisits                                                    &
          \multicolumn{1}{c|}{1.12\%}                                                               &
          COAGFlag                                                                                  &
          \multicolumn{1}{c|}{1.13\%}                                                               &
          RENLFAILFlag                                                                              &
          0.99\%                                                                                      \\
          HX\_Sepsis                                                                                &
          0.71\%                                                                                    &
          CADFlag                                                                                   &
          \multicolumn{1}{c|}{0.69\%}                                                               &
          OBESEFlag                                                                                 &
          \multicolumn{1}{c|}{0.75\%}                                                               &
          DEPRESSFlag                                                                               &
          \multicolumn{1}{c|}{0.87\%}                                                               &
          HX\_HTN                                                                                   &
          \multicolumn{1}{c|}{0.72\%}                                                               &
          HYPOTHYFlag                                                                               &
          \multicolumn{1}{c|}{1.08\%}                                                               &
          DEPRESSFlag                                                                               &
          \multicolumn{1}{c|}{1.11\%}                                                               &
          HX\_HTN                                                                                   &
          0.98\%                                                                                      \\
          \begin{tabular}[c]{@{}l@{}}1-3HoursToFirst\\ AntibioticAdmin\end{tabular}                 &
          0.71\%                                                                                    &
          \begin{tabular}[c]{@{}l@{}}3-6HoursToFirst\\ AntibioticAdmin\end{tabular}                 &
          \multicolumn{1}{c|}{0.69\%}                                                               &
          HYPOTHYFlag                                                                               &
          \multicolumn{1}{c|}{0.73\%}                                                               &
          CADFlag                                                                                   &
          \multicolumn{1}{c|}{0.62\%}                                                               &
          \cellcolor[HTML]{E49EDD}\begin{tabular}[c]{@{}l@{}}Race\\ Description\end{tabular}        &
          \multicolumn{1}{c|}{0.69\%}                                                               &
          \cellcolor[HTML]{BFCEF0}\begin{tabular}[c]{@{}l@{}}EthnicGroup\\ Description\end{tabular} &
          \multicolumn{1}{c|}{1.06\%}                                                               &
          RENLFAILFlag                                                                              &
          \multicolumn{1}{c|}{1.05\%}                                                               &
          HYPOTHYFlag                                                                               &
          0.96\%                                                                                      \\
          HX\_HTN                                                                                   &
          0.70\%                                                                                    &
          DEPRESSFlag                                                                               &
          \multicolumn{1}{c|}{0.69\%}                                                               &
          HX\_HTN                                                                                   &
          \multicolumn{1}{c|}{0.72\%}                                                               &
          HX\_LYTES                                                                                 &
          \multicolumn{1}{c|}{0.61\%}                                                               &
          VALVEFlag                                                                                 &
          \multicolumn{1}{c|}{0.68\%}                                                               &
          HX\_HTN                                                                                   &
          \multicolumn{1}{c|}{0.94\%}                                                               &
          CHFFlag                                                                                   &
          \multicolumn{1}{c|}{1.00\%}                                                               &
          HX\_OBESE                                                                                 &
          0.95\%                                                                                      \\
          NEUROFlag                                                                                 &
          0.66\%                                                                                    &
          HX\_HTN                                                                                   &
          \multicolumn{1}{c|}{0.68\%}                                                               &
          HX\_Sepsis                                                                                &
          \multicolumn{1}{c|}{0.72\%}                                                               &
          \begin{tabular}[c]{@{}l@{}}1-3HoursToFirst\\ AntibioticAdmin\end{tabular}                 &
          \multicolumn{1}{c|}{0.61\%}                                                               &
          CADFlag                                                                                   &
          \multicolumn{1}{c|}{0.68\%}                                                               &
          COAGFlag                                                                                  &
          \multicolumn{1}{c|}{0.92\%}                                                               &
          PERIVASCFlag                                                                              &
          \multicolumn{1}{c|}{0.92\%}                                                               &
          COAGFlag                                                                                  &
          0.88\%                                                                                      \\
          RENLFAILFlag                                                                              &
          0.66\%                                                                                    &
          HYPOTHYFlag                                                                               &
          \multicolumn{1}{c|}{0.65\%}                                                               &
          \begin{tabular}[c]{@{}l@{}}3-6HoursToFirst\\ AntibioticAdmin\end{tabular}                 &
          \multicolumn{1}{c|}{0.65\%}                                                               &
          CHFFlag                                                                                   &
          \multicolumn{1}{c|}{0.61\%}                                                               &
          DMCXFlag                                                                                  &
          \multicolumn{1}{c|}{0.66\%}                                                               &
          CHFFlag                                                                                   &
          \multicolumn{1}{c|}{0.91\%}                                                               &
          \cellcolor[HTML]{E49EDD}\begin{tabular}[c]{@{}l@{}}Race\\ Description\end{tabular}        &
          \multicolumn{1}{c|}{0.89\%}                                                               &
          PULMCIRCFlag                                                                              &
          0.87\%                                                                                      \\
          HX\_ANEMDEF                                                                               &
          0.64\%                                                                                    &
          HX\_DEPRESS                                                                               &
          \multicolumn{1}{c|}{0.64\%}                                                               &
          HX\_CHRNLUNG                                                                              &
          \multicolumn{1}{c|}{0.61\%}                                                               &
          HX\_DM                                                                                    &
          \multicolumn{1}{c|}{0.54\%}                                                               &
          WGHTLOSSFlag                                                                              &
          \multicolumn{1}{c|}{0.62\%}                                                               &
          VALVEFlag                                                                                 &
          \multicolumn{1}{c|}{0.83\%}                                                               &
          \cellcolor[HTML]{BFCEF0}\begin{tabular}[c]{@{}l@{}}EthnicGroup\\ Description\end{tabular} &
          \multicolumn{1}{c|}{0.88\%}                                                               &
          HX\_LYTES                                                                                 &
          0.87\%                                                                                      \\
        \end{tabular}%
      }
    \end{tabular}
  \end{table*}

\end{appendices}

\end{document}